%% file: main_pami.tex
\documentclass[10pt,journal,compsoc]{IEEEtran}
\ifCLASSOPTIONcompsoc
  \usepackage[nocompress]{cite}
\else
  \usepackage{cite}
\fi
\ifCLASSINFOpdf
   \usepackage[pdftex]{graphicx}
\else
\fi
\usepackage{pgfplots}
\usepackage{comment}

\usepackage{amsmath}
\usepackage[caption=false,font=footnotesize]{subfig}
\usepackage{hyperref}

\usepackage{soul}
\usepackage{xcolor}

\newcommand{\revised}[1]{{\color{black}#1}}
\usepackage{multirow}

\newdimen\figrasterwd
\figrasterwd\columnwidth

\usepackage{comment}
\usepackage{booktabs}
\pgfplotsset{compat=1.12}

\usepackage{gensymb}

\begin{document}
%
\title{Generalized Contrastive Optimization of Siamese Networks for Place Recognition}
%
%
%
%

\author{Mar\'ia~Leyva-Vallina,
        Nicola~Strisciuglio,
        and~Nicolai~Petkov
\IEEEcompsocitemizethanks{\IEEEcompsocthanksitem M. Leyva-Vallina and N. Petkov are with the Intelligent Systems Group of the Bernoulli Institute, at the University of Groningen, the Netherlands.
\IEEEcompsocthanksitem 
N. Strisciuglio is with the Faculty of Electrical Engineering, Mathematics and Computer Science of the University of Twente, the Netherlands.
\IEEEcompsocthanksitem 
E-mail: m.leyva.vallina@rug.nl, n.strisciuglio@utwente.nl
}
}

%
%

\markboth{
}%
{Leyva-Vallina \MakeLowercase{\textit{et al.}}: Generalized Contrastive Optimization of Siamese Networks for Place Recognition}
%



\IEEEtitleabstractindextext{%
\begin{abstract}
Visual place recognition is a challenging task in computer vision and a key component of camera-based localization and navigation systems. 
Recently, Convolutional Neural Networks (CNNs) achieved high results and good generalization capabilities. They are usually trained using pairs or triplets of images labeled as either \emph{similar} or \emph{dissimilar}, in a binary fashion. In practice, the similarity between two images is not binary, but continuous. Furthermore, training these CNNs is computationally complex and involves costly pair and triplet mining strategies.  

We propose a Generalized Contrastive loss (GCL) function that relies on image similarity as a continuous measure, and use it to train a siamese CNN. Furthermore, we \revised{present} three techniques for automatic annotation of image pairs with labels indicating their degree of similarity, and deploy them to re-annotate the MSLS, TB-Places, and 7Scenes datasets. 

We demonstrate that siamese CNNs trained using the GCL function and the improved annotations consistently outperform their binary counterparts. Our models trained on MSLS outperform the state-of-the-art methods, including NetVLAD\revised{, NetVLAD-SARE, AP-GeM and Patch-NetVLAD}, and generalize well on the Pittsburgh\revised{30k, Tokyo 24/7, RobotCar Seasons v2 and Extended CMU Seasons}  datasets. Furthermore, training a siamese network using the GCL function does not require complex pair mining. We release the source code at \url{https://github.com/marialeyvallina/generalized_contrastive_loss}.
\end{abstract}

}

\maketitle

\IEEEdisplaynontitleabstractindextext

%
\IEEEpeerreviewmaketitle

\IEEEraisesectionheading{\section{Introduction}}
\input{sections/1_intro}

\section{Related works}

\input{sections/2_previouswork}

\section{Methodology}
\input{sections/3_methodology}
\section{Training data and automatic labelling}
\input{sections/4_data}

\section{Experimental framework}
\input{sections/5_experiments}

\section{Results and discussion}
\input{sections/6_results}

\section{Conclusions}
\input{sections/8_conclusions}

\section*{Acknowledgements}
This research received partial support from the EU H2020 TrimBot2020 project (grant no. 688007).
\\
We would like to thank the Center for Information Technology of the University of Groningen for their support and for providing access to the Peregrine HPC cluster.
\\ Finally, we would like to thank Dr. Manuel López Antequera for his assistance with the MSLS dataset.

\bibliographystyle{IEEEtran}
\bibliography{egbib}
%

%
\input{sections/bios}




\break
\appendices

\input{sections/appendix_extra_results_msls}

\input{sections/appendix_fov}
\input{sections/appendix_gradients}

\end{document}

%% file: sections/1_intro.tex
\begin{figure*}
    \centering
\subfloat[Query]{
    \includegraphics[scale=.19]{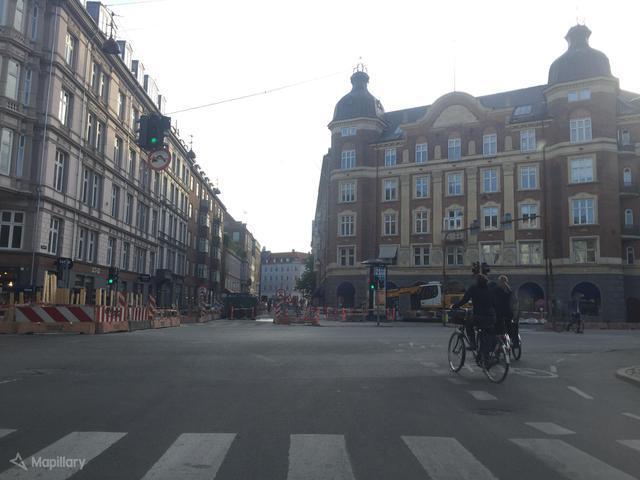}
    \label{fig:msls_query_example}
    }
   \subfloat[]{
    \includegraphics[scale=.19]{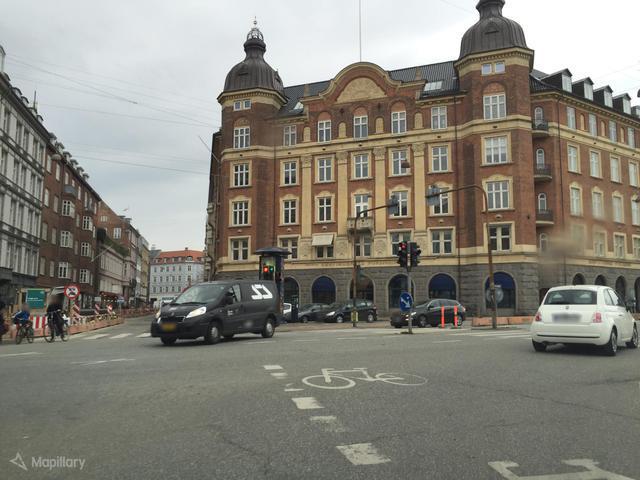}
    \label{fig:msls_match_5539}
     }\subfloat[]{
    \includegraphics[scale=.19]{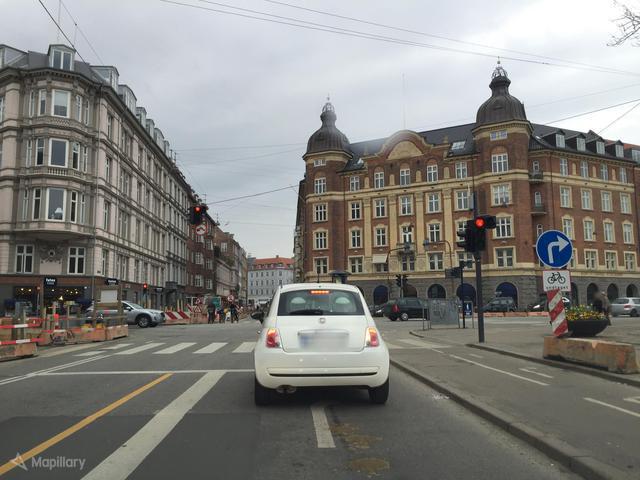}
        \label{fig:msls_match_7561}
     }
     \subfloat[]{
    \includegraphics[scale=.19]{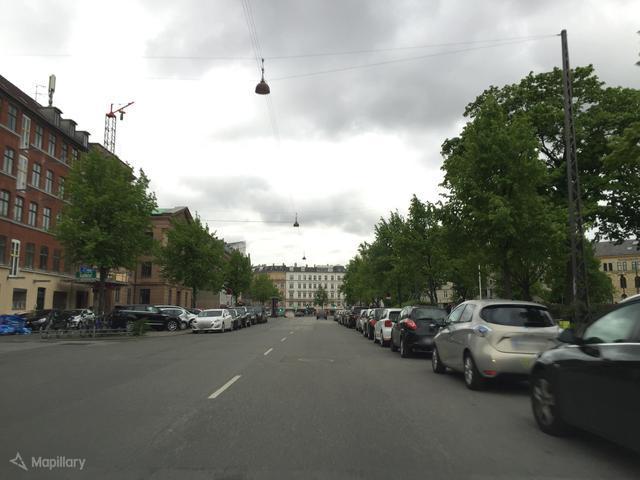}
    \label{fig:msls_match_9316}
    }
    \caption{Examples from MSLS dataset of a query image (a) and three matches with different degrees of similarity. A very close match is (b) with 86\% of overlap, while (c) is a borderline case with 52\% of commonality, and (d) is a negative match, with no features in common with the query.}
    \label{fig:msls_similarity_example}
\end{figure*}
\IEEEPARstart{V}isual place recognition has received large interest from researchers in computer vision, machine learning and information retrieval. It consists of, given a query image, seeking an image depicting a similar scene in a map or database. Possible instances of this problem are the retrieval of an image containing a specific distinctive landmark~\cite{philbin2007object, philbin2008lost} or the recognition of a previously visited place~\cite{Lowry2016,Arandjelovic2017}. Algorithms for image retrieval and place recognition are deployed in and are a key component of many visual localization~\cite{Torii-PAMI2015} and Simultaneous Localization and Mapping (SLAM) systems~\cite{Milford2012}. In these applications, effectively retrieving similar images from a map set is of utmost importance to facilitate a reliable estimation of the camera pose in an environment and perform the subsequent 3D reconstruction tasks. Methods for place recognition have been tested on data recorded in different environments, each including different challenges. Indoor scenes can be subject to heavy viewpoint variations~\cite{Shotton2013}. Outdoor environments include drastic changes in illumination~\cite{Arandjelovic2017}, weather~\cite{Milford2012}, seasonal~\cite{Sunderhauf2013} as well as long-term variations~\cite{msls}.

Recently, approaches based on Convolutional Neural Networks (CNNs) achieved very good results for place recognition~\cite{zhang2020visual,masone2021survey}. Their great generalization capabilities allow to deploy them as feature extractors, yielding acceptable results~\cite{lopez2019place,babenko2015aggregating,chen2014convolutional,sunderhauf2015performance}. It is, however, with end-to-end trained methods that the best performance results were achieved. 
These methods usually optimize a triplet loss function, using triplets made of a query image, a positive and a negative match~\cite{Lopez-Antequera2017}, or a query image, a set of similar images and a set of dissimilar images~\cite{Arandjelovic2017}. Other approaches rely on a contrastive loss function~\cite{radenovic2018fine,leyvavallina2019caip, liu2019stochastic}, and are trained using as input similar or dissimilar image pairs. 

All these methods are built upon considering image similarity as a binary option, i.e. two images either depict the same scene/object or not. In practice, image similarity is not a binary attribute, but rather a continuous one. A pair of images may be 100\% similar if they are the same, or not similar at all, if they do not share features. In between, one should consider a range where two images share some features, while not being exactly the same. We illustrate some examples in Fig.~\ref{fig:msls_similarity_example}. \revised{Several} existing methods~\cite{Arandjelovic2017,radenovic2018fine,yu2019spatial} are trained using binary similarity labels, ignoring the range of partial similarity that exists in practice. 

Since deep learning methods rely on the amount and quality of the training data, noise or errors in the labels may cause the models to not be effectively trained, lowering their performance. \revised{We } more complete and precise label information can contribute to increasing the performance of existing methods. This improvement can be achieved by re-annotating a dataset~\cite{weyand2020google}, by implementing a multi-task architecture~\cite{kendall2018multi,yu2020bdd100k,baslamisli2018joint,LeBMG18}, or by also using semantic~\cite{talavera2019hierarchical,zhang2020semantics,chang2020weakly,wald2020learning}, geometric~\cite{kendall2015posenet,kendall2017geometric,naha2020pose}, or sequential~\cite{Milford2012} information to train the models. 
\revised{In~\cite{balntas2018relocnet,ding2019camnet}, the authors involved scene geometry to construct more meaningful labels to learn pose regression models. They considered the 3D camera frustum overlap as a direct measure of camera pose similarity and used it as a regression target to learn image representations in a continuous setting. The formulation of the loss functions and the measure of similarity in these works are, however, strictly linked to the geometry of the scene and the visual localization task.} 

In this work, we build on the concept of partial image similarity for image retrieval, and propose a generalized formulation of the Contrastive Loss (GCL) function\revised{, to learn a suitable embedding space for image retrieval}. We deploy it as objective function to train siamese network architectures\revised{, and apply to a visual place recognition task}. The proposed GCL function enforces the representation of similar images to be closer in the latent space, while pushing apart the representations of dissimilar images proportionally to their annotated degree of similarity.

We \revised{implement} three strategies to automatically re-annotate existing datasets for place recognition with labels that indicate the degree of similarity of image pairs: Weak 2D Field-of-View overlap, Strong 2D Field-of-View overlap and 3D Field-of-View overlap. These strategies rely on different available information recorded in the original data sets.  The Weak 2D Field-of-View overlap labeling strategy exploits the GPS position and compass angle information associated to the images to estimate the 2D Field-of-View (FoV) of the camera that has taken the concerned picture. The intersection area between the FoV of two cameras determines the ground truth degree of similarity of the image pair. We use this to re-label the Mapillary Street Level Sequences dataset~\cite{msls}, which contains images recorded in urban, suburban and countryside environments in 30 cities all over the world. The Strong 2D Field-of-View overlap also computes the intersection of the FoV of two cameras, but it is calculated based on the 6DOF camera pose information.  We use it to re-annotate the TB-Places dataset, where the 6DOF ground truth camera pose was recorded using a laser tracker and an IMU sensor~\cite{leyvavallina2019access}. The 3D Field-of-View overlap strategy computes the intersection of the 3D FoV of two cameras, based on the 3D reconstruction of the scene. We apply it to re-label the 7Scenes dataset~\cite{Shotton2013}, which contains images and 3D models of indoor scenes. 

We use the proposed automatic annotations to train a siamese CNN architecture by optimizing the Generalized Contrastive Loss and we compare it to existing approaches that rely on binary image similarity ground truth. The method that we propose consists of a fully convolutional backbone with a simple pooling operation on top (i.e. Global Average or GeM pooling). We do not incorporate complex pair mining strategies into the training process of our method. We only ensure that each batch contains approximately the same amount of positive and negative pairs, each composed of two images: a query and a positive or negative counterpart.
We carry out experiments on the MSLS~\cite{msls} (for which we report new state-of-the-art results), Pittsburgh30k~\cite{Torii-PAMI2015,Arandjelovic2017}, Tokyo 24/7~\cite{Torii-CVPR2013}, \revised{RobotCar Seasons v2~\cite{sattler2018benchmarking, Maddern2017}, Extended CMU Seasons~\cite{sattler2018benchmarking, Badino2011}}, TB-Places~\cite{leyvavallina2019access} and 7Scenes~\cite{Shotton2013} benchmark data sets.

The contributions of this work are \revised{two}-fold: 
\begin{enumerate}
    \item a novel Generalized Contrastive Loss (GCL) function, which we use to train siamese networks using graded image similarity annotations \revised{ and a na\"ive pair mining strategy that does not involve memory-intensive computations on the GPU;}
    \item new state-of-the-art results on the MSLS data set.
\end{enumerate}
\noindent \revised{
The na\"ive pair mining strategy uses only the graded similarity associated to image pairs as in~\cite{ding2019camnet}, 
and ensures that a training batch contains pairs with a roughly uniform distribution of graded similarity.  With respect to~\cite{ding2019camnet}, in which three matches are selected for each anchor (easy, moderate and hard sample on the basis of the camera frustum overlap), we simplify the sampling by selecting only one match per query based on the annotated similarity. Our strategy does not select hard-pairs by computing distances in a latent-space like commonly used computationally-intensive mining strategies~\cite{msls,kim2019deep,Arandjelovic2017,liu2019stochastic}.
}
\revised{Additionally, we introduce three automatic data annotation techniques, based on GPS coordinates and compass angle, 6DOF camera pose and 3D reconstruction information and deploy them to reannotate the MSLS, TB-Places and 7Scenes datasets. We make the new labels available. }

The paper is organized as follows. We discuss related works in Section~\ref{sec:previouswork}. We introduce the Generalized Contrastive Loss function, the model architecture and mining strategy in Section~\ref{sec:methodology}, and provide details about our automatic labeling techniques in Section~\ref{sec:data}. We explain the experimental framework and evaluation procedure in Section~\ref{sec:experiments}. We present and discuss the results that we achieved in Section~\ref{sec:results}. Finally, we draw conclusions in Section~\ref{sec:conclusions}.

%% file: sections/2_previouswork.tex
\label{sec:previouswork}
\subsection{\revised{Metric learning for p}lace recognition}
Traditional place recognition algorithms rely on local image features and holistic representations, such as Fisher Vectors~\cite{perronnin2010large,jegou2010fisher}, Bag of Words~\cite{bow2012,Torii-PAMI2015,philbin2007object} and VLAD~\cite{jegou2011VLAD,arandjelovic2013all}, or on exploiting image sequences~\cite{Milford2012,Sunderhauf2013} or panoramas~\cite{Torii-PAMI2015}.

Deep learning methods achieved state-of-the-art results~\cite{zhang2020visual,masone2021survey}, initially using pre-trained CNNs as feature extractors~\cite{lopez2019place, babenko2015aggregating,chen2014convolutional,Sunderhauf2015,sharif2014cnn}, and subsequently as end-to-end trainable models.  Several methods optimize a contrastive loss function~\cite{radenovic2018fine}, with pairs of positive and negative samples. In~\cite{thoma2020soft}, the authors demonstrate the benefits of using a ground truth based on soft assignments to positive and negative classes, \revised{based on the Euclidean distance between the annotated GPS coordinates of images in their dataset, without considering orientation. This leads to some discrepancies in the soft assignment, e.g. two images taken in very close places but facing opposite directions do not share visual cues. It makes the use of a hard mining strategy necessary for convergence.}  \revised{In~\cite{kim2021embedding}, the authors propose a Contrastive Loss function that weights similar pairs of images and dissimilar images on the basis of the distance of their representations in the embedding space. As the weighing depends on the distance in the embedding space, it requires extra regularization during training. In~\cite{balntas2018relocnet}, the authors proposed to regress the 3D camera frustum overlap as part of an image retrieval pipeline for image localization, learning image retrieval and pose regression simultaneously. The authors of ~\cite{ding2019camnet} build further into the idea of frustum overlap based retrieval, learning a relocalization pipeline with two modules for coarse and fine localization.}
Other methods are trained by optimizing a triplet loss function, with image tuples consisting of a query, a positive match and a negative match ~\cite{Lopez-Antequera2017}. NetVLAD~\cite{Arandjelovic2017} is trained using a triplet loss function with a query, a set of potential positive matches and a set of definite negative matches. They optimize a weak triplet loss function that enforces higher distances between the query and all the negative matches than between the query and any potential positive. PointNetVLAD~\cite{angelina2018pointnetvlad} exploits 3D information by combining PointNet~\cite{qi2017pointnet} and NetVLAD for large-scale place recognition. In~\cite{yu2019spatial}, the authors use spatial pyramid pooling to encode spatial and structural information into the NetVLAD descriptors, and a weighted triplet loss to maximize the distances between the negative pairs and maximize that between the positive pairs. \revised{In~\cite{liu2019stochastic}, the authors train a NetVLAD by optimizing a SARE (Stochastic Attraction-Repulsion Embedding) loss function, which aims to minimize the distance among similar scences while maximizing the distance among dissimilar scenes in a probabilistic framework. The use of the SARE loss function required a reformulation of hard-pair mining, which is also necessary for the training process to converge. In~\cite{kim2019deep}, a measure of the semantic degree of similarity between object pairs was incorporated into the learning process to optimize a novel log-ratio triplet loss, and applied to human pose, room layout, and caption-aware image retrieval. This approach, however, required the reformulation of the triplet mining strategy to perform dense sampling in the neighborhood of the anchor sample.
In~\cite{revaud2019learning}, the global retrieval ranking is optimized, rather than a distance loss function for image retrieval, which was used for visual localization systems~\cite{benchmarking_ir3DV2020}. In~\cite{hausler2021patch}, the authors use the patch-level features of VLAD layer to do local feature matching and re-rank the retrieval predictions, obtaining an improvement in the recall.}

Several challenges of visual place recognition in different environments were studied and several benchmark datasets were publicly released. The KITTI~\cite{Geiger2013}, Oxford RobotCar~\cite{Maddern2017}, \revised{RobotCar Seasons~\cite{sattler2018benchmarking},} Pittsburgh250K~\cite{Torii-CVPR2013} and TokyoTM~\cite{Arandjelovic2017} datasets contain images taken in urban environments. \revised{The CMU dataset~\cite{Badino2011} and the Extended CMU Seasons dataset~\cite{sattler2018benchmarking} contain images taken in urban, suburban and park environments.} Tokyo 24/7~\cite{Arandjelovic2017} and Aachen~\cite{Sattler2012} also provide images taken at day and night that depict drastic changes in illumination conditions. The Alderley dataset also includes weather variations~\cite{Milford2012}. The Mapillary Street Level Sequences dataset includes images taken with different cameras in 30 cities in six continents~\cite{msls}. It contains changes in viewpoint, weather, illumination and long-term variations, as well as urban, suburban and countryside environments. To the best of our knowledge, it is the largest dataset for long-term place recognition. The Norland dataset includes images taken in countryside and contains variations of weather conditions and season~\cite{Sunderhauf2013}. The TB-Places dataset was designed for place recognition in gardens~\cite{leyvavallina2019access,leyvavallina2019caip}. It includes challenging long-term variation, and viewpoint changes, as well as very repetitive textures and scenes dominated by objects of green color appearance.
\subsection{Data annotation}

CNNs, which constitute the current state-of-the-art in most computer vision tasks, are trained on large-scale annotated datasets, such as ImageNet~\cite{deng2009imagenet} and Places~\cite{zhou2017places} for object and scene classification, COCO~\cite{lin2014microsoft} and Pascal VOC~\cite{Everingham15} for object detection, and Pittsburgh250k~\cite{Torii-CVPR2013} and Mapillary Street Level Sequences~\cite{msls} for visual place recognition. The quality, diversity, and richness of the ground truth annotations have a direct impact on the performance of the trained models and contribute to the strength of their generalization capabilities~\cite{sun2017revisiting}. For instance, in~\cite{biasedai} it was shown that models trained using inaccurate ground truth data can lead to gender and racial biased predictions, as they fail to learn the actual distribution of the real-world problem. 

Thus, a correct, informative and complete definition of the ground truth labels is of utmost importance to learn more accurate and unbiased models. Previous works on data curation include dataset expansion, e.g. Places2~\cite{zhou2017places}, and manual re-annotations, e.g. Google Landmarks~\cite{weyand2020google}. 
Another trend in the literature is Multi-Task learning, which consists of learning two (or more) related but yet diverse tasks simultaneously, benefiting from their commonalities. This generally results in higher prediction accuracy for the individual tasks, compared to learning them separately~\cite{caruana1997multitask}. Recent works include the combination of semantic and geometric tasks for pose regression~\cite{kendall2018multi} or joint learning of semantic segmentation and image intrinsic decomposition~\cite{baslamisli2018joint}. Although its efficacy, Multi-Task learning is limited by the scarcity of adequate datasets, which are required to provide samples with multiple labels, for each task.
Another option is to incorporate available complementary information in the training process, such as for SeqSLAM~\cite{Milford2012,Sunderhauf2013} where image sequences were used for Simultaneous Localization and Mapping. 

%% file: sections/3_methodology.tex
\label{sec:methodology}
In this section, we propose the Generalized Contrastive Loss function and show how to use it to train a siamese network architecture. Furthermore, we describe the place recognition pipeline and how we apply it to search and retrieve similar images from a reference map.
\subsection{Fully convolutional backbone and pooling}

\begin{figure}[t!]
    \centering
    {\fontsize{7pt}{11pt}
    \def\svgwidth{\columnwidth}
      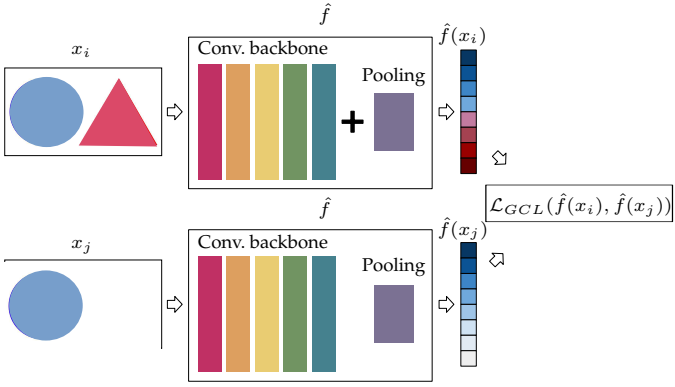}
    \caption{Sketch of a siamese architecture where $x_i$ and $x_j$ are the input images, $\hat{f}$ represents the convolutional backbone with a pooling layer and $\hat{f}(x_i)$ and $\hat{f}(x_j)$ are the representations of the input images. They are used as input for the Generalized Contrastive Loss function $\mathcal{L}_{GCL}(\hat{f}(x_i), \hat{f}(x_j))$.}
    \label{fig:siamese}
\end{figure}

We deploy a siamese architecture with a fully convolutional backbone. Given an input image $x \in R^{w_n \times h_n \times d_n}$, we consider a convolutional network that computes a representation $f(x) \in R^{w_m \times h_m \times d_m}$, where $w_m$ and $h_m$ are the width and height of the last convolutional activation map, and $d_m$ corresponds to the number of kernels of the last convolutional layer. The output tensor of the last convolutional layer of the network is fed to a global pooling layer, which computes the image representation $\hat{f}(x)\in R^{d_m}$. In this work, we experiment with versions of the network with a Global Average Pooling and a GeM pooling layer~\cite{radenovic2018fine}. However, one can explore the use of other global pooling strategies, in accordance with the application at hand. Given a pair of images $(x_i, x_j)$, their representation is computed as $(\hat{f}(x_i), \hat{f}(x_j))$. We illustrate the architecture in Fig.~\ref{fig:siamese}.
Our methodology can be applied using any convolutional architecture as backbone: in this work we use DenseNet~\cite{huang2017densely}, ResNet~\cite{he2016deep}\revised{, VGG16~\cite{simonyan2015very} and ResNeXt~\cite{xie2017aggregated}}.

\subsection{Generalized Contrastive Loss}

Siamese architectures are used to learn representations that disentangle the (dis-)similarity of pairs of input images. They consist of two identical networks (i.e. they share their weights) that process input image pairs. Applications include signature verification~\cite{bromley1994signature}, face identification~\cite{chopra2005learning}, visual place recognition~\cite{Arandjelovic2017,Lopez-Antequera2017,leyvavallina2019caip} and image retrieval~\cite{radenovic2018fine}. The training of a siamese architecture is generally carried out by optimizing a Contrastive Loss function.

\subsubsection{Contrastive Loss function}
Let us consider two input images $x_i$ and $x_j$, and their representations $\hat{f}(x_i)$ and $\hat{f}(x_j)$. We define the distance between the representations of the input images $x_i$ and $x_j$ in the latent space as $d(x_i, x_j) = \left \| \hat{f}(x_i)-\hat{f}(x_j) \right \|_2$.
The Contrastive Loss~\cite{hadsell2006dimensionality} function $\mathcal{L}_{CL}$ is defined as:
\begin{equation}
 \mathcal{L}_{CL}(x_i,x_j)=\begin{cases}
     \frac{1}{2}d(x_i, x_j)^2 ,& \text{if } y= 1\\
    \frac{1}{2}\max(\tau-d(x_i, x_j),0)^2,& \text{if } y=0
\end{cases}
\label{eq:contrastive}
\end{equation}
where $\tau$ is the margin, i.e. a threshold for the descriptor distance above which a pair of images is not considered as depicting the same place. The margin $\tau$ is an hyperparameter defined by the user. The ground truth label $y$ is such that $1$ indicates a pair of similar images, and $0$ a not-similar pair of images.  Similarity, however, is not a binary attribute, and defining it as such may cause the trained models to produce unreliable predictions. 

\begin{figure}[t!]
    \centering
\subfloat[A]{
    \fbox{\includegraphics[scale=0.075]{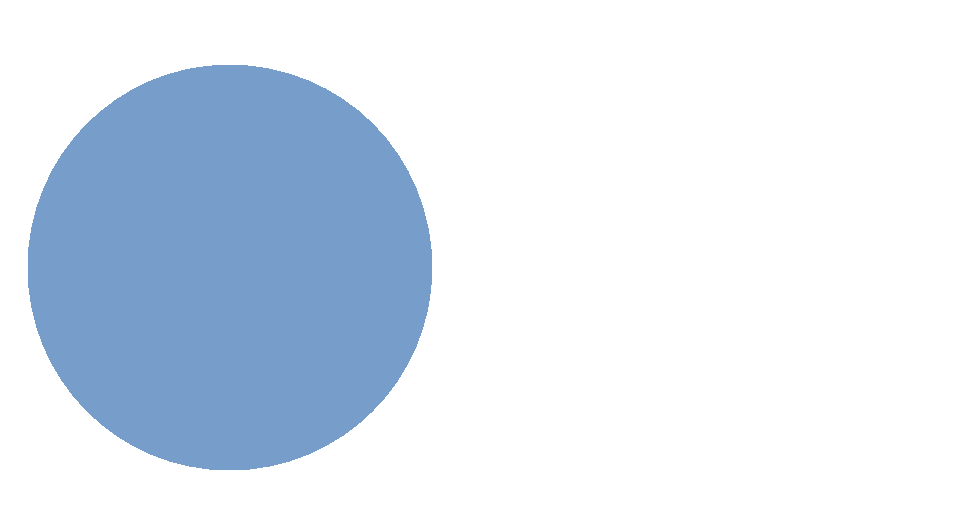}}
    \label{fig:image_a}
    }
  \subfloat[B]{
    \fbox{\includegraphics[scale=0.075]{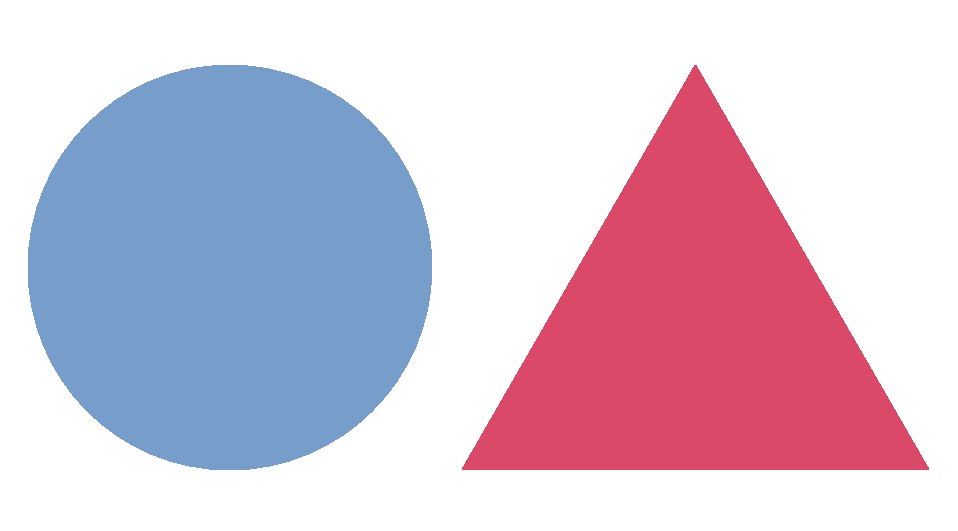}}
    \label{fig:image_b}
    }
    \subfloat[C]{
    \fbox{\includegraphics[scale=0.075]{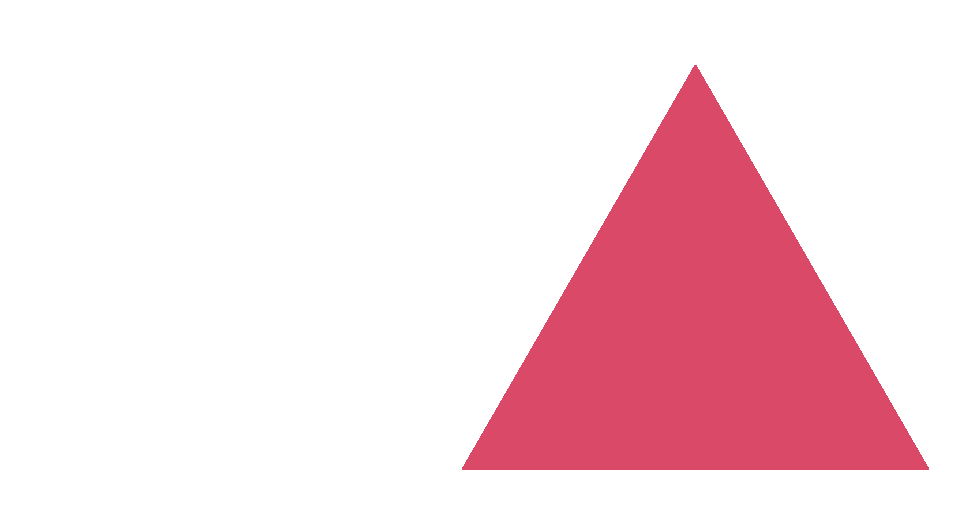}}
    \label{fig:image_c}
    }

    \caption{Example dataset consisting of three images. ${(A,B)}$ and ${(B,C)}$ are considered similar ($y=1$) because they share part of their content, while ${(A,C)}$ are dissimilar ($y=0$).}
    \label{fig:toy_dataset}
\end{figure}

Let us consider an example dataset consisting of three images, as depicted in Fig.~\ref{fig:toy_dataset}. The pairs ${(A,B)}$ and ${(B,C)}$ are labeled as similar, while the pair ${(A,C)}$ is labeled as not-similar. The aim of a siamese network is to learn a function $\hat{f}$ that maps similar images to points in a latent space that are close together, and non-similar image pairs to points with larger distance in the latent space.

The optimization of the Contrastive Loss function aims at minimizing the distance between the representations of the similar image pairs ${(A,B)}$ and ${(B,C)}$, such that:
\begin{equation}
\nonumber
\left \| \hat{f}(A)-\hat{f}(B) \right \|_2 \approx 0 \,\land\,  \left \| \hat{f}(B)-\hat{f}(C) \right \|_2 \approx 0
\end{equation}

\noindent that corresponds to:
\begin{equation}
\nonumber
\hat{f}(A)\approx\hat{f}(B)
\,\land\, 
\hat{f}(B)\approx\hat{f}(C)
\end{equation}

Therefore, by the transitive property of equality, the Euclidean distance of the representation of $A$ and $C$ is also ensured to be close in the latent space:
\begin{equation}
 \left [  \left ( \hat{f}(A) \approx  \hat{f}(B) \right )\land \left (\hat{f}(B)\approx  \hat{f}(C) \right )\right ] \rightarrow  \hat{f}(C)\approx  \hat{f}(A)
 \label{eq:dissimilar}
 \end{equation}
In the example we considered, however, ${(A,C)}$ is labeled as dissimilar. The result of Eq.~\ref{eq:dissimilar} is in contrast with the optimization ensured by Eq.~\ref{eq:contrastive}, where the distance of the representation of dissimilar images is maximized.
This inconsistency in the training process is due to the fact that the Contrastive Loss function does not take into account the partial similarity between the input pairs.

\subsubsection{Generalized Contrastive Loss function}
We propose a generalized formulation of the Contrastive Loss that relies on a definition of continuous, rather than binary, similarity. We define the Generalized Contrastive Loss function $ \mathcal{L}_{GCL}$ as:

\begin{align}
    \begin{split} 
 \mathcal{L}_{GCL}(x_i,x_j)= {}&\psi_{i,j}\cdot \frac{1}{2}d(x_i, x_j)^2 +\\  & (1-\psi_{i,j}) \cdot \frac{1}{2}\max(\tau-d(x_i, x_j),0)^2
\end{split}
\label{eq:generalized_contrastive_loss}
\end{align}
where $x_i$ and $x_j$ denote the two input images, $\hat{f}(x)$ is the representation of the input $x$ and $\psi_{i,j} \in \left[0, 1 \right]$ is the ground truth degree of similarity of $x_i$ and $x_j$. \revised{In contrast to Eq.~\ref{eq:contrastive}, here the similarity $\psi_{i,j}$ is} a continuous value ranging from 0 (completely dissimilar) to 1 (identical). 
By minimising a Generalized Contrastive Loss function, the distance of each image pair in the latent space is optimized 
proportionally to the corresponding degree of similarity. 

\subsubsection{Gradient of the Generalized Contrastive Loss}
In the training phase, the loss function is minimized by gradient descent optimization and the weights of the network are updated by backpropagation. In the case of the Constrastive Loss function, the gradient is:
\begin{equation}
     \nabla  \mathcal{L}_{CL}(x_i,x_j)=\begin{cases}
     d(x_i, x_j) ,& \text{if } y= 1\\
    \min(d(x_i, x_j)-\tau,0),& \text{if } y=0
\end{cases}
\label{eq:contrastive_loss_grad}
\end{equation}
It is worth noting that the $\min$ function ensures that the gradient is computed for the negative pairs whose distance in the latent space is lower than the margin $\tau$ (see the supplementary materials for the derivation of the gradient).
The weights of the model are updated in order to minimize the distance between similar images, and to maximize the distance between dissimilar ones. A partial similarity is therefore not considered, and the training process may incur in learning inconsistent representations.

In contrast, the proposed Generalized Constrastive loss function takes into account the degree of similarity between the input images during the training process. Its gradient is:

\begin{equation}
     \nabla  \mathcal{L}_{GCL}(x_i,\!x_j)\!=\!\begin{cases}d(x_i,\!x_j)\!+\!\tau(\psi_{i,j}\!-\!1),& \text{if } d(x_i,\!x_j)\!<\!\tau\\
     d(x_i, x_j) \cdot \psi_{i,j},& \text{if } d(x_i,\!x_j)\!\geq\!\tau\\
     \end{cases}
     \label{eq:generalized_contrastive_loss_grad}
\end{equation} 

\noindent It is worth pointing out that the magnitude of the gradient of the Generalized Contrastive Loss function is modulated by the ground truth degree of similarity of the input image pairs, $\psi_{i,j}$. In the case the distance $d(x_i,x_j)$ is larger than the margin $\tau$, the weights of the network are updated proportionally to $\psi_{i,j}$. 
More interestingly, when the distance $d(x_i,x_j) < \tau$, the Generalized Contrastive Loss function has an intrinsic regularization effect of the learned latent space.
The network weights are indeed updated so that the vector representations of the training images $x_i$ and $x_j$ are moved closer in the latent space if their ground truth degree of similarity is $\psi_{i,j} > 1-\frac{d(x_i,x_j)}{\tau}$, otherwise they are pushed away. 
For the boundary cases $\psi_{i,j}=0$ (completely dissimilar input images) and $\psi_{i,j}=1$ (same exact input images), the gradient is the same as in Eq.~\ref{eq:contrastive_loss_grad}.

\subsection{Image search and place recognition}
Let us consider a set $X$ of map images, for which the location
in the environment of the camera that took them is known. We are presented with a set $Y$ of query images taken from unknown positions. In order to effectively localize the camera that took the query images in the environment, we are required to retrieve similar images to the query from the map set.
We compute a representation of the map images $\hat{f}(x)  \,\forall x \in X$, and of the query images $\hat{f}(y)\, \forall y \in Y$ using our network. For a given query representation $\hat{f}(y)$, image retrieval is performed by an exhaustive nearest neighbor search among the representations of the map set $\hat{f}(x)\, \forall x \in X$, retrieving the map images with the closest representations in the latent space.

It is worth pointing out that in this paper we do not address the camera localization and pose estimation problems. We focus on training networks that are able to compute an effective representation for better retrieval results, which can further benefit the localization task.

%% file: figures/3_methodology/Siamese_tex.pdf_tex
\begingroup%
  \makeatletter%
  \providecommand\color[2][]{%
    \errmessage{(Inkscape) Color is used for the text in Inkscape, but the package 'color.sty' is not loaded}%
    \renewcommand\color[2][]{}%
  }%
  \providecommand\transparent[1]{%
    \errmessage{(Inkscape) Transparency is used (non-zero) for the text in Inkscape, but the package 'transparent.sty' is not loaded}%
    \renewcommand\transparent[1]{}%
  }%
  \providecommand\rotatebox[2]{#2}%
  \newcommand*\fsize{\dimexpr\f@size pt\relax}%
  \newcommand*\lineheight[1]{\fontsize{\fsize}{#1\fsize}\selectfont}%
  \ifx\svgwidth\undefined%
    \setlength{\unitlength}{669bp}%
    \ifx\svgscale\undefined%
      \relax%
    \else%
      \setlength{\unitlength}{\unitlength * \real{\svgscale}}%
    \fi%
  \else%
    \setlength{\unitlength}{\svgwidth}%
  \fi%
  \global\let\svgwidth\undefined%
  \global\let\svgscale\undefined%
  \makeatother%
  \begin{picture}(1,0.60426009)%
    \lineheight{1}%
    \setlength\tabcolsep{0pt}%
    \put(0,0){\includegraphics[width=\unitlength,page=1]{figures/3_methodology/Siamese_tex.pdf}}%
    
    \put(0.29,0.49){\color[rgb]{0,0,0}\makebox(0,0)[lt]{\lineheight{1.25}\smash{\begin{tabular}[t]{l}Conv. backbone\end{tabular}}}}%

    \put(0,0){\includegraphics[width=\unitlength,page=2]{figures/3_methodology/Siamese_tex.pdf}}%
    
    \put(0.535,0.45){\color[rgb]{0,0,0}\makebox(0,0)[lt]{\lineheight{1.25}\smash{\begin{tabular}[t]{l}Pooling\end{tabular}}}}%
    
    \put(0,0){\includegraphics[width=\unitlength,page=3]{figures/3_methodology/Siamese_tex.pdf}}%
    
    \put(0.29,0.2){\color[rgb]{0,0,0}\makebox(0,0)[lt]{\lineheight{1.25}\smash{\begin{tabular}[t]{l}Conv. backbone\end{tabular}}}}%
    
    \put(0,0){\includegraphics[width=\unitlength,page=4]{figures/3_methodology/Siamese_tex.pdf}}%
    
    \put(0.535,0.16492191){\color[rgb]{0,0,0}\makebox(0,0)[lt]{\lineheight{1.25}\smash{\begin{tabular}[t]{l}Pooling\end{tabular}}}}%
    
    \put(0,0){\includegraphics[width=\unitlength,page=5]{figures/3_methodology/Siamese_tex.pdf}}%
    
    \put(0.1,0.49){\makebox(0,0)[lt]{\lineheight{1.25}\smash{\begin{tabular}[t]{l}$x_i$\end{tabular}}}}%
    
    \put(0.1,0.2){\makebox(0,0)[lt]{\lineheight{1.25}\smash{\begin{tabular}[t]{l}$x_j$\end{tabular}}}}%
    
    \put(0.47,0.54){\makebox(0,0)[lt]{\lineheight{1.25}\smash{\begin{tabular}[t]{l}$\hat{f}$\end{tabular}}}}%
    
    \put(0.47,0.25){\makebox(0,0)[lt]{\lineheight{1.25}\smash{\begin{tabular}[t]{l}$\hat{f}$\end{tabular}}}}%
    
    \put(0.648,0.51){\makebox(0,0)[lt]{\lineheight{1.25}\smash{\begin{tabular}[t]{l}$\hat{f}(x_i)$\end{tabular}}}}%
    
    \put(0.648,0.22){\makebox(0,0)[lt]{\lineheight{1.25}\smash{\begin{tabular}[t]{l}$\hat{f}(x_j)$\end{tabular}}}}%
    
    \put(0.725,0.255){\makebox(0,0)[lt]{\lineheight{1.25}\smash{\begin{tabular}[t]{l}$\mathcal{L}_{GCL}(\hat{f}(x_i), \hat{f}(x_j))$\end{tabular}}}}%
    
    \put(0,0){\includegraphics[width=\unitlength,page=6]{figures/3_methodology/Siamese_tex.pdf}}%
  \end{picture}%
\endgroup%

%% file: sections/4_data.tex
\label{sec:data}


The optimization of the Generalized Contrastive Loss function relies on the ground truth similarity of image pairs defined in the range $[0,1]$. In this section we present three approaches to automatically label the degree of similarity of image pairs, which we used depending on the data available together with the images in the concerned datasets. We \revised{describe} two measures of image similarity based on the 2D Field-of-View overlap: weakly labeled, relying on GPS data, and strongly labeled, relying on the 6DOF camera pose information associated to the images. Additionally, we used the 3D Field-of-View overlap, estimated from 3D reconstruction and 6DOF camera pose.

\begin{figure}[t!]
  \centering
  \parbox{\figrasterwd}{
    \parbox{.5\figrasterwd}{%
      \subfloat[]{
      \def\svgwidth{.5\columnwidth}
      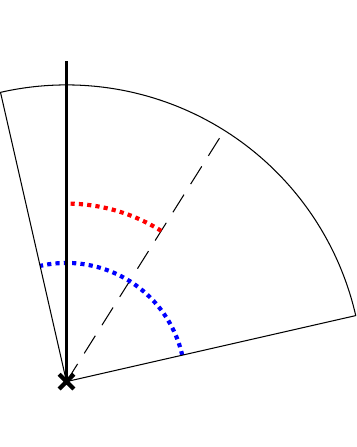
      \label{fig:fov_graph}
      }
    }
    \hspace{-3pt}
    \parbox{.2\figrasterwd}{%
      \subfloat[]{\includegraphics[width=.45\columnwidth, trim=2cm 2cm 2cm 3.25cm, clip]{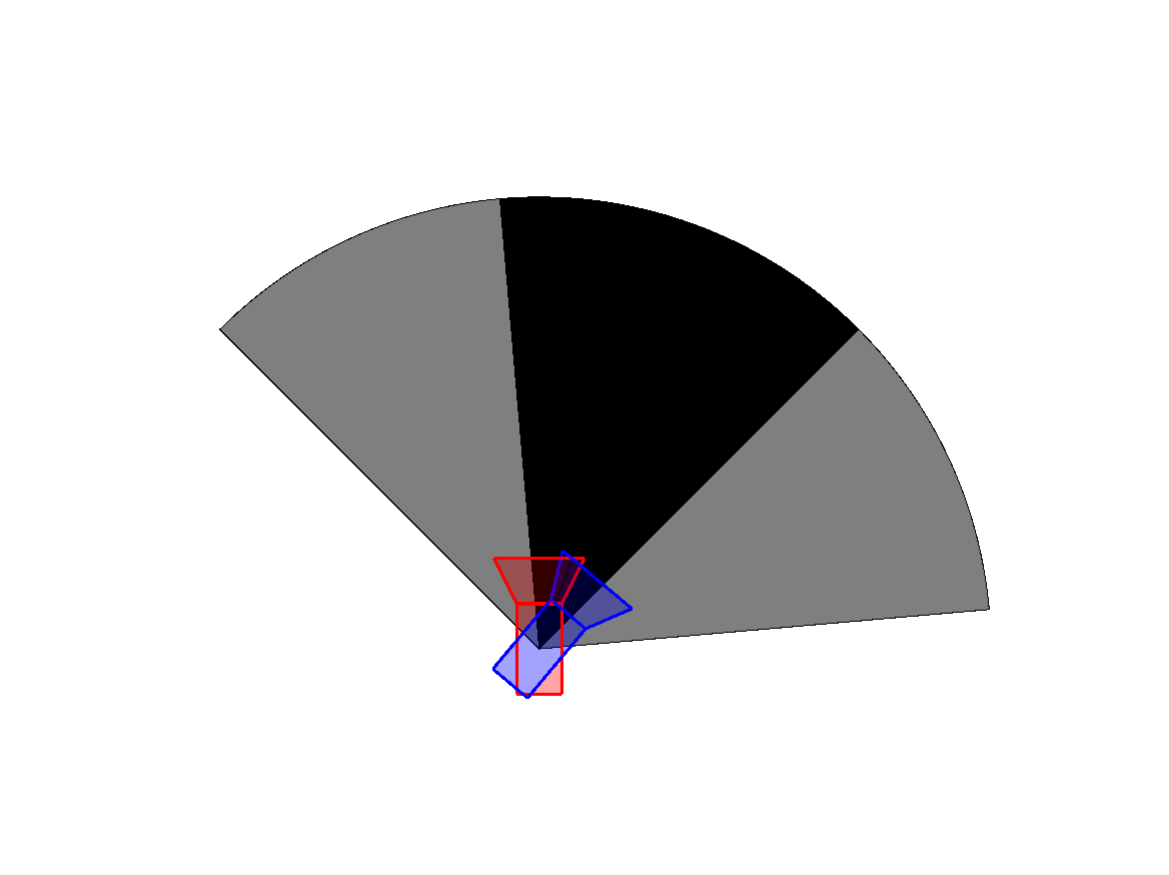}
          \label{fig:msls_0m_40deg}
          }
      \vskip1em
      \subfloat[]{\includegraphics[width=.45\columnwidth, trim=2cm 2cm 2cm 3.5cm, clip]{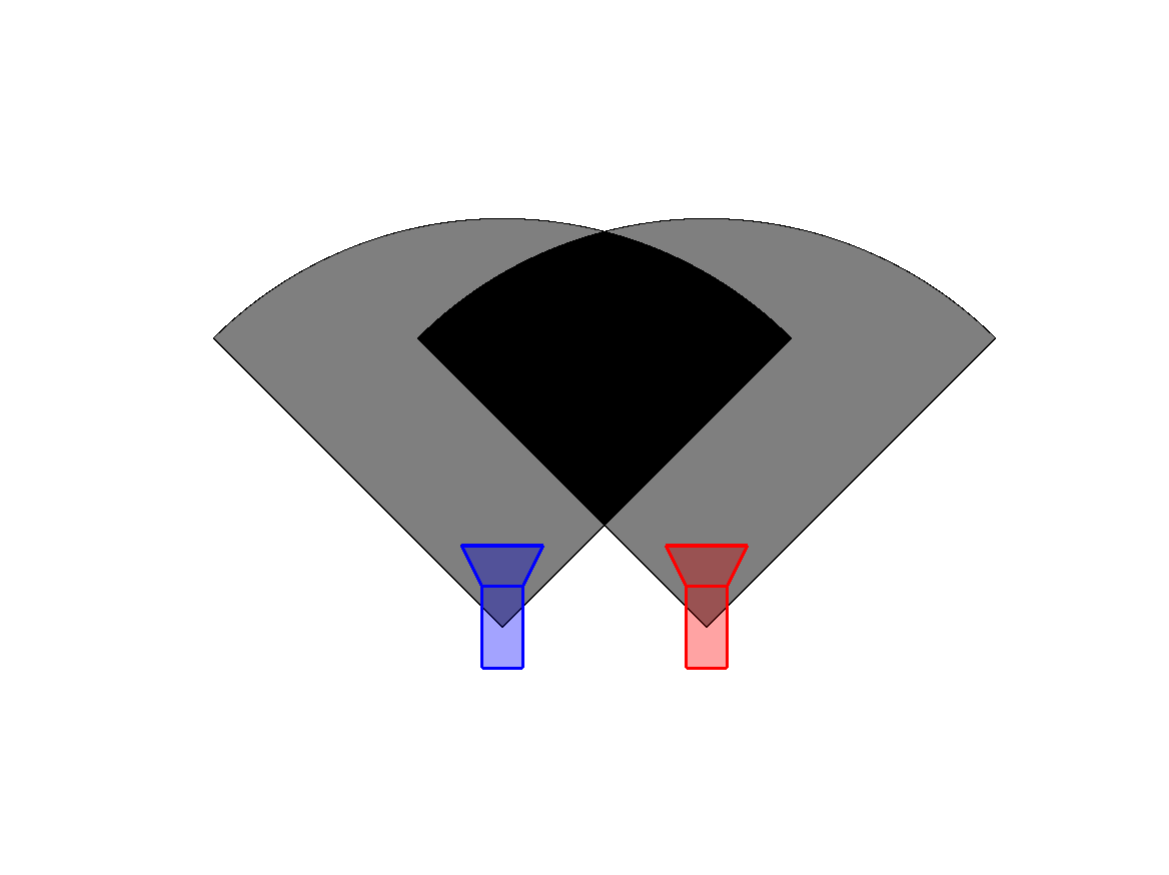}
          \label{fig:msls_25m_0deg}}
    }
    \label{fig:fov}
    \caption{(a) 2D Field-of-View representation with angle $\theta$ and radius $r$. The point $(t_0,t_1)$ is the camera location in the environment, and $\alpha$ is the camera orientation in the form of a compass angle with respect to the north $N$. A (b) soft positive match: two cameras (with  $\theta=90^{\circ}$ and $r=50m$) in the same position but with orientations $40 ^{\circ}$  apart. A (c) soft negative example: two cameras (with  $\theta=90^{\circ}$ and $r=50m$) located 25m apart but with the same orientation. }
  }

\end{figure}

\subsection{2D Field-of-View overlap}
\label{sec:2Dfov}
\begin{figure*}[t!]
    \centering
\subfloat[Query]{
    \includegraphics[width=.22\textwidth]{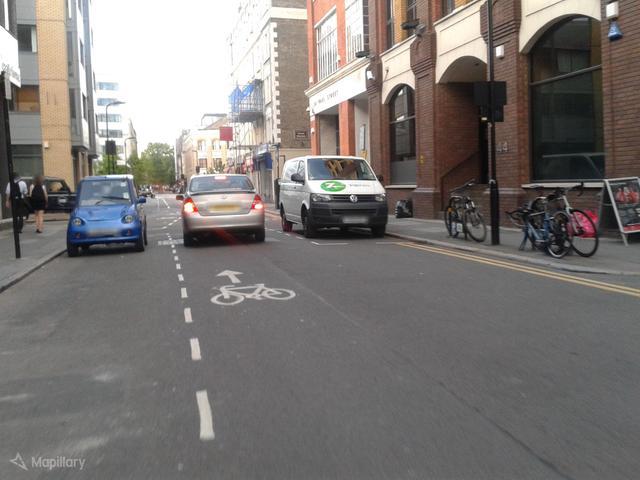}
    \label{fig:msls_london_query}
    }
     \subfloat[Query]{
    \includegraphics[width=.22\textwidth]{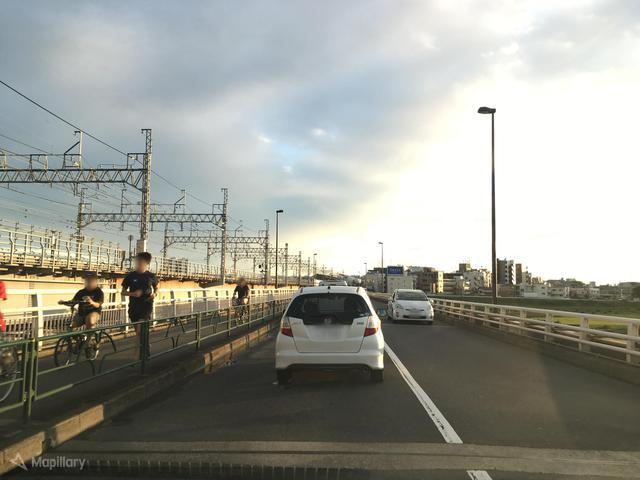}
    \label{fig:msls_tokyo_query}
    }
    \subfloat[Query]{
    \includegraphics[width=.22\textwidth]{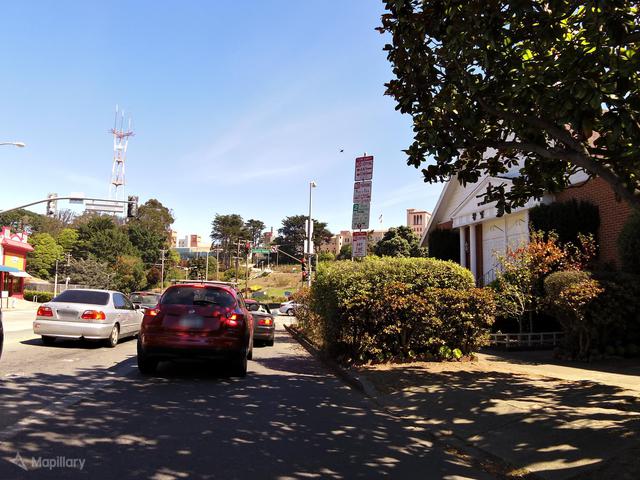}
    \label{fig:msls_sf_query}
    }
    \subfloat[Query]{
    \includegraphics[width=.22\textwidth]{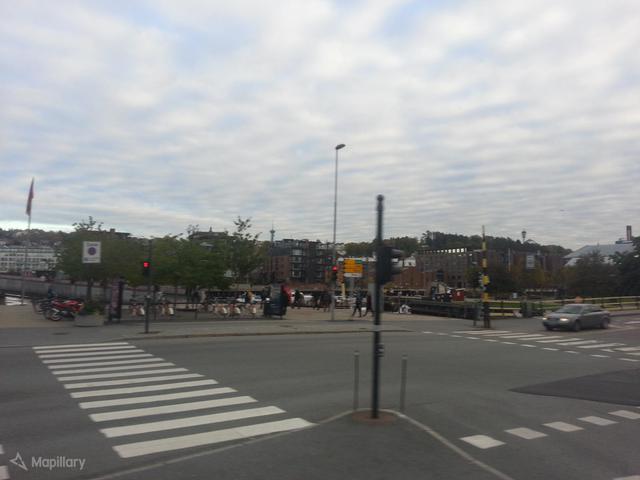}
    \label{fig:msls_trondheim_query}
    }
    
   \subfloat[Positive match]{
    \includegraphics[width=.22\textwidth]{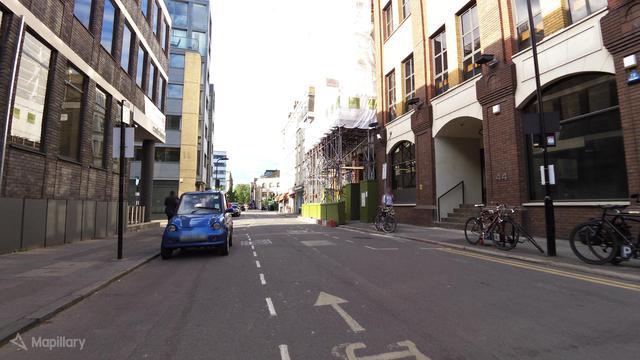}
    \label{fig:msls_london_db}
    }
    \subfloat[Positive match]{
    \includegraphics[width=.22\textwidth]{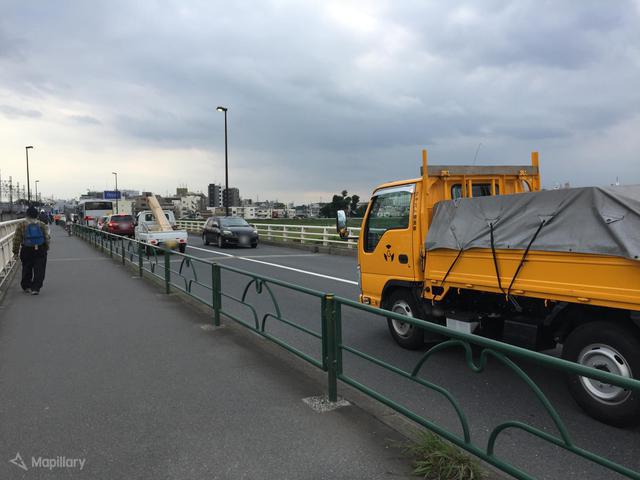}
    \label{fig:msls_tokyo_db}
    }
     \subfloat[Soft negative match]{
    \includegraphics[width=.22\textwidth]{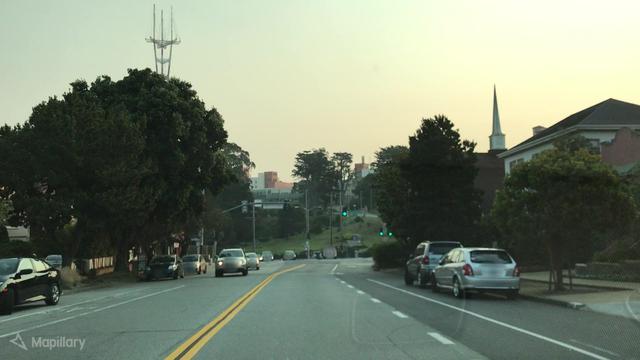}
    \label{fig:msls_sf_db}
    }
    \subfloat[Hard negative match]{
    \includegraphics[width=.22\textwidth]{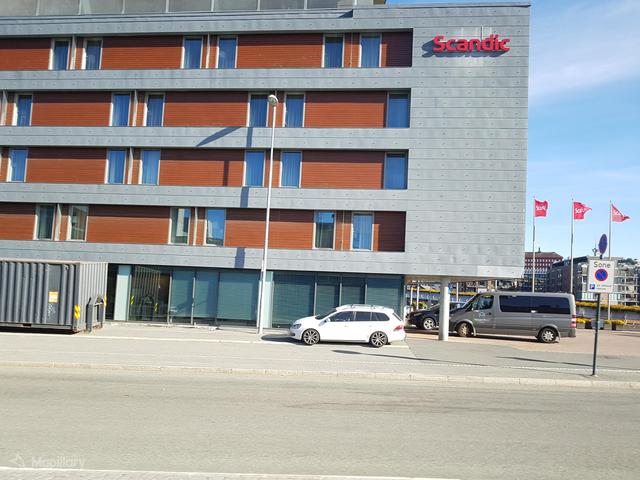}
    \label{fig:msls_trondheim_db}
    }
    
    \subfloat[75.5\% FoV overlap]{
    \includegraphics[width=.22\textwidth]{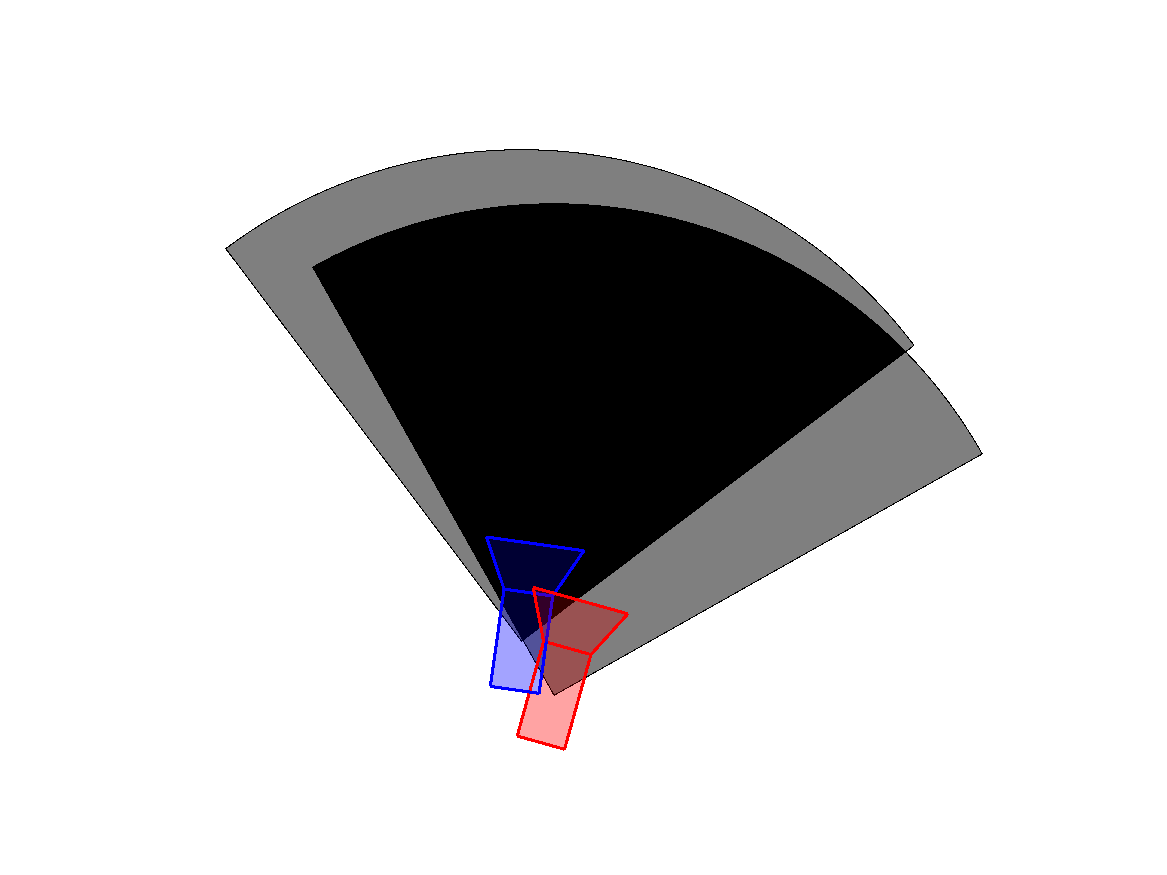}
    \label{fig:msls_london_fov}
    }
    \subfloat[Fov overlap 50.31\% ]{
    \includegraphics[width=.22\textwidth]{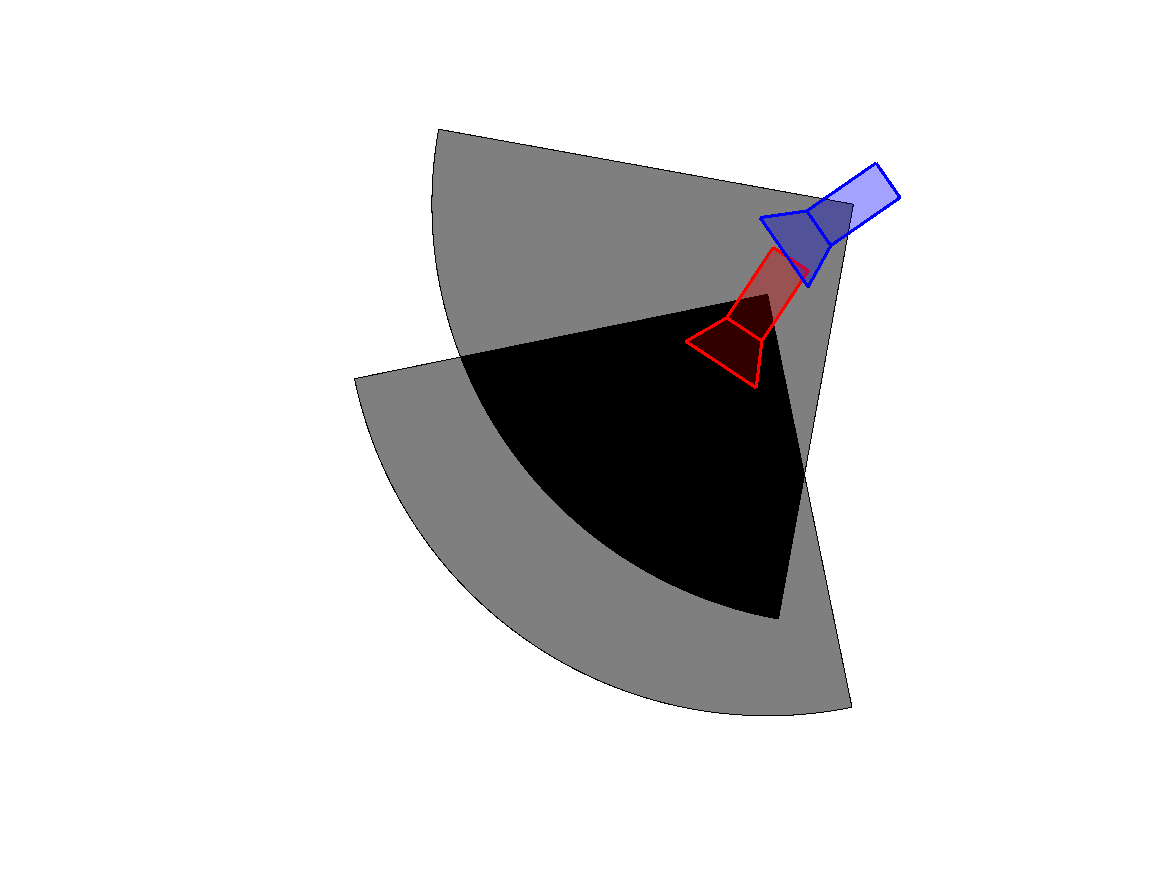}
    \label{fig:msls_tokyo_fov}
    }
    \subfloat[FoV overlap 16.78\%]{
    \includegraphics[width=.22\textwidth]{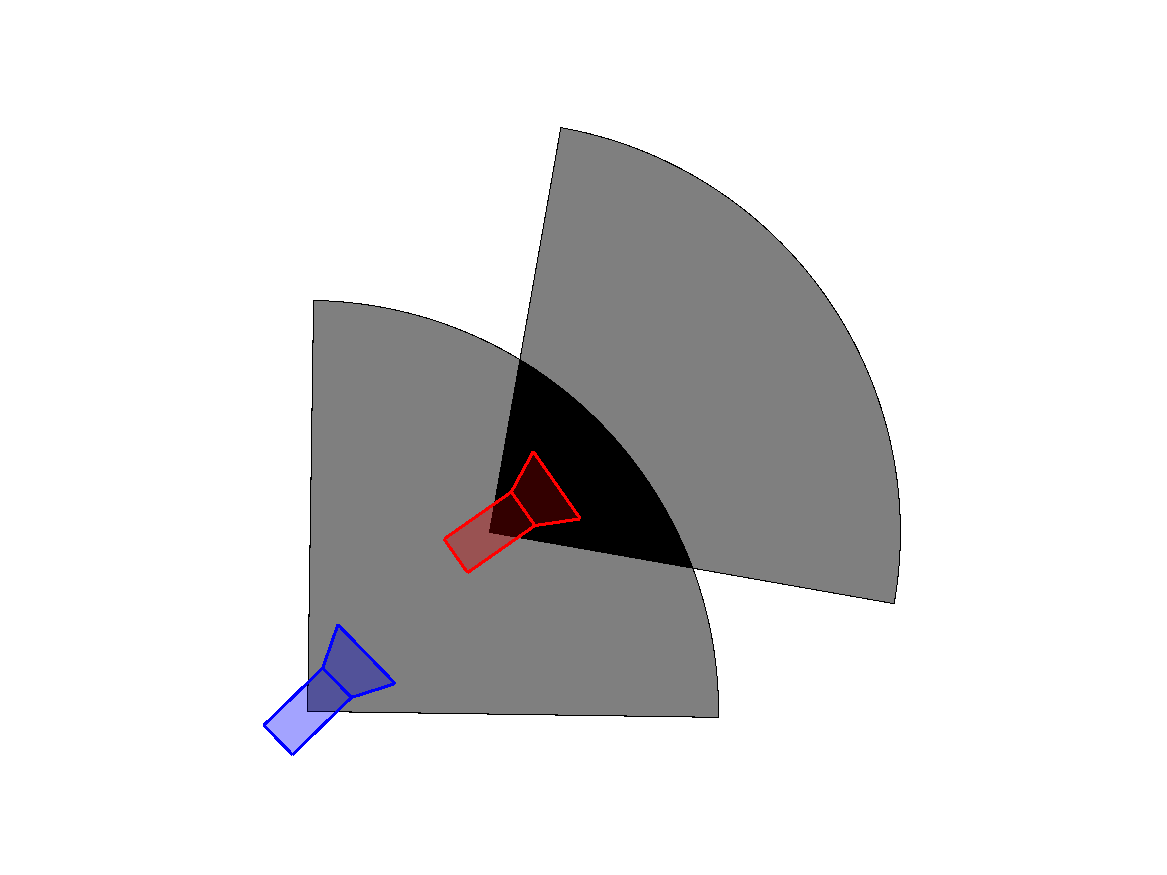}
    \label{fig:msls_sf_fov}
    }
    \subfloat[FoV overlap 0\%]{
    \includegraphics[width=.22\textwidth, trim=1cm 1cm 1cm 1cm, clip]{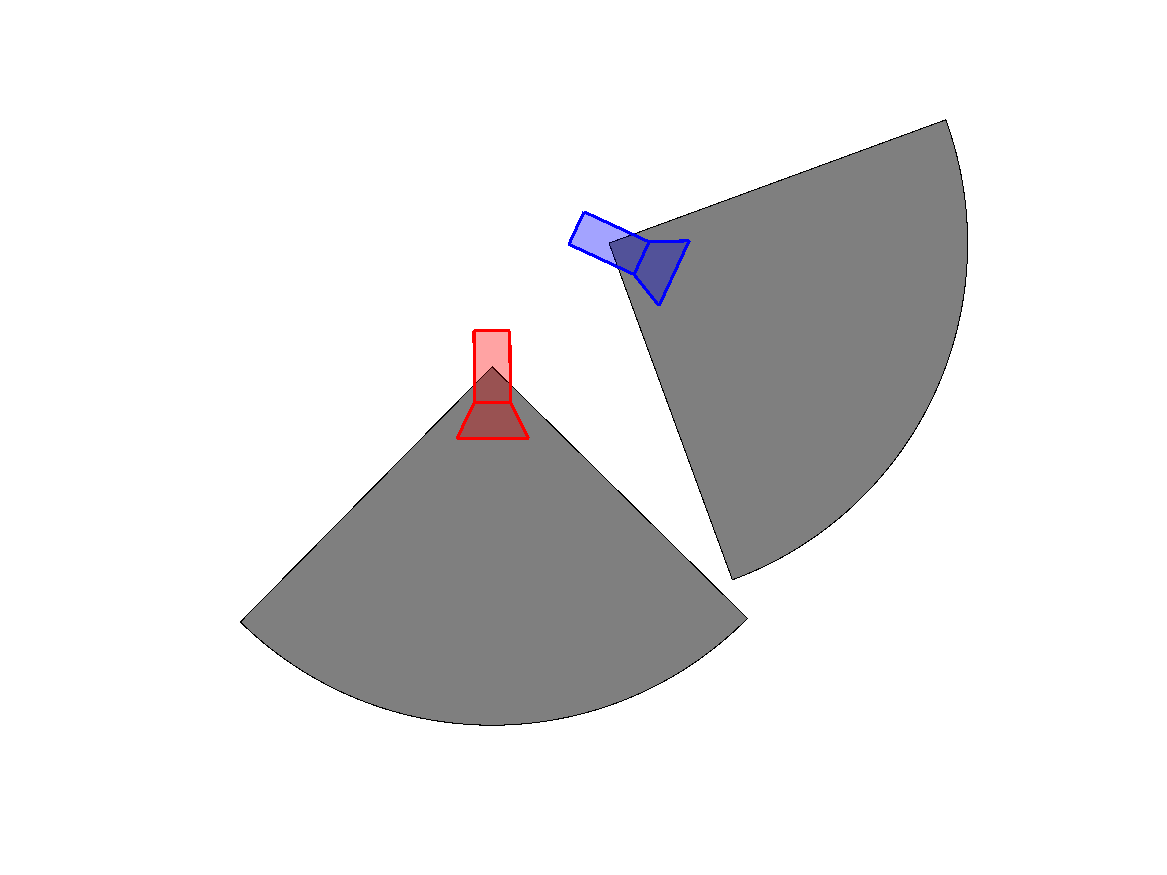}
    \label{fig:msls_trondheim_fov}
    }
   
    \caption{Example image pairs from the MSLS dataset. The first row shows the query images, the second row shows the corresponding matches from the map set, and the third row shows the estimated 2D FoV overlap. The query image is associated with the red camera, while the map image with the blue camera. The first column shows a positive match with 75.5\% FoV overlap, and many visual features in common. The second column shows a borderline pair: the two images have 50.31\% FoV overlap and some features in common. The third column shows a soft negative match, where the two images have FoV overlap of 16.78\%. The fourth column shows a hard negative match, where the two images are taken by cameras looking in opposite directions, and the FoV overlap is 0\%.}
    \label{fig:msls_fov}
\end{figure*}

We estimate the similarity of two images by approximating a measure of the overlap of their two-dimensional Field-of-View (FoV) in the horizontal plane.
Let us consider a camera with a FoV defined by the angle $\theta$ and radius $r$, positioned in an environment according to a 2D translation vector $(t_0,t_1)$ with respect to the origin of the reference system. The camera is oriented at an angle $\alpha$ with respect to the north direction of the reference system. We define the 2D FoV as the sector of the circle denoted by the center $(t_0,t_1)$ and radius $r$ enclosed in the angle range delimited by $[\alpha - \frac{\theta}{2}, \alpha +\frac{\theta}{2}]$ (see Fig.~\ref{fig:fov_graph}).
The 2D Field-of-View overlap between two images corresponds to the intersection-over-union (IoU) of their FoVs. \revised{Differently from~\cite{balntas2018relocnet,ding2019camnet} that compute the overlap of  3D camera frusta for camera localization, we relax this concept by only considering the overlap of the camera field-of-views in the horizontal plane.} We consider as positives the image pairs with determined ground truth FoV overlap higher than $50\%$, and the rest as negatives. More specifically, the negative pairs with a similarity higher than $0\%$ are soft negatives, while the pairs with a similarity degree equal to $0\%$ are hard negatives.

\subsubsection{Weak 2D Field-of-View overlap}

We \revised{present} the \emph{weak 2D Field-of-View overlap} to estimate the similarity of image pairs for which GPS position labels (UTM format) and compass angle information are available. We consider the UTM data as the translation vector $(t_0,t_1)$ and the compass angle as the orientation $\alpha$ necessary to estimate the 2D FoV of the cameras.

We use the weak 2D Field-of-View overlap to re-annotate the Mapillary Street Level Sequences (MSLS) dataset~\cite{msls}, a large-scale outdoor visual place recognition dataset. It contains images of 30 cities across 6 continents that depict urban, suburban and countryside environments, and are annotated with the UTM  position and compass angle. The FoV angle of the cameras and the intrinsics are not provided. We thus estimate the value of the FoV angle $\theta$ according to the following reasoning. The authors of MSLS define a positive match when the retrieved map image falls within 25m and $40^{\circ}$ from the query. We define a similarity measure that satisfies those constraints. Image pairs taken at locations that are closer than 25m and with orientation differences lower than $40^{\circ}$ are expected to have a similarity higher than $50\%$. Moreover, the borderline cases with distances close to 25m and/or orientation difference near to $40^{\circ}$ should have a similarity close to $50\%$. Hence, we define $r=25m\times2=50m$ and estimate a $\theta$ that gives approximately a $50\%$ FoV overlap for the borderline cases,  i.e. 0m@$40^{\circ}$, and 25m@$0^{\circ}$. For the former, the optimal $\theta$ corresponds to $80^{\circ}$, and for the second latter to $102^{\circ}$. We settle for a value in the middle and define $\theta=90^{\circ}$, which gives $55.63\%$ and $45.01\%$ FoV overlaps, as shown in Fig.~\ref{fig:msls_0m_40deg} and~\ref{fig:msls_25m_0deg}. We display  examples of the similarity ground truth for MSLS in Fig.~\ref{fig:msls_fov}. \revised{We compute the similarity ground truth for each possible query-map pair, per city.}

\subsubsection{Strong 2D Field-of-View overlap}

We \revised{present} the \emph{strong 2D Field-of-View overlap} to estimate the similarity of image pairs in the case 6DOF camera pose information is available together with the recorded images in the dataset. The translation vector $(t_0,t_1)$ and orientation angle $\alpha$ are extracted from the pose vector.

\begin{figure}
    \centering
\subfloat[Query]{
    \includegraphics[width=.3\columnwidth]{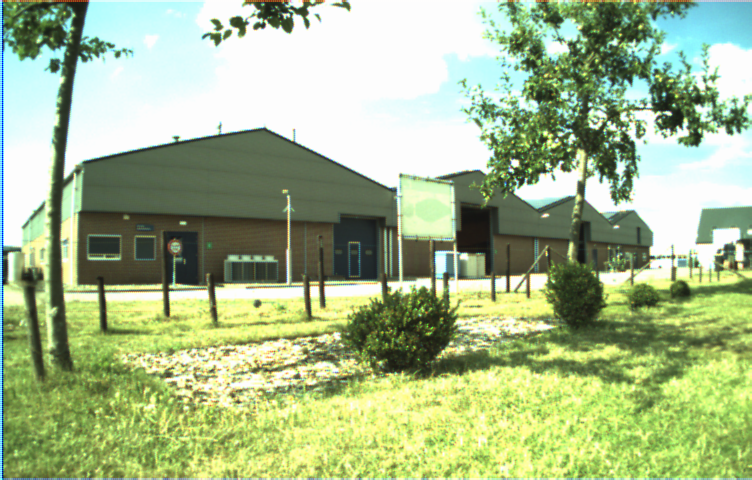}
    \label{fig:tbplaces_positive_query}
    }
   \subfloat[Map]{
    \includegraphics[width=.3\columnwidth]{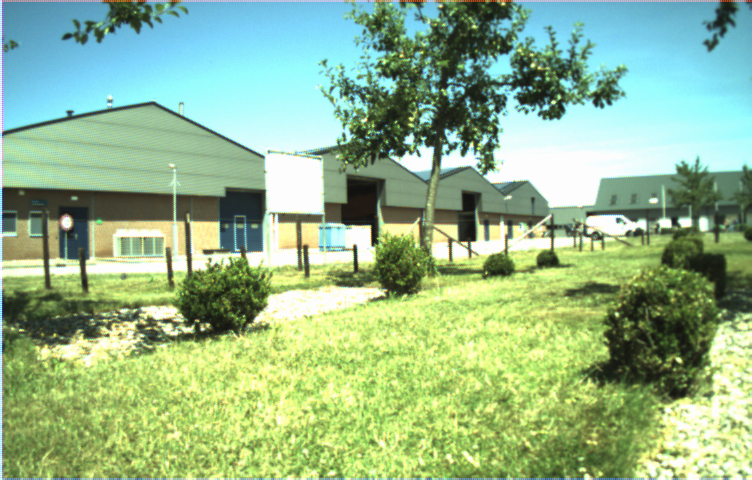}
    \label{fig:tbplaces_positive_db}
    }
    \subfloat[FoV overlap (74\%)]{
    \includegraphics[width=.33\columnwidth, trim=1cm 2cm 1cm 2cm, clip]{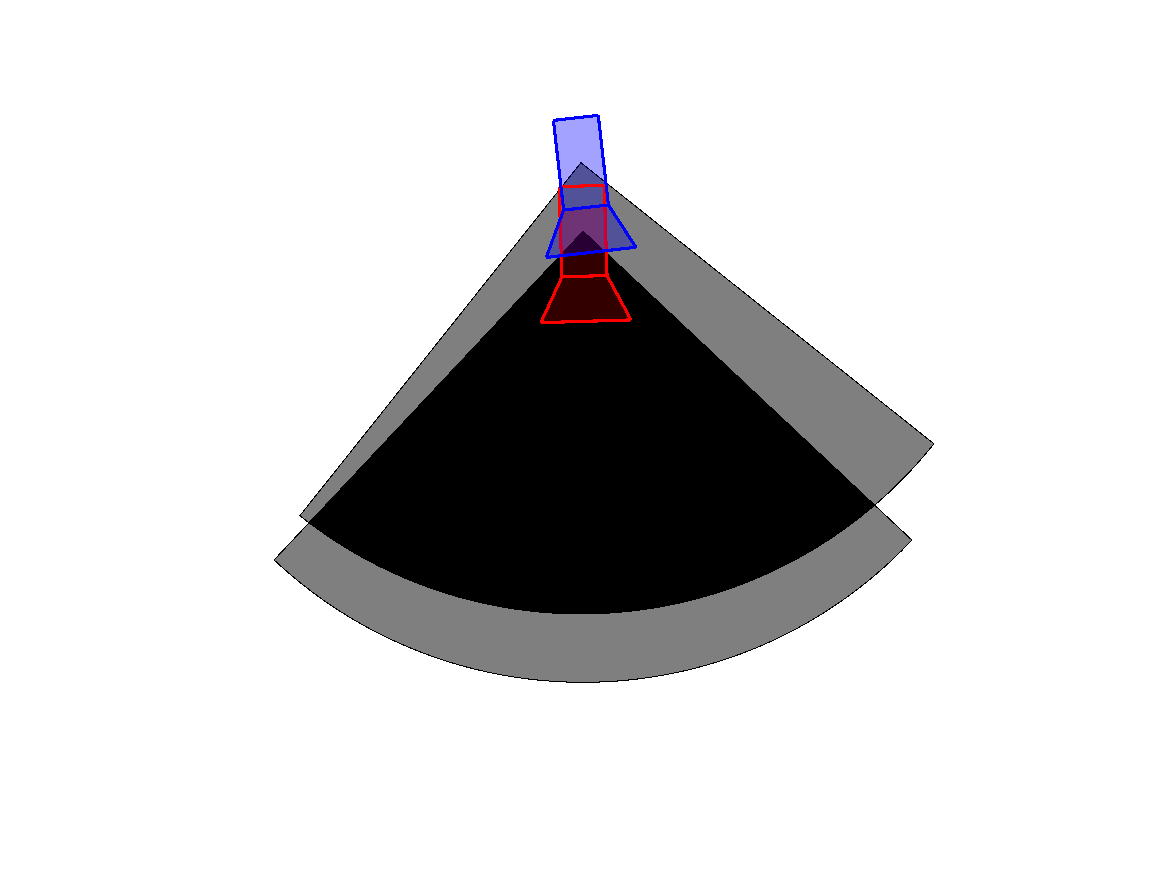}
    \label{fig:tbplaces_positive_fov}
    }\\
   \subfloat[Query]{
    \includegraphics[width=.3\columnwidth]{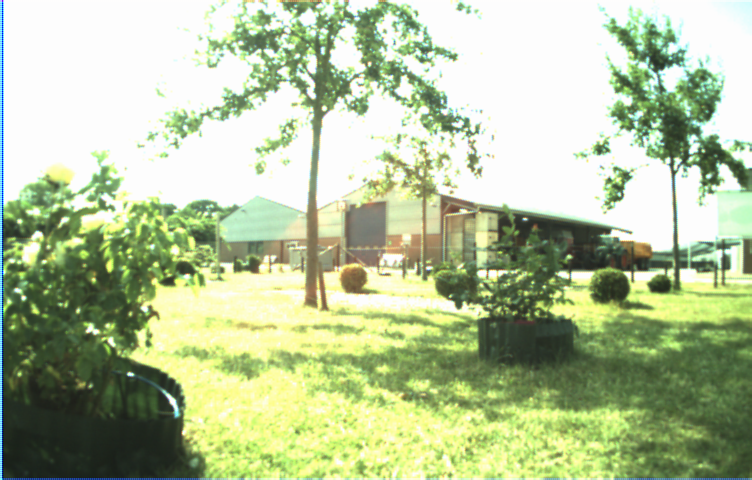}
    \label{fig:tbplaces_negative_query}
    }
   \subfloat[Map]{
    \includegraphics[width=.3\columnwidth]{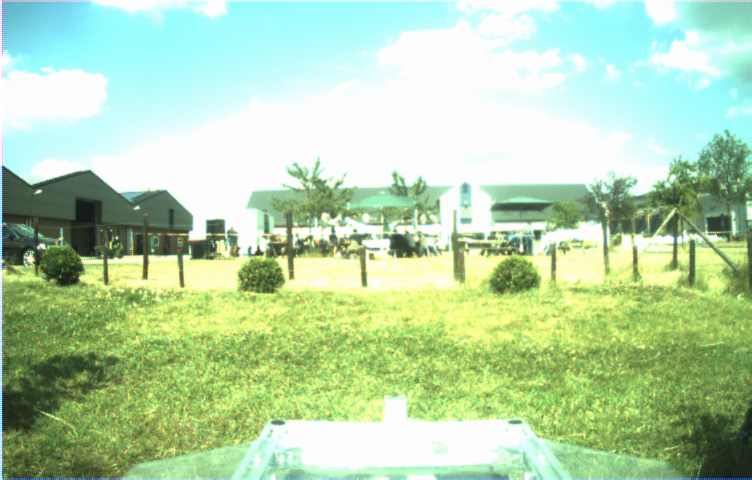}
    \label{fig:tbplaces_negative_db}
    }
    \subfloat[FoV overlap (41\%)]{
    \includegraphics[width=.33\columnwidth, trim=1cm 2cm 1cm 2cm, clip]{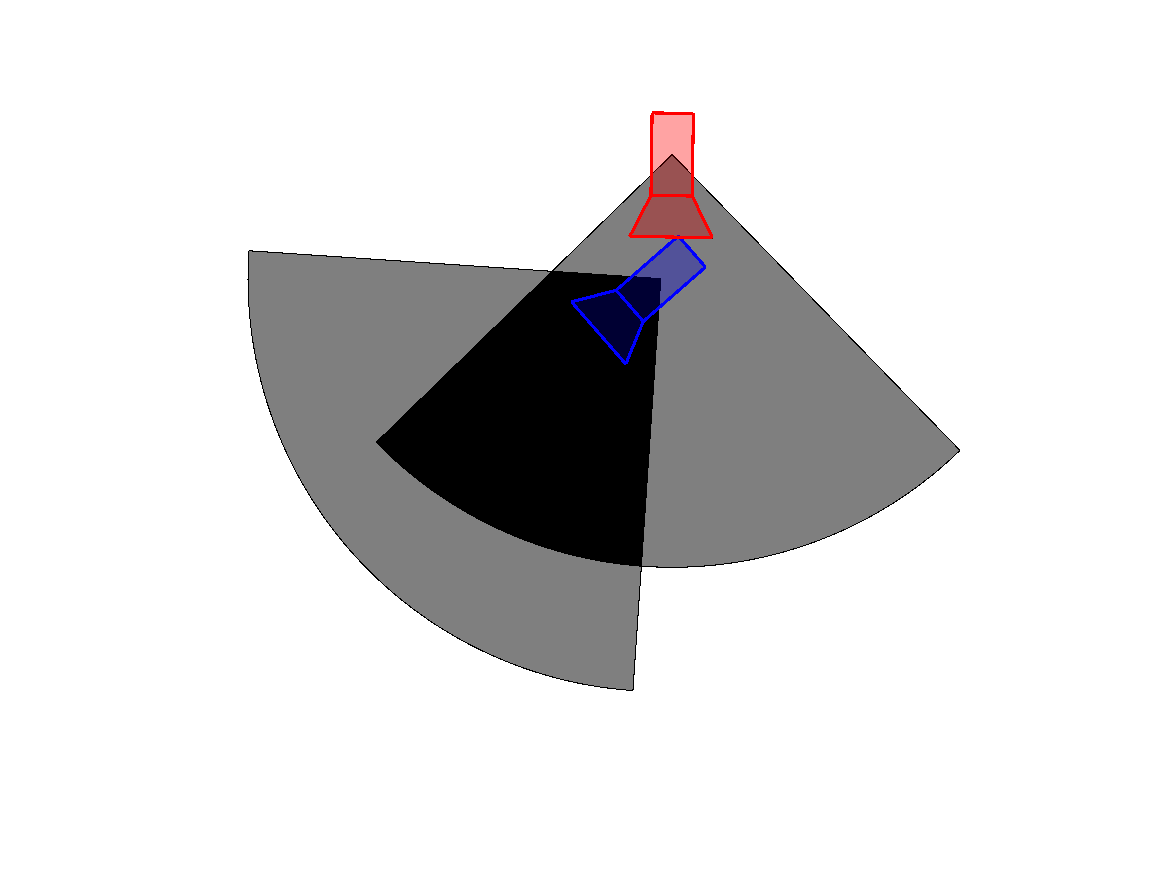}
    \label{fig:tbplaces_negative_fov}
    }
    \caption{Example image pairs from the TB-Places dataset. The first row shows a positive pair with FoV overlap of 74\%. The second row depicts a soft negative pair with FoV overlap equal to 41\%. The red camera corresponds to the query image, while the blue one to the map image.}
    \label{fig:tbplaces_fov}
\end{figure}

This is the case of the TB-Places dataset~\cite{leyvavallina2019access}, for visual place recognition in garden environments, created for the Trimbot2020 project~\cite{strisciuglio2018trimbot2020}. It contains images taken in an experimental garden over three years and includes variations in illumination, season and viewpoint.  Each image comes with a 6DOF camera pose, which allows us to estimate a very precise 2D FoV. According to the original paper of the TB-places dataset, we set the FoV angle of the cameras as $\theta = 90^{\circ}$ and the radius as $r=3.5m$. We thus estimate the 2D Field-of-View overlap and use it to re-label the pairs of images contained in the dataset. \revised{We provide the similarity ground truth for all possible pairs within W17, the training set.} We show some examples of image pairs and their 2D FoV overlap in Fig.~\ref{fig:tbplaces_fov}.

\subsection{3D Field-of-View overlap}
\label{sec:3Dfov}

When a 3D reconstruction of a concerned environment is available, we propose to estimate the degree of similarity of image pairs by computing the \emph{3D Field-of-View overlap}. We project a given image with an associated 6DOF camera pose onto the reconstructed pointcloud of the environment. 
We select the subset of 3D points that falls within the boundaries of the image as the image 3D FoV. For an image pair, we compute their 3D FoV overlap as the intersection-over-union (IoU) of the sets of 3D points associated with the two images. \revised{This computation is similar to the \emph{maximum inliers} measure proposed in~\cite{radenovic2018fine}, where it was used as part of a pair-mining strategy that also involved the computation of a distance in the latent-space.} We consider, instead, the computed 3D FoV overlap as a measure of the degree of similarity of a pair of images\revised{, and use it to compute the graded ground truth.} 

\begin{figure}[t]
    \centering
\subfloat[Query]{
    \includegraphics[width=.3\columnwidth]{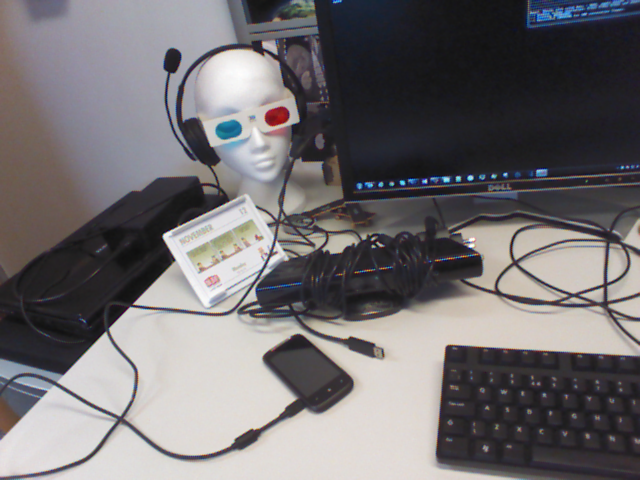}
    \label{fig:heads_query}
    }
   \subfloat[Map]{
    \includegraphics[width=.3\columnwidth]{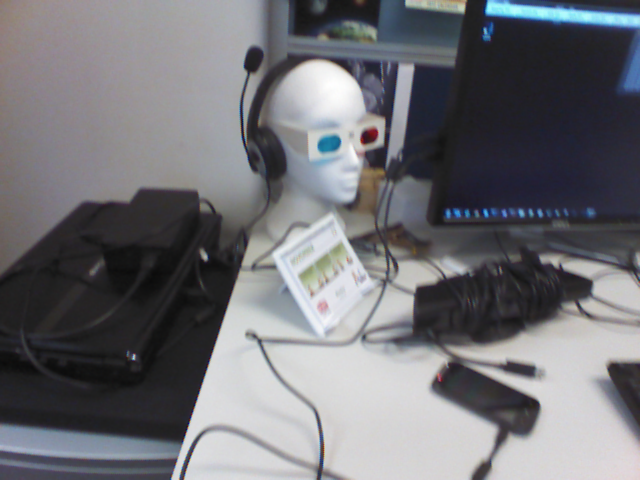}
    \label{fig:heads_db}
    }
    \hspace{-4pt}\subfloat[3D FoV overlap (75\%)]{
    \includegraphics[width=.35\columnwidth,trim=0 0 0 7cm, clip]{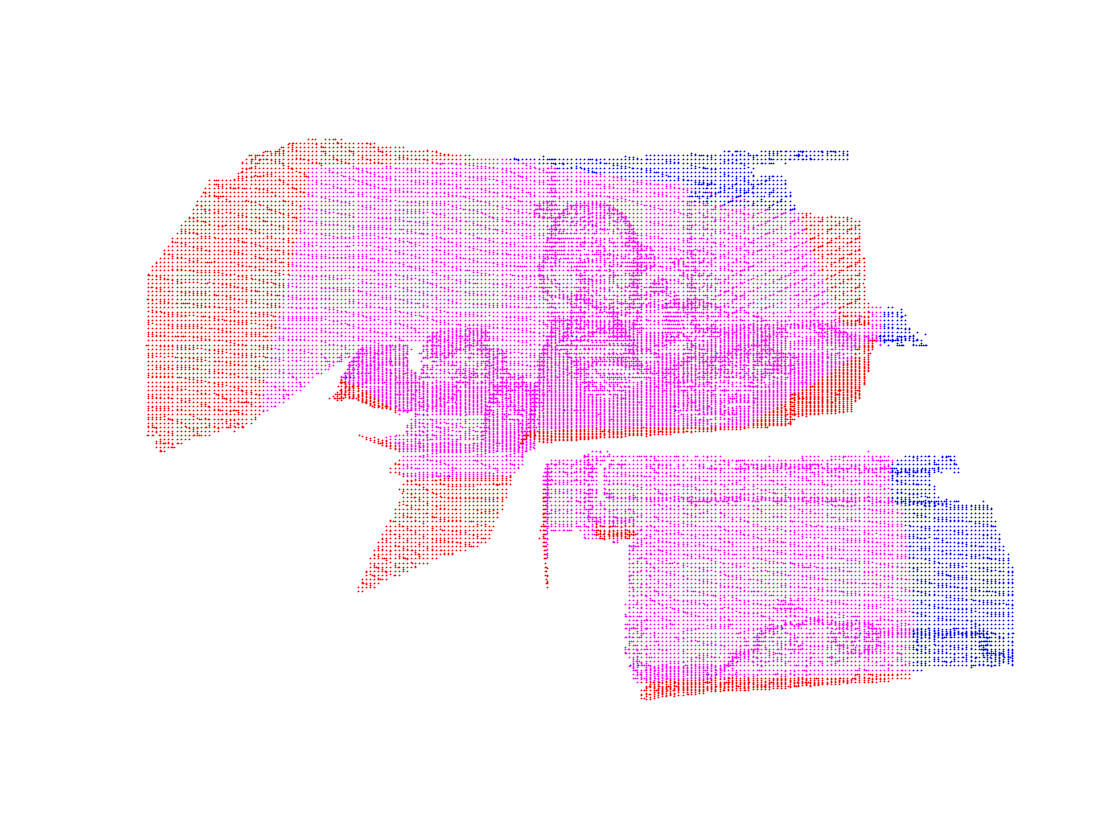}
    \label{fig:heads_fov}
    }
   
   \subfloat[Query]{
    \includegraphics[width=.3\columnwidth]{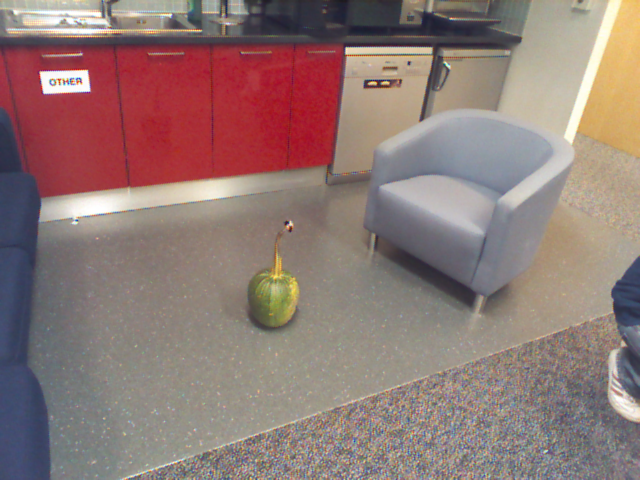}
    \label{fig:pumpkin_query}
    }
   \subfloat[Map]{
    \includegraphics[width=.3\columnwidth]{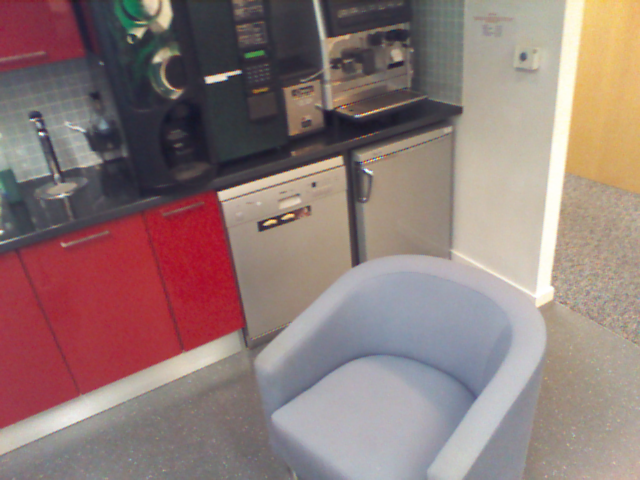}
    \label{fig:pumpkin_db}
    }
    \hspace{-4pt}\subfloat[3D FoV overlap (50\%)]{
    \includegraphics[width=.35\columnwidth]{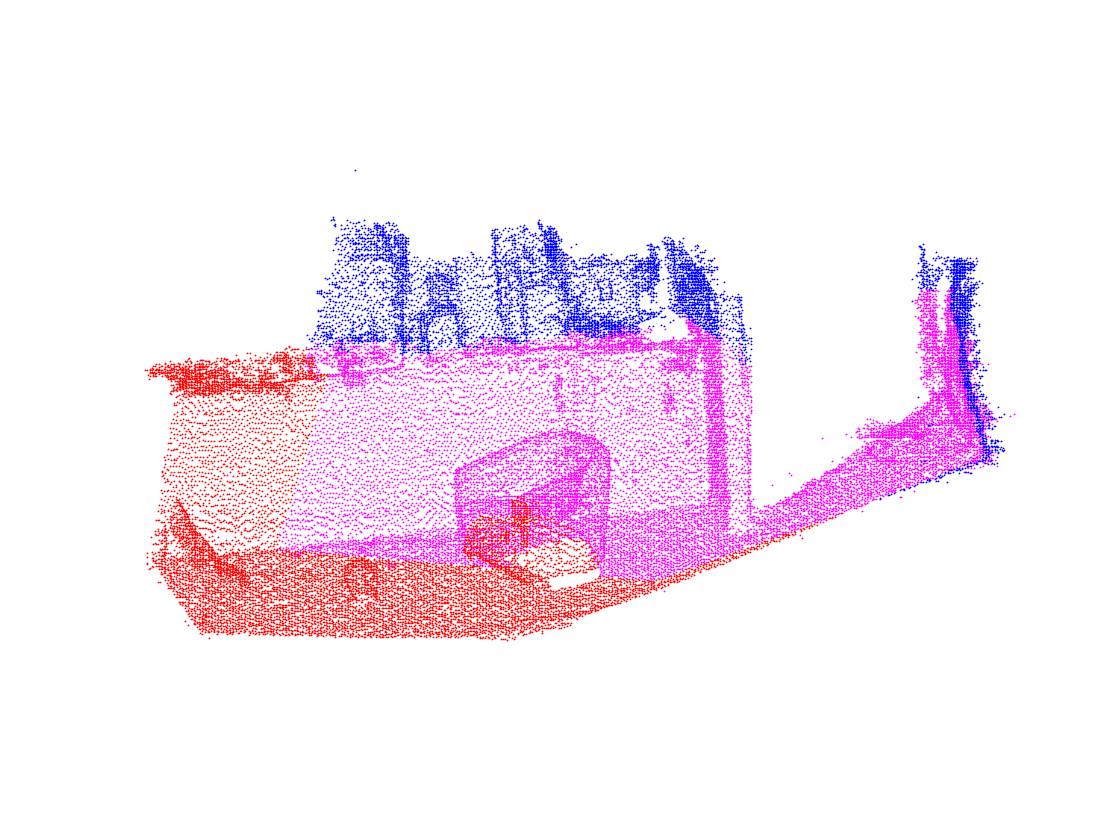}
    \label{fig:pumpkin_fov}
    }
       
   \subfloat[Query]{
    \includegraphics[width=.3\columnwidth]{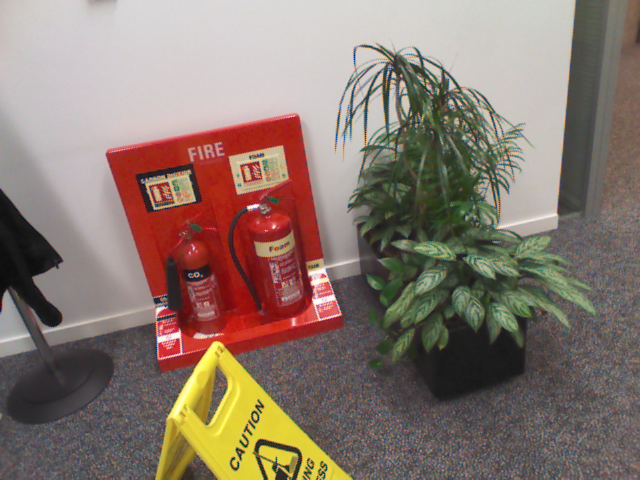}
    \label{fig:fire_query}
    }
   \subfloat[Map]{
    \includegraphics[width=.3\columnwidth]{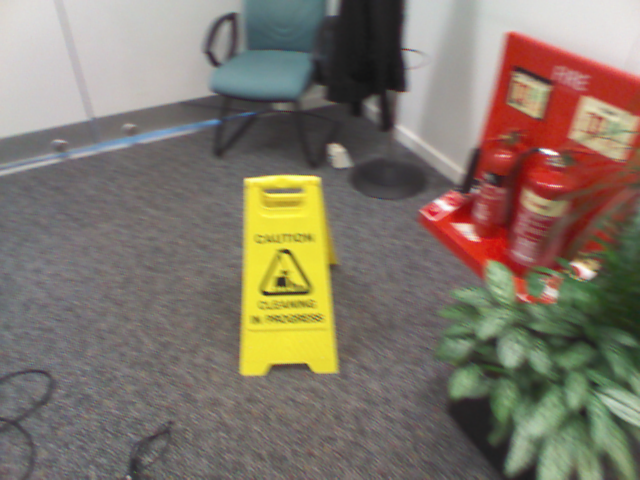}
    \label{fig:fire_db}
    }
    \hspace{-4pt}\subfloat[3D FoV overlap (25\%)]{
    \includegraphics[width=.35\columnwidth]{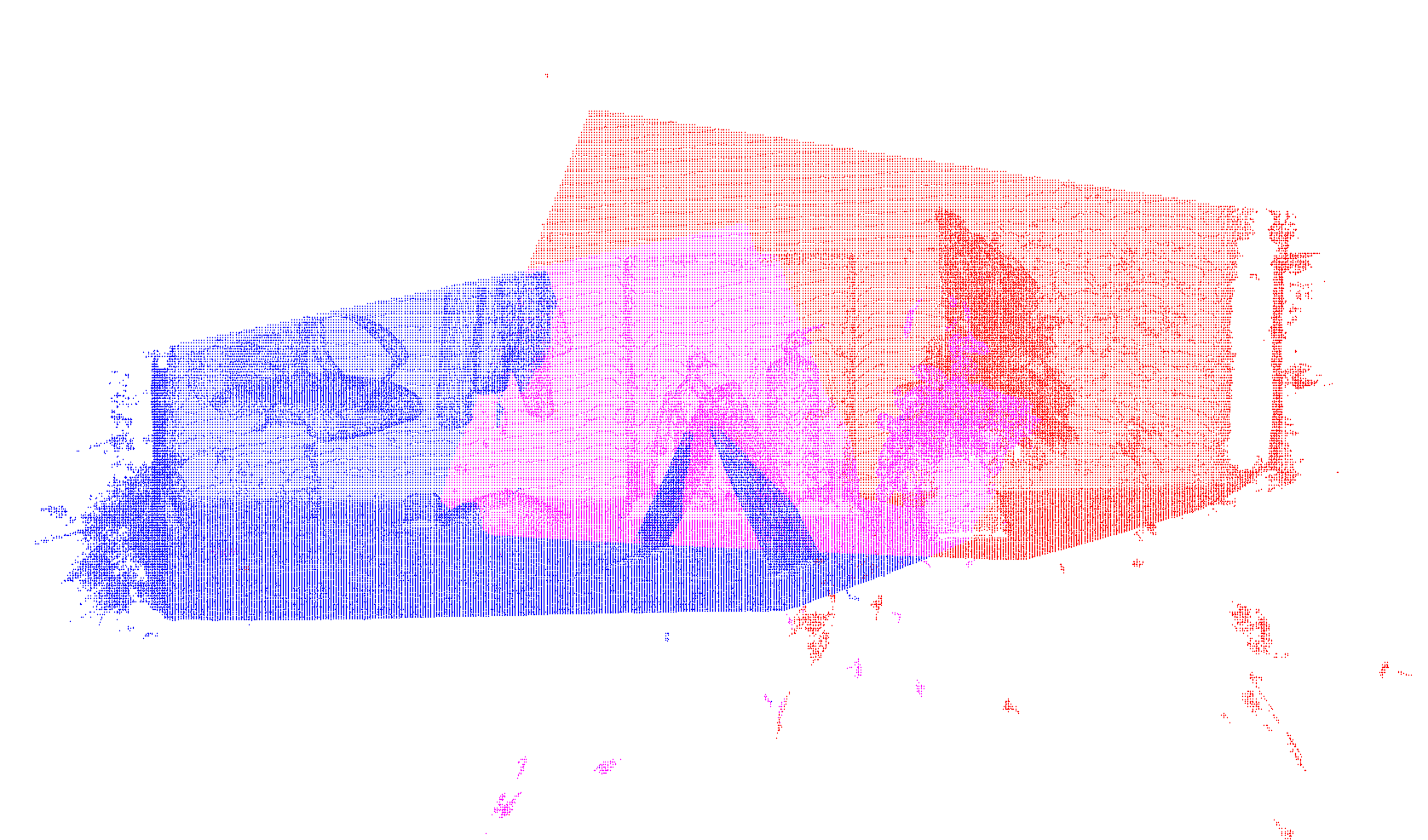}
    \label{fig:fire_fov}
    }
    \caption{3D Field-of-View overlap examples from the 7Scenes dataset. The first row shows a positive pair with 3D FoV overlap of 75\%. The second row depicts a borderline pair with 50\% 3D FoV overlap. The last row shows a soft negative image pair with 25\% 3D FoV overlap.
    The red pointcloud corresponds to the query 3D FoV, the blue one is the map, and the magenta represents the overlap between them.}
    \label{fig:7scenes_fov}
\end{figure}

We use the \emph{3D Field-of-View overlap} to re-annotate the 7Scenes dataset~\cite{Shotton2013}, an indoor localization benchmark, that contains RGBD images taken in seven environments. Each image has an associated 6DOF pose, and a 3D reconstruction of each scene is provided. \revised{We provide annotations for each possible pair within the training set per scene. We also compute the test-vs-train similarities for evaluation purposes. }We show some examples of the 3D FoV overlap in Fig.~\ref{fig:7scenes_fov}. We display the 3D FoV associated to the query image in red, that of the map image in blue and their overlap in magenta. 

\subsection{Selection of training pairs}
\label{sec:training_pairs}
Training a model for visual place recognition is usually formulated as a binary classification problem. The aim is to determine whether two images depict a similar (class 1) or dissimilar (class 2) place. 
To train binary classifiers, it is generally desirable to have balanced datasets, to ensure that the two classes are equally weighted. This is often not the case for visual place recognition, where the dissimilar image pairs significantly outnumber the positive pairs. Moreover, when training a siamese architecture, it is necessary to form batches with meaningful image pairs or triplets. For instance, if the training batch consists of pairs that are too easy, the learning process might stall, and the weights of the network not be updated. If the pairs are too difficult, the training dynamics can become unstable~\cite{shi2016embedding}. Hence, the selection of image pairs is a crucial element of the training. 

Existing approaches use complex image pair mining strategies. In the case of  NetVLAD~\cite{Arandjelovic2017} \revised{and NetVLAD-based methods like Patch-NetVLAD~\cite{hausler2021patch} or NetVLAD-SARE~\cite{liu2019stochastic}}, the authors use a Hard Negative Mining strategy. For each query image, they select a set of potential positive map images and a set with its 10 hardest negative map images. The best combination of positive and negative samples is selected by means of their contribution to a Triplet Ranking Loss function. This implies that all the images need to be forwarded to the network to compute the value of the loss function, although only three contribute to the learning. In~\cite{msls}, the authors deploy a similar strategy, selecting only the top-5 hardest negatives. For each triplet, many image latent representations are computed. Even with a caching strategy, bigger backbones like VGG cannot be trained in their entirety~\cite{Arandjelovic2017} and large training batches do not fit in the memory of a regular GPU. In~\cite{msls}, indeed, the authors use a batch size of 4 triplets. 

\revised{A continuous similarity ground truth can lead to improved learning~\cite{thoma2020soft}, but it does not dispense with the need for hard mining strategies by itself. When relying only on the GPS without taking in consideration the orientation, for instance, the ground truth annotations are likely to contain visual inaccuracies, i.e. two images taken 20cm apart facing in opposite directions will not be visually similar, despite having a high annotated similarity. These contradictions make necessary the use of further curation of the selected pairs or triplets by means of a costly hard mining strategy.}

We argue that a better curated and more informative ground truth can replace these complicate mining techniques. We use our graded similarity annotations, presented in Sections~\ref{sec:2Dfov} and~\ref{sec:3Dfov} to select the image pairs that compose a training batch. For our training process, we ensure that each training batch contains a balanced amount of positive and negative pairs. In the case of the negative pairs, we also ensure that half of them are soft negatives (i.e. their annotated similarity is higher than 0). \revised{We explore several batch composition strategies and discuss them in Section~\ref{sec:discussion}.} This means that each selected image pair consist of a query and a match image, and we do not require complicated mining strategies or to compute any latent representations to form the pairs. Hence, we are able to train our models with a batch size of 64.

%% file: figures/4_data/twod_fov_tex.pdf_tex
\begingroup%
  \makeatletter%
  \providecommand\color[2][]{%
    \errmessage{(Inkscape) Color is used for the text in Inkscape, but the package 'color.sty' is not loaded}%
    \renewcommand\color[2][]{}%
  }%
  \providecommand\transparent[1]{%
    \errmessage{(Inkscape) Transparency is used (non-zero) for the text in Inkscape, but the package 'transparent.sty' is not loaded}%
    \renewcommand\transparent[1]{}%
  }%
  \providecommand\rotatebox[2]{#2}%
  \newcommand*\fsize{\dimexpr\f@size pt\relax}%
  \newcommand*\lineheight[1]{\fontsize{\fsize}{#1\fsize}\selectfont}%
  \ifx\svgwidth\undefined%
    \setlength{\unitlength}{171.00101852bp}%
    \ifx\svgscale\undefined%
      \relax%
    \else%
      \setlength{\unitlength}{\unitlength * \real{\svgscale}}%
    \fi%
  \else%
    \setlength{\unitlength}{\svgwidth}%
  \fi%
  \global\let\svgwidth\undefined%
  \global\let\svgscale\undefined%
  \makeatother%
  \begin{picture}(1,1.25386037)%
    \lineheight{1}%
    \setlength\tabcolsep{0pt}%
    \put(0,0){\includegraphics[width=\unitlength,page=1]{figures/4_data/twod_fov_tex.pdf}}%
    \put(0.3,0.51){\makebox(0,0)[lt]{\lineheight{1.25}\smash{\begin{tabular}[t]{l}\color[rgb]{0,0,1}{$\theta$}\end{tabular}}}}%
    \put(0,0){\includegraphics[width=\unitlength,page=2]{figures/4_data/twod_fov_tex.pdf}}%
    \put(0.1,0.1){\makebox(0,0)[lt]{\lineheight{1.25}\smash{\begin{tabular}[t]{l}$(t_0,t_1)$\end{tabular}}}}%
    \put(0.5,0.6){\color[rgb]{0,0,0}\makebox(0,0)[lt]{\lineheight{1.25}\smash{\begin{tabular}[t]{l}$r$\end{tabular}}}}%
    \put(0.2,1.1){\color[rgb]{0,0,0}\makebox(0,0)[lt]{\lineheight{1.25}\smash{\begin{tabular}[t]{l}$N$\end{tabular}}}}%
    \put(0.3,0.7){\color[rgb]{1,0,0}\makebox(0,0)[lt]{\lineheight{1.25}\smash{\begin{tabular}[t]{l}$\alpha$\end{tabular}}}}%
  \end{picture}%
\endgroup%

%% file: sections/5_experiments.tex
\label{sec:experiments}
\subsection{Datasets}
\textbf{Mapillary Street Level Sequences (MSLS). }
This is a large scale place recognition dataset that contains images taken in 30 different cities across six continents~\cite{msls}. It includes challenging variations of camera viewpoint, season, time and illumination. Moreover, the images have been taken with different cameras. The training set contains over 500k query images and 900k map images, taken in 22 cities. The validation set consists of 19k map images and 11k query images, taken in two cities, and the test set has 39k map images and 27k query images, taken in six different cities.
 
We evaluate our models on the MSLS validation set and on the MSLS test set. For the former, we use the evaluation script published by the authors of the dataset. For the latter, since the ground truth is not publicly available, we submit our predictions to the official evaluation server. Following the protocol established by the authors of the dataset~\cite{msls}, two images are considered as similar if they are taken by cameras located within $25m$ of distance, and with less than $40^{\circ}$ of viewpoint variation.

\textbf{Pittsburgh. }
This dataset contains images of urban environments gathered from Google Street View in the city of Pittsburgh, Pennsylvania, USA~\cite{Torii-PAMI2015}, taken in the span of several years. We  \revised{use the Pitts30k as test set~\cite{Arandjelovic2017}, a subset of the Pittsburgh dataset which consists of }10k map images and 7k query images, to evaluate the generalization of our models trained on MSLS.

\textbf{Tokyo 24/7. }
It consists of images taken in Tokyo, Japan, containing large variations of illumination, as they are taken during day and night~\cite{Torii-CVPR2013}. The query set consists of 315 images, and the map contains 76k photos.

\revised{
\textbf{RobotCar Seasons v2.}
This dataset~\cite{sattler2018benchmarking} is a subset of Oxford RobotCar~\cite{Maddern2017}, a visual localization dataset with images taken in the city of Oxford, UK, within the span of twelve months. Its query set contains 1872 images under different lighting and weather conditions, and its reference set consists of 21k images taken during at daytime, with overcast weather. 

\textbf{Extended CMU Seasons. }
This is an expansion of the CMU Seasons dataset~\cite{sattler2018benchmarking}, which in turn is a subset of the CMU Dataset~\cite{Badino2011}. It contains images taken in urban, suburban and park environments, in the city of Pittsburgh, over a period of 1 year. Its query set contains 61k images, and its reference set contains 57k images.
}

\textbf{TB-Places. }
This dataset was designed for place recognition in garden environments~\cite{leyvavallina2019access, leyvavallina2019caip}. It contains images taken by a rover robot in a garden at the University of Wageningen, the Netherlands, over the course of three years. The dataset was collected for the TrimBot2020 project~\cite{strisciuglio2018trimbot2020}. It includes drastic viewpoint variations, as well as illumination changes. The garden is a very challenging small environment with repetitive textures. 
\begin{figure}[t!]
    \centering
   \subfloat[]{
    \includegraphics[width=.48\columnwidth]{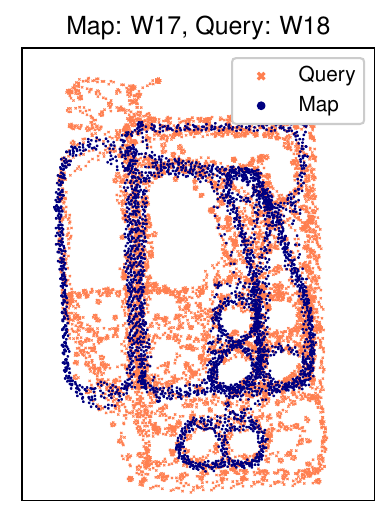}
    \label{fig:w17_18}
    }\hspace{-10pt}~\subfloat[]{
    \includegraphics[width=.48\columnwidth]{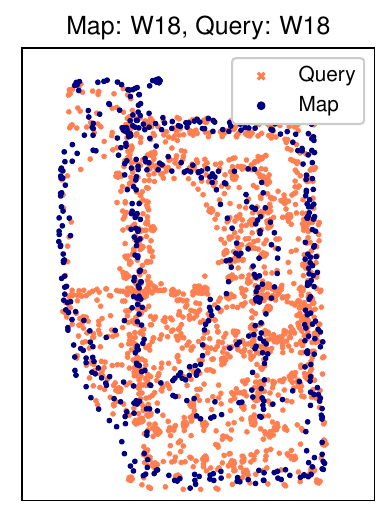}
    \label{fig:w18_w18}
    }
    \caption{Configurations of the experiments on the TB-Places dataset. (a) W17 subset is the map set, and W18 is the query. (b) We divide W18 into map and query. For visualization purposes, the trajectories have been downsampled.}
    \label{fig:tb_places_test_sets}
\end{figure}

The dataset consists of three subsets, i.e. W16, that contains 41k images taken in 2016; W17, that includes 11k images taken in 2017; and W18, that has 23k images taken in 2018. As in~\cite{leyvavallina2019caip} we use the W17 subset to train our models.
We design two experiments to evaluate our models. For the first one we establish W17 as map set (11k images) and W18 as query (23k images). With this configuration we aim to test the robustness of our models w.r.t. changes between the map and the query sets. For the second experiment, we divide W18 into query (17k images) and map (6k images) to test the generalization capabilities of our models in the case both map and query sets were not used for training. In Fig.~\ref{fig:tb_places_test_sets}, we show a sketch of the trajectory that the robot covered in the TrimBot2020 garden for the recording of the reference map (blue trajectory) and query (orange trajectory) images. It is worth pointing out that the query images were taken from locations not covered by map images, thus including substantial viewpoint variations.

\textbf{7Scenes. }
It is a benchmark dataset for indoor camera localization algorithms~\cite{Shotton2013}. It includes 26k training images and 17k test images, taken in seven different environments. Each image has an associated ground truth 6DOF camera pose. Additionally, a 3D reconstruction of each scene is available. We use it to test our models for visual place recognition in indoor environments. For evaluation purposes, we define an image pair as a positive match if their annotated degree of similarity is higher than $50\%$. We use the training set as map, and the test set as query.


\subsection{Implementation details}
We implemented our models using the PyTorch framework~\footnote{\url{http://pytorch.org/}} and trained them on a \revised{NVIDIA} V100 GPU card. 
We optimized the weights of our models with Stochastic Gradient Descent. To train the networks that use the binary Contrastive Loss function, we set an initial learning rate $l_0=0.01$. For those that deploy the Generalized Contrastive Loss function, we used as initial learning rate $l_0=0.1$. In both cases the learning rate is divided by $10$ after every \revised{250k pairs} for MSLS, or every 15 epochs for TB-Places and 7Scenes. We set the margin value $\tau=0.5$ and we use a batch size of 64. We trained the two last convolutional blocks of the backbone networks for all experiments. The necessary code to replicate the experiments is publicly available~\footnote{\url{http://github.com/marialeyvallina/generalized_contrastive_loss}}.

\begin{table*}[t!]
\centering
\caption{\revised{Ablation study on the considered datasets: all the models are trained on the MSLS train set. CL stands for Contrastive Loss and while GCL for our Generalized Contrastive Loss. For the cases in which PCA whitening is applied we report the dimensionality that achieves the best results on the MSLS validation set. } }
\label{tab:msls-ablation}
\resizebox{\textwidth}{!}{%
\setlength\tabcolsep{1.5pt}
\revised{
\begin{tabular}{@{}llcc@{\hspace{4\tabcolsep}}ccc@{\hspace{4\tabcolsep}}ccc@{\hspace{4\tabcolsep}}ccc@{\hspace{4\tabcolsep}}ccc@{\hspace{4\tabcolsep}}ccc@{\hspace{4\tabcolsep}}ccc@{}}
 &  &  &  & \multicolumn{3}{c}{\textbf{MSLS-Val}} & \multicolumn{3}{c}{\textbf{MSLS-Test}} & \multicolumn{3}{c}{\textbf{Pitts30k}} & \multicolumn{3}{c}{\textbf{Tokyo24/7}} & \multicolumn{3}{c}{\textbf{RobotCar Seasons v2}} & \multicolumn{3}{c}{\textbf{Extended CMU Seasons}} \\

\textbf{Method} & \textbf{Loss} & \textbf{PCA$_w$} & \textbf{Dim} & \textbf{R@1} &\textbf{ R@5} & \textbf{R@10} & \textbf{R@1} & \textbf{R@5} & \textbf{R@10} &\textbf{ R@1} & \textbf{R@5} & \textbf{R@10} & \textbf{R@1} & \textbf{R@5 }& \textbf{R@10} & \textbf{0.25m/2$\degree$} & \textbf{0.5m/5$\degree$} & \textbf{5.0m/10$\degree$} & \textbf{0.25m/2$\degree$} & \textbf{0.5m/5$\degree$} & \textbf{5.0m/10$\degree$} \\
\toprule
\multirow{4}{*}{VGG-GeM} & CL & N & 512 & 47.0 & 60.3 & 65.5 & 27.9 & 40.5 & 46.5 & 51.2 & 71.9 & 79.7 & 24.1 & 39.4 & 47.0 & 3.1 & 13.2 & 55.0 & 2.8 & 8.6 & 44.5 \\
 & GCL & N & 512 & 65.9 & 77.8 & 81.4 & 41.7 & 55.7 & 60.6 & 61.6 & 80.0 & 86.0 & 34.0 & 51.1 & 61.3 & 3.7 & 15.8 & 59.7 & 3.6 & 11.2 & 55.8 \\
 & CL & Y & 512 & 61.4 & 75.1 & 78.5 & 36.3 & 49.0 & 54.1 & 64.7 & 81.5 & 86.8 & 36.2 & 54.0 & 57.8 & 4.2 & 18.7 & 62.5 & 4.4 & 13.4 & 56.5 \\
 & GCL & Y & 512 & 72.0 & 83.1 & 85.8 & 47.0 & 60.8 & 65.5 & 73.3 & 85.9 & 89.9 & 47.6 & 61.0 & 69.2 & 5.4 & 21.9 & 69.2 & 5.7 & 17.1 & 66.3 \\ \midrule
\multirow{4}{*}{ResNet50-GeM} & CL & N & 2048 & 51.4 & 66.5 & 70.8 & 29.7 & 44.0 & 50.7 & 61.5 & 80.0 & 86.9 & 30.8 & 46.0 & 56.5 & 3.2 & 15.4 & 61.5 & 3.2 & 9.6 & 49.5 \\
 & GCL & N & 2048 & 66.2 & 78.9 & 81.9 & 43.3 & 59.1 & 65.0 & 72.3 & 87.2 & 91.3 & 44.1 & 61.0 & 66.7 & 2.9 & 14.0 & 58.8 & 3.8 & 11.8 & 61.6 \\
 & CL & Y & 1024 & 63.2 & 76.6 & 80.7 & 37.9 & 53.0 & 58.5 & 66.2 & 82.2 & 87.3 & 36.2 & 51.8 & 61.0 & 5.0 & 21.1 & 66.5 & 4.7 & 13.4 & 51.6 \\
 & GCL & Y & 1024 & 74.6 & 84.7 & 88.1 & 52.9 & 65.7 & 71.9 & 79.9 & 90.0 & 92.8 & 58.7 & 71.1 & 76.8 & 4.7 & 20.2 & 70.0 & 5.4 & 16.5 & 69.9 \\ \midrule
\multirow{4}{*}{ResNet152-GeM} & CL & N & 2048 & 58.0 & 72.7 & 76.1 & 34.1 & 50.8 & 56.8 & 66.5 & 83.8 & 89.5 & 34.6 & 57.1 & 63.5 & 3.3 & 15.2 & 64.0 & 3.2 & 9.7 & 52.2 \\
 & GCL & N & 2048 & 70.3 & 82.0 & 84.9 & 45.7 & 62.3 & 67.9 & 72.6 & 87.9 & 91.6 & 34.0 & 51.8 & 60.6 & 2.9 & 13.1 & 63.5 & 3.6 & 11.3 & 63.1 \\
 & CL & Y & 2048 & 66.9 & 80.9 & 83.8 & 44.8 & 59.2 & 64.8 & 71.2 & 85.8 & 89.8 & 54.3 & 68.9 & 75.6 & \textbf{6.1 } & \textbf{23.5} & 68.9 & 4.8 & 14.2 & 55.0 \\
 & GCL & Y & 2048 & 79.5 & 88.1 & 90.1 & 57.9 & 70.7 & 75.7 & \textbf{80.7} & \textbf{91.5} & \textbf{93.9} & \textbf{69.5} & \textbf{81.0} & \textbf{85.1} & 6.0 & 21.6 & 72.5 & 5.3 & 16.1 & 66.4 \\ \midrule
\multirow{4}{*}{ResNeXt-GeM} & CL & N & 2048 & 62.6 & 76.4 & 79.9 & 40.8 & 56.5 & 62.1 & 56.0 & 77.5 & 85.0 & 37.8 & 54.9 & 62.5 & 1.9 & 10.4 & 54.8 & 2.9 & 9.0 & 52.6 \\
 & GCL & N & 2048 & 75.5 & 86.1 & 88.5 & 56.0 & 70.8 & 75.1 & 64.0 & 81.2 & 86.6 & 37.8 & 53.6 & 62.9 & 2.7 & 13.4 & 65.2 & 3.5 & 10.5 & 58.8 \\
 & CL & Y & 1024 & 74.3 & 87.0 & 89.6 & 49.9 & 63.8 & 69.4 & 70.9 & 85.7 & 90.2 & 50.8 & 67.6 & 74.3 & 3.8 & 17.2 & 68.2 & 4.9 & 14.4 & 61.7 \\
 & GCL & Y & 1024 & \textbf{80.9} & \textbf{90.7} & \textbf{92.6} & \textbf{62.3} & \textbf{76.2} & \textbf{81.1} & 79.2 & 90.4 & 93.2 & 58.1 & 74.3 & 78.1 & 4.7 & 21.0 & \textbf{74.7} & \textbf{6.1} & \textbf{18.2} & \textbf{74.9} \\ \bottomrule
\end{tabular}
}}
\end{table*}

\begin{table*}[t!]
\centering
\caption{\revised{Comparison of our results with those of state-of-the-art approaches on the considered datasets. For each model, we report if PCA whitening is used and the dimensionality of the learned image latent vector.The best result by our method is underlined, and the overall best in presented in bold font.}}
\label{tab:msls-sota}
\resizebox{\textwidth}{!}{%
\setlength\tabcolsep{1.5pt}
\revised{
\begin{tabular}{@{}lcc@{\hspace{4\tabcolsep}}ccc@{\hspace{4\tabcolsep}}ccc@{\hspace{4\tabcolsep}}ccc@{\hspace{4\tabcolsep}}ccc@{\hspace{4\tabcolsep}}ccc@{\hspace{4\tabcolsep}}ccc@{}}
 &  &  & \multicolumn{3}{c}{\textbf{MSLS-Val}} & \multicolumn{3}{c}{\textbf{MSLS-Test}} & \multicolumn{3}{c}{\textbf{Pitts30k}} & \multicolumn{3}{c}{\textbf{Tokyo24/7}} & \multicolumn{3}{c}{\textbf{RobotCar Seasons v2}} & \multicolumn{3}{c}{\textbf{Extended CMU Seasons}} \\ 
\textbf{Method} & \textbf{PCA$_w$} & \textbf{Dim} & \textbf{R@1} & \textbf{R@5} & \textbf{R@10} & \textbf{R@1} & \textbf{R@5} & \textbf{R@10} & \textbf{R@1} & \textbf{R@5} & \textbf{R@10} & \textbf{R@1} & \textbf{R@5 }& \textbf{R@10} & \textbf{0.25m/2$\degree$} & \textbf{0.5m/5$\degree$ }& \textbf{5.0m/10$\degree$ }& \textbf{0.25m/2$\degree$ }& \textbf{0.5m/5$\degree$ }& \textbf{5.0m/10$\degree$} \\\toprule 
NetVLAD 64 Pitts & N & 32768 & 47.7 & 62.8 & 70.9 & 30.7 & 41.9 & 46.4 & 82.1 & 91.4 & 93.8 & 62.2 & 73.7 & 78.4 & 5.6 & 22.0 & 71.0 & 10.7 & 27.8 & 84.1 \\
NetVLAD 64 Pitts & Y & 4096 & 70.7 & 81.4 & 84.6 & 30.6 & 41.9 & 47.5 & 83.7 & 91.8 & 94.0 & 67.0 & 77.8 & 80.3 & 5.8 & 23.1 & 73.2 & 11.6 & 30.3 & 87.5 \\
NetVLAD 64 MSLS & N & 32768 & 44.6 & 61.1 & 66.4 & 28.8 & 44.0 & 50.7 & 40.4 & 64.5 & 74.2 & 11.4 & 24.1 & 31.4 & 2.0 & 9.2 & 45.5 & 1.3 & 4.5 & 31.9 \\
NetVLAD 64 MSLS & Y & 4096 & 70.1 & 80.8 & 84.9 & 45.1 & 58.8 & 63.7 & 68.6 & 84.7 & 88.9 & 34.0 & 47.6 & 57.1 & 4.2 & 18.0 & 68.1 & 3.9 & 12.1 & 58.4 \\
NetVLAD 16 MSLS & N & 8192 & 49.5 & 65.0 & 71.8 & 29.3 & 43.5 & 50.4 & 48.7 & 70.6 & 78.9 & 13.0 & 33.0 & 43.8 & 1.8 & 9.2 & 48.4 & 1.7 & 5.5 & 39.1 \\
NetVLAD 16 MSLS & Y & 4096 & 70.5 & 81.1 & 84.3 & 39.4 & 53.0 & 57.5 & 70.3 & 84.1 & 89.1 & 37.8 & 53.3 & 61.0 & 4.8 & 17.9 & 65.3 & 4.4 & 13.7 & 61.4 \\
Patch NetVLAD & Y & 4096 & 79.5 & 86.2 & 87.7 & 48.1 & 57.6 & 60.5 & \textbf{88.7} & \textbf{94.5} & \textbf{95.9} & \textbf{86.0} & \textbf{88.6} & \textbf{90.5} & \textbf{9.6} & \textbf{35.3} & \textbf{90.9} & \textbf{11.8} & \textbf{36.2} & \textbf{96.2} \\
ResNet101-AP-GeM & Y & 2048 & 64.1 & 75.0 & 78.2 & 33.7 & 44.5 & 49.4 & 80.7 & 91.4 & 94.0 & 11.4 & 22.9 & 30.5 & 5.1 & 20.5 & 66.1 & 4.9 & 14.7 & 65.2 \\
NetVLAD-SARE & Y & 4096 & 68.1 & 77.3 & 82.4 & 34.4 & 44.3 & 48.8 & 87.8 & 94.3 & \textbf{95.9} & 79.7 & 86.7 & 90.5 & 7.4 & 26.5 & 81.3 & 6.4 & 19.4 & 75.5 \\ \midrule
VGG-GeM-GCL  & N & 512 & 65.9 & 77.8 & 81.4 & 41.7 & 55.7 & 60.6 & 61.6 & 80.0 & 86.0 & 34.0 & 51.1 & 61.3 & 3.7 & 15.8 & 59.7 & 3.6 & 11.2 & 55.8 \\
VGG-GeM-GCL  & Y & 512 & 72.0 & 83.1 & 85.8 & 47.0 & 60.8 & 65.5 & 73.3 & 85.9 & 89.9 & 47.6 & 61.0 & 69.2 & 5.4 & 21.9 & 69.2 & 5.7 & 17.1 & 66.3 \\
ResNet50-GeM-GCL & N & 2048 & 66.2 & 78.9 & 81.9 & 43.3 & 59.1 & 65.0 & 72.3 & 87.2 & 91.3 & 44.1 & 61.0 & 66.7 & 2.9 & 14.0 & 58.8 & 3.8 & 11.8 & 61.6 \\
ResNet50-GeM-GCL & Y & 1024 & 74.6 & 84.7 & 88.1 & 52.9 & 65.7 & 71.9 & 79.9 & 90.0 & 92.8 & 58.7 & 71.1 & 76.8 & 4.7 & 20.2 & 70.0 & 5.4 & 16.5 & 69.9 \\
ResNet152-GeM-GCL& N & 2048 & 70.3 & 82.0 & 84.9 & 45.7 & 62.3 & 67.9 & 72.6 & 87.9 & 91.6 & 34.0 & 51.8 & 60.6 & 2.9 & 13.1 & 63.5 & 3.6 & 11.3 & 63.1 \\
ResNet152-GeM-GCL & Y & 2048 & 79.5 & 88.1 & 90.1 & 57.9 & 70.7 & 75.7 & \underline{80.7} & \underline{91.5} & \underline{93.9} & \underline{69.5} & \underline{81.0} & \underline{85.1} & \underline{6.0} & \underline{21.6} & 72.5 & 5.3 & 16.1 & 66.4 \\
ResNeXt-GeM-GCL & N & 2048 & 75.5 & 86.1 & 88.5 & 56.0 & 70.8 & 75.1 & 64.0 & 81.2 & 86.6 & 37.8 & 53.6 & 62.9 & 2.7 & 13.4 & 65.2 & 3.5 & 10.5 & 58.8 \\
ResNeXt-GeM-GCL & Y & 1024 & \underline{\textbf{ 80.9}} & \underline{\textbf{90.7}} & \underline{\textbf{92.6}} & \underline{\textbf{62.3}} & \underline{\textbf{76.2}} & \underline{\textbf{81.1}} & 79.2 & 90.4 & 93.2 & 58.1 & 74.3 & 78.1 & 4.7 & 21.0 & \underline{74.7} & \underline{6.1} & \underline{18.2} & \underline{74.9} \\ \bottomrule

\end{tabular}
}}
\end{table*}

\begin{figure*}[t!]
\centering

\vspace{0.5cm}

\subfloat[MSLS validation set]{
\begin{tikzpicture}[thick,scale=1, every node/.style={scale=.8}]
\begin{axis}
[xlabel=K,
ylabel=Recall@K (\%), ylabel style={at={(axis description cs:-0.08,0.5)}},
legend pos =south east,grid, height=6cm, width=7.5cm]
\addplot[magenta, mark size=2pt, thick , mark=triangle,mark options=dashed, style=dashed] table [x=k, y=r, col sep=comma] {plot_data/MSLS/val/Pitts_vgg16_NetVLAD_64_TL_results_latex.csv};
\addplot[magenta, mark size=2pt, thick , mark=triangle,mark options=solid, style=solid] table [x=k, y=r, col sep=comma] {plot_data/MSLS/val/Pitts_vgg16_NetVLAD_64_TL_pca_results_latex.csv};
\addplot[black, mark size=2pt, thick , mark=pentagon,mark options=dashed, style=dashed] table [x=k, y=r, col sep=comma] {plot_data/MSLS/val/MSLS_vgg16_NetVLAD_16_TL_results_latex.csv};
\addplot[black, mark size=2pt, thick , mark=pentagon,mark options=solid, style=solid] table [x=k, y=r, col sep=comma] {plot_data/MSLS/val/MSLS_vgg16_NetVLAD_16_TL_pca_results_latex.csv};
\addplot[black, mark size=2pt, thick , mark=triangle,mark options=dashed, style=dashed] table [x=k, y=r, col sep=comma] {plot_data/MSLS/val/MSLS_vgg16_NetVLAD_64_TL_results_latex.csv};
\addplot[black, mark size=2pt, thick , mark=triangle,mark options=solid, style=solid] table [x=k, y=r, col sep=comma] {plot_data/MSLS/val/MSLS_vgg16_NetVLAD_64_TL_pca_results_latex.csv};
\addplot[green, mark size=2pt, thick , mark=triangle,mark options=solid, style=solid] table [x=k, y=r, col sep=comma] {plot_data/MSLS/val/MSLS_vgg16_NetVLAD_64_SARE_results_latex.csv};
\addplot[red, mark size=2pt, thick , mark=triangle,mark options=dashed, style=dashed] table [x=k, y=r, col sep=comma] {plot_data/MSLS/val/MSLS_vgg16_GeM_480_GCL_results_latex.csv};
\addplot[red, mark size=2pt, thick , mark=triangle,mark options=solid, style=solid] table [x=k, y=r, col sep=comma] {plot_data/MSLS/val/MSLS_vgg16_GeM_480_GCL_PCA_results_latex.csv};
\addplot[red, mark size=2pt, thick , mark=square,mark options=dashed, style=dashed] table [x=k, y=r, col sep=comma] {plot_data/MSLS/val/MSLS_resnet50_GeM_480_GCL_results_latex.csv};
\addplot[red, mark size=2pt, thick , mark=square,mark options=solid, style=solid] table [x=k, y=r, col sep=comma] {plot_data/MSLS/val/MSLS_resnet50_GeM_480_GCL_PCA_results_latex.csv};
\addplot[red, mark size=2pt, thick , mark=*,mark options=dashed, style=dashed] table [x=k, y=r, col sep=comma] {plot_data/MSLS/val/MSLS_resnet152_GeM_480_GCL_results_latex.csv};
\addplot[red, mark size=2pt, thick , mark=*,mark options=solid, style=solid] table [x=k, y=r, col sep=comma] {plot_data/MSLS/val/MSLS_resnet152_GeM_480_GCL_PCA_results_latex.csv};
\addplot[red, mark size=2pt, thick , mark=diamond,mark options=dashed, style=dashed] table [x=k, y=r, col sep=comma] {plot_data/MSLS/val/MSLS_resnext_GeM_480_GCL_results_latex.csv};
\addplot[red, mark size=2pt, thick , mark=diamond,mark options=solid, style=solid] table [x=k, y=r, col sep=comma] {plot_data/MSLS/val/MSLS_resnext_GeM_480_GCL_PCA_results_latex.csv};

\addplot[blue, mark size=2pt, thick , mark=star,mark options=solid, style=solid] table [x=k, y=r, col sep=comma] {plot_data/MSLS/val/Pitts_ResNet101_GeM_480_apgem_results_latex.csv};

\end{axis}

\end{tikzpicture}\label{fig:msls_recall_val}
}
\subfloat[MSLS test set]{
\begin{tikzpicture}[thick,scale=1, every node/.style={scale=.8}]
\begin{axis}
[xlabel=K,
legend pos =south east,grid, height=6cm, width=7.5cm]
\addplot[magenta, mark size=2pt, thick , mark=triangle,mark options=dashed, style=dashed] table [x=k, y=r, col sep=comma] {plot_data/MSLS/test/Pitts_vgg16_netvlad_64_TL_latex.csv};
\addplot[magenta, mark size=2pt, thick , mark=triangle,mark options=solid, style=solid] table [x=k, y=r, col sep=comma] {plot_data/MSLS/test/Pitts_vgg16_netvlad_64_TL_pca_latex.csv};
\addplot[black, mark size=2pt, thick , mark=pentagon,mark options=dashed, style=dashed] table [x=k, y=r, col sep=comma] {plot_data/MSLS/test/MSLS_vgg16_netvlad_16_TL_latex.csv};
\addplot[black, mark size=2pt, thick , mark=pentagon,mark options=solid, style=solid] table [x=k, y=r, col sep=comma] {plot_data/MSLS/test/MSLS_vgg16_netvlad_16_TL_pca_latex.csv};
\addplot[black, mark size=2pt, thick , mark=triangle,mark options=dashed, style=dashed] table [x=k, y=r, col sep=comma] {plot_data/MSLS/test/MSLS_vgg16_netvlad_64_TL_latex.csv};
\addplot[green, mark size=2pt, thick , mark=triangle,mark options=dashed, style=dashed] table [x=k, y=r, col sep=comma] {plot_data/MSLS/test/MSLS_vgg16_netvlad_64_SARE_latex.csv};
\addplot[red, mark size=2pt, thick , mark=triangle,mark options=dashed, style=dashed] table [x=k, y=r, col sep=comma] {plot_data/MSLS/test/MSLS_vgg16_GeM_480_GCL_results_latex.csv};
\addplot[red, mark size=2pt, thick , mark=triangle,mark options=solid, style=solid] table [x=k, y=r, col sep=comma] {plot_data/MSLS/test/MSLS_vgg16_GeM_480_GCL_PCA_results_latex.csv};
\addplot[red, mark size=2pt, thick , mark=square,mark options=dashed, style=dashed] table [x=k, y=r, col sep=comma] {plot_data/MSLS/test/MSLS_resnet50_GeM_480_GCL_results_latex.csv};
\addplot[red, mark size=2pt, thick , mark=square,mark options=solid, style=solid] table [x=k, y=r, col sep=comma] {plot_data/MSLS/test/MSLS_resnet50_GeM_480_GCL_PCA_results_latex.csv};
\addplot[red, mark size=2pt, thick , mark=*,mark options=dashed, style=dashed] table [x=k, y=r, col sep=comma] {plot_data/MSLS/test/MSLS_resnet152_GeM_480_GCL_results_latex.csv};
\addplot[red, mark size=2pt, thick , mark=*,mark options=solid, style=solid] table [x=k, y=r, col sep=comma] {plot_data/MSLS/test/MSLS_resnet152_GeM_480_GCL_PCA_results_latex.csv};
\addplot[red, mark size=2pt, thick , mark=diamond,mark options=dashed, style=dashed] table [x=k, y=r, col sep=comma] {plot_data/MSLS/test/MSLS_resnext_GeM_480_GCL_results_latex.csv};
\addplot[red, mark size=2pt, thick , mark=diamond,mark options=solid, style=solid] table [x=k, y=r, col sep=comma] {plot_data/MSLS/test/MSLS_resnext_GeM_480_GCL_PCA_results_latex.csv};

\addplot[blue, mark size=2pt, thick , mark=triangledown,mark options=solid, style=solid] table [x=k, y=r, col sep=comma] {plot_data/MSLS/test/Pitts_ResNet101_GeM_480_apgem_results_latex.csv};

\end{axis}
\end{tikzpicture}
\label{fig:msls_recall_test}
}
\subfloat{
\begin{minipage}[t]{.65\columnwidth}
\vspace{-5.15cm}
\begin{tikzpicture}[thick, scale=1, every node/.style={scale=.8}] 
    \begin{axis}[%
    hide axis,
    xmin=20,
    xmax=50,
    ymin=0,
    ymax=0.4,
    height=6cm, width=2cm,
    legend style={draw=white!15!black,legend cell align=left, font=\footnotesize}
    ]
\addlegendimage{magenta, mark size=2pt, thick , mark=triangle, style=dashed,mark options=dashed}
\addlegendentry{VGG-NetVLAD-TL-64-Pitts};
\addlegendimage{magenta, mark size=2pt, thick , mark=triangle, style=solid,mark options=solid}
\addlegendentry{VGG-NetVLAD-TL-64-Pitts-PCA$_w$};
\addlegendimage{black, mark size=2pt, thick , mark=pentagon, style=dashed,mark options=dashed}
\addlegendentry{VGG-NetVLAD-TL-16-MSLS};
\addlegendimage{black, mark size=2pt, thick , mark=pentagon, style=solid,mark options=solid}
\addlegendentry{VGG-NetVLAD-TL-16-MSLS-PCA$_w$};
\addlegendimage{black, mark size=2pt, thick , mark=triangle, style=dashed,mark options=dashed}
\addlegendentry{VGG-NetVLAD-TL-64-MSLS};
\addlegendimage{black, mark size=2pt, thick , mark=triangle, style=solid,mark options=solid}
\addlegendentry{VGG-NetVLAD-TL-64-MSLS-PCA$_w$};
\addlegendimage{green, mark size=2pt, thick , mark=triangle, style=solid,mark options=solid}
\addlegendentry{VGG-NetVLAD-SARE-64};

\addlegendimage{blue, mark size=2pt, thick , mark=star, style=solid,mark options=solid}
\addlegendentry{ResNet101-GeM-AP};

\addlegendimage{red, mark size=2pt, thick , mark=triangle, style=dashed,mark options=dashed}
\addlegendentry{VGG16-GeM-GCL};
\addlegendimage{red, mark size=2pt, thick , mark=triangle, style=solid,mark options=solid}
\addlegendentry{VGG16-GeM-GCL-PCA$_w$};
\addlegendimage{red, mark size=2pt, thick , mark=square, style=dashed,mark options=dashed}
\addlegendentry{ResNet50-GeM-GCL};
\addlegendimage{red, mark size=2pt, thick , mark=square, style=solid,mark options=solid}
\addlegendentry{ResNet50-GeM-GCL-PCA$_w$};
\addlegendimage{red, mark size=2pt, thick , mark=*, style=dashed,mark options=dashed}
\addlegendentry{ResNet152-GeM-GCL};
\addlegendimage{red, mark size=2pt, thick , mark=*, style=solid,mark options=solid}
\addlegendentry{ResNet152-GeM-GCL-PCA$_w$};
\addlegendimage{red, mark size=2pt, thick , mark=diamond, style=dashed,mark options=dashed}
\addlegendentry{ResNeXt-GeM-GCL};
\addlegendimage{red, mark size=2pt, thick , mark=diamond, style=solid,mark options=solid}
\addlegendentry{ResNeXt-GeM-GCL-PCA$_w$};
    \end{axis}
    \end{tikzpicture} 
\end{minipage}
}

   \caption{Comparison of the results achieved by our methods with those of state-of-the-art methods and models trained with a binary contrastive loss for (a) MSLS validation and (b) MSLS test. TL stands for models trained with the Triplet Loss, CL for Contrastive Loss and GCL for Generalized Contrastive Loss. \revised{The results obtained by our models are displayed in red, with a solid line for those trained with a GCL function, and a dashed one for those trained with a CL function. }}
       \label{fig:msls_recall}

\end{figure*}
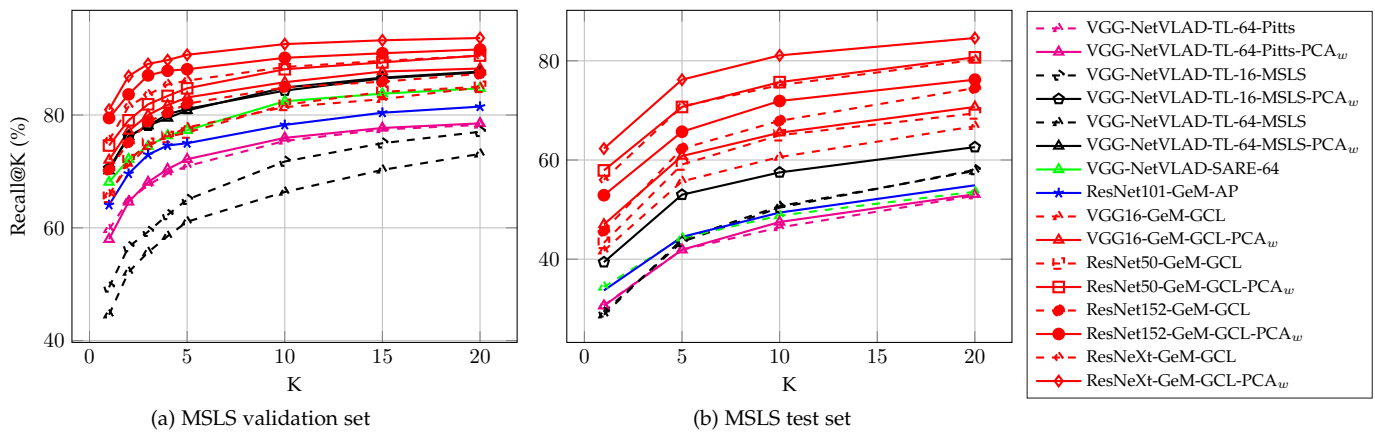

\subsection{Evaluation metrics}
We followed a widely used place recognition evaluation protocol and considered a query as correctly identified if any of the top-k retrieved images are annotated as a positive match~\cite{Sattler2012, Arandjelovic2017,msls}. We computed the following metrics.

\textbf{Top-k recall (R@k)} is the percentage of queries for which we retrieved at least a correct map image among their $k$ nearest neighbors. 

\revised{\textbf{Percentage of correctly localized queries} is the amount of images that are correctly retrieved for a given translation and rotation threshold. We use it for evaluating on the RobotCar Seasons v2 and the Extended CMU datasets.}

\textbf{Average Precision (AP),} an approximation of the area under the precision-recall curve for the classification of all the possible pairs of images. We use this measure for the  7Scenes dataset~\cite{Shotton2013}.

%% file: sections/6_results.tex
\label{sec:results}

\subsection{Large scale outdoor place recognition}
We trained several models on the MSLS training set and report results on the MSLS validation and test set. We evaluate the generalization capabilities of our models by also testing them on the Pittsburgh30k, Tokyo~24/7, \revised{RobotCar Seasons v2, and ExtendedCMU Seasons} datasets. 
In the following, we use the naming \mbox{\emph{backbone}-\emph{pooling}-\emph{loss}} for the considered models. For example, \mbox{ResNet50-GeM-GCL} indicates a ResNet50 backbone with GeM pooling trained using the Generalized Contrastive Loss function. 

\textbf{Mapillary Street Level Sequences.} We carried out experiments using \revised{four} backbones, namely \revised{VGG16}, ResNet50, ResNet152, and \revised{ResNeXt101-32x8d (hereinafter ResNeXt)}, and \revised{the GeM~\cite{radenovic2018fine}} global pooling layer. \revised{Additional results with an average global pooling layer are included in the supplementary material}. For each combination of backbone network and pooling layer, we optimize the models with the binary Contrastive Loss and with our Generalized Contrastive Loss. As shown in Table~\ref{tab:msls-ablation} and Fig.~\ref{fig:msls_recall}, the models trained with the GCL function consistently outperform their counterparts trained with the CL. For \revised{all models}, we also perform a postprocessing step consisting of PCA whitening and observe that it further boosts the performance. 

We compared the results of our models with those obtained by state-of-the-art methods\revised{, namely NetVLAD~\cite{Arandjelovic2017}, NetVLAD-SARE~\cite{liu2019stochastic}, AP-GeM~\cite{revaud2019learning} and Patch-NetVLAD~\cite{hausler2021patch}}. We achieved better results than existing approaches on the MSLS test set. We demonstrate that a simple\revised{r} architecture, \revised{consisting only of a convolutional backbone and a simple pooling layer arranged as a two-way siamese network,} trained using the proposed GCL loss function outperforms more complex architectures, which deploy triplet networks and VLAD layers. For instance, NetVLAD achieved a top-5 recall equal to \revised{44\%} on the MSLS test set~\cite{msls}\revised{, while our \mbox{VGG16-GeM-GCL} model, which uses the same backbone as NetVLAD, achieved a top-5 recall of 55.7\%. The superiority of our \mbox{VGG16-GeM-GCL} model holds also in the case we post-process the descriptors using PCA whitening.} For \mbox{ResNet50-GeM-GCL}, we obtained a top-5 recall equal to 59.2\%. Our \mbox{ResNet152-GeM-GCL} \revised{and our ResNeXt-GeM-GCL achieved a top-5 recall equal to 62.3\% and 70.8\%, respectively.} When whitening is applied, the performance of our models further improved, reaching a top-5 recall of \revised{60.8\% for \mbox{VGG-GeM-GCL},} 65.7\% for \mbox{ResNet50-GeM-GCL} 70.7\% for \mbox{ResNet152-GeM-GCL}, \revised{and 76.2\% for \mbox{ResNeXt-GeM-GCL}. 

Our results are higher than those of all the other methods on the MSLS dataset, including NetVLAD (with different configurations)~\cite{Arandjelovic2017}, NetVLAD-SARE~\cite{liu2019stochastic}, AP-GeM~\cite{revaud2019learning} and Patch-NetVLAD~\cite{hausler2021patch}. We used the models and code from the original papers, as also done in~\cite{benchmarking_ir3DV2020,hausler2021patch}}.
It is worth pointing out that our learned representations have a size of 2048 \revised{for the ResNet50, ResNet152 and ResNeXt models, and 512 for the VGG16}, while the NetVLAD vectors consist of 32768 features \revised{when using 64 clusters or 8192 when using 16 clusters}  \revised{before dimensionality reduction. When PCA whitening is applied, the NetVLAD-based representations are reduced to a size of 4096, while ours remain 512 for VGG16, 1024 for ResNet50 and ResNeXt, and 2048 for ResNet152}.

\revised{\textbf{Generalization to Pittsburgh\revised{30K}, Tokyo 24/7,  \revised{RobotCar-Seasons v2 and ExtendedCMU Seasons.}} In Table~\ref{tab:msls-ablation} and Table~\ref{tab:msls-sota}, we also report the results that our models achieved on the Pittsburgh30K, Tokyo 24/7, RobotCar Seasons v2 and Extended CMU Seasons datasets.}

\revised{We compare our results with those achieved by different configurations of NetVLAD, trained on Pittsburgh30k with 64 clusters and trained on MSLS with 16 and 64 clusters. Additionally, we compare to ResNet101-AP-GeM, NetVLAD-SARE and Patch-NetVLAD.} \revised{As shown in Table~\ref{tab:msls-sota}, }the models trained with the GCL function on the MSLS dataset generalize well to unseen datasets, in some cases better than existing methods\revised{, despite using descriptors of lower dimensionality}. \revised{In the case of Pittsburgh30k, our models achieve competitive results, up to 91.5\% top-5 recall, which is very close to the 91.8\% reported by NetVLAD-64 trained on the Pittsburgh30k dataset itself, and slightly lower than NetVLAD-SARE, which reaches a top-5 recall of 94.3\%, and Patch-NetVLAD, with a top-5 recall of 94.5\% } 

\revised{In the case of Tokyo24/7 we record a lower performance, which we attribute to the drastic variations in illumination included in the dataset, with images taken at day and at night. Such conditions are underrepresented in the MSLS training set: only 4k query images are taken at night, out of 500k query images that compose the dataset. However, we observe that applying a whitening transform greatly enhances the generalization capabilities of our models, and it boosts the performance for all datasets. On Tokyo 24/7, we achieved an improvement of 28.2\% on the top-5 recall for the ResNet152-GeM-GCL model, up to 81\%, which outperforms NetVLAD and AP-GeM. NetVLAD-SARE and Patch-NetVLAD generalize better to this dataset, with top-5 recall of 86.7\% and 88.6\% respectively. 
Our models also generalize well to urban localization datasets like RobotCar Seasons V2 and Extended CMU Seasons, achieving up to 21.9\% and 19\% of correctly localized queries within 0.5m and 5$\degree$, respectively. For both datasets, the results are comparable to those obtained by NetVLAD trained on MSLS, but are outperformed by NetVLAD trained on Pittsburgh, AP-GeM, NetVLAD-SARE and Patch-NetVLAD.
} \revised{Note that we do not perform 6DOF pose estimation, but estimate the pose of a query image by inheriting that of the best retrieved match. We thus do not aim at comparing with methods that perform refined pose estimation.}

\subsection{Small scale outdoor place recognition}
\begin{figure}[t!]
\subfloat{
\begin{minipage}[t]{\columnwidth}

\hspace{1.1cm}
\begin{tikzpicture}[thick, scale=1, every node/.style={scale=.8}] 
    \begin{axis}[%
    hide axis,
    xmin=10,
    xmax=50,
    ymin=0,
    ymax=0.4, height=6cm, width=\columnwidth,
    legend style={draw=white!15!black,legend columns=2,legend style={minimum width=.41\columnwidth}},
    legend cell align={left}
    ]
    
\addlegendimage{blue, thick, mark size=2pt, mark=square};
\addlegendentry{ResNet18-CL};

\addlegendimage{red, thick, mark size=2pt, mark=square};
\addlegendentry{ResNet18-GCL (ours)};
\addlegendimage{blue, thick, mark size=2pt, mark=o};
\addlegendentry{ResNet34-CL};

\addlegendimage{red, thick, mark size=2pt, mark=o};
\addlegendentry{ResNet34-GCL (ours)};

\addlegendimage{blue, thick, mark size=2pt, mark=diamond};
\addlegendentry{DenseNet161-CL};

\addlegendimage{red, thick, mark size=2pt, mark=diamond};
\addlegendentry{DenseNet161-GCL (ours)};
\addlegendimage{black, thick, mark size=2pt, mark=triangle};
\addlegendentry{NetVLAD off-the-shelf};

\end{axis}
\end{tikzpicture}
\end{minipage}
}
\vspace{-5pt}

\centering
\setcounter{subfigure}{0}
\subfloat[Map: W17, Query: W18]{
\begin{tikzpicture}[thick, scale=1, every node/.style={scale=.8}]
\begin{axis}
[xlabel=K,
ylabel=Recall@K (\%), ylabel style={at={(axis description cs:-0.08,0.5)}},grid, height=6cm, width=\columnwidth]
\addplot[black, thick, mark size=2pt, mark=triangle] table [x=k, y=r, col sep=comma] {plot_data/TB_Places/TB_Places_W18_W17_NetVLAD_recalls.csv};
\addplot[red, thick, mark size=2pt, mark=square] table [x=k, y=r, col sep=comma] {plot_data/TB_Places/TB_Places_ResNet18_avg_2last_soft_siamese_w18_w17_recall_toplot.txt};
\addplot[blue, thick, mark size=2pt, mark=square] table [x=k, y=r, col sep=comma] {plot_data/TB_Places/TB_Places_ResNet18_avg_2last_binary_siamese_w18_w17_recall_toplot.txt};

\addplot[red, thick, mark size=2pt, mark=o] table [x=k, y=r, col sep=comma] {plot_data/TB_Places/TB_Places_ResNet34_avg_2last_soft_siamese_w18_w17_recall_toplot.txt};
\addplot[blue, thick, mark size=2pt, mark=o] table [x=k, y=r, col sep=comma] {plot_data/TB_Places/TB_Places_ResNet34_avg_2last_binary_siamese_w18_w17_recall_toplot.txt};

\addplot[blue, thick, mark size=2pt, mark=diamond] table [x=k, y=r, col sep=comma] {plot_data/TB_Places/TB_Places_W18_W17_CAIP_recalls.csv};
\addplot[red, thick, mark size=2pt, mark=diamond] table [x=k, y=r, col sep=comma] {plot_data/TB_Places/TB_Places_W18_W17_Soft_recalls.csv};
\end{axis}
\end{tikzpicture}
    \label{fig:w17_w18_recall}
}
\\
\subfloat[Map: W18, Query: W18]{
\begin{tikzpicture}[thick, scale=1, every node/.style={scale=.8}]
\begin{axis}
[xlabel=K,
ylabel=Recall@K (\%), ylabel style={at={(axis description cs:-0.08,0.5)}},
legend pos =south east,grid, height=6cm, width=\columnwidth]
\addplot[black, thick, mark size=2pt, mark=triangle] table [x=k, y=r, col sep=comma] {plot_data/TB_Places/TB_Places_map_query_NetVLAD_recalls.csv};
\addplot[red, thick, mark size=2pt, mark=square] table [x=k, y=r, col sep=comma] {plot_data/TB_Places/TB_Places_ResNet18_avg_2last_soft_siamese_w18_map_query_recall_toplot.txt};
\addplot[blue, thick, mark size=2pt, mark=square] table [x=k, y=r, col sep=comma] {plot_data/TB_Places/TB_Places_ResNet18_avg_2last_binary_siamese_w18_map_query_recall_toplot.txt};

\addplot[red, thick, mark size=2pt, mark=o] table [x=k, y=r, col sep=comma] {plot_data/TB_Places/TB_Places_ResNet34_avg_2last_soft_siamese_w18_map_query_recall_toplot.txt};
\addplot[blue, thick, mark size=2pt, mark=o] table [x=k, y=r, col sep=comma] {plot_data/TB_Places/TB_Places_ResNet34_avg_2last_binary_siamese_w18_map_query_recall_toplot.txt};

\addplot[blue, thick, mark size=2pt, mark=diamond] table [x=k, y=r, col sep=comma] {plot_data/TB_Places/TB_Places_map_query_CAIP_recalls.csv};
\addplot[red, thick, mark size=2pt, mark=diamond] table [x=k, y=r, col sep=comma] {plot_data/TB_Places/TB_Places_map_query_Soft_recalls.csv};
\end{axis}
\end{tikzpicture}

    \label{fig:w18_map_query_recall}
}\\
   \caption{Comparison of the results for the TB-Places dataset. In (a) we report the results when using W17 as map and W18 as query set. In (b), we show the top-k recall achieved when dividing W18 into map and query. The results achieved by our method are shown in red, the results of the models trained with the binary Contrastive Loss are in blue, and the results of NetVLAD are displayed in black.}
       \label{fig:tb_places_recall}

\end{figure}
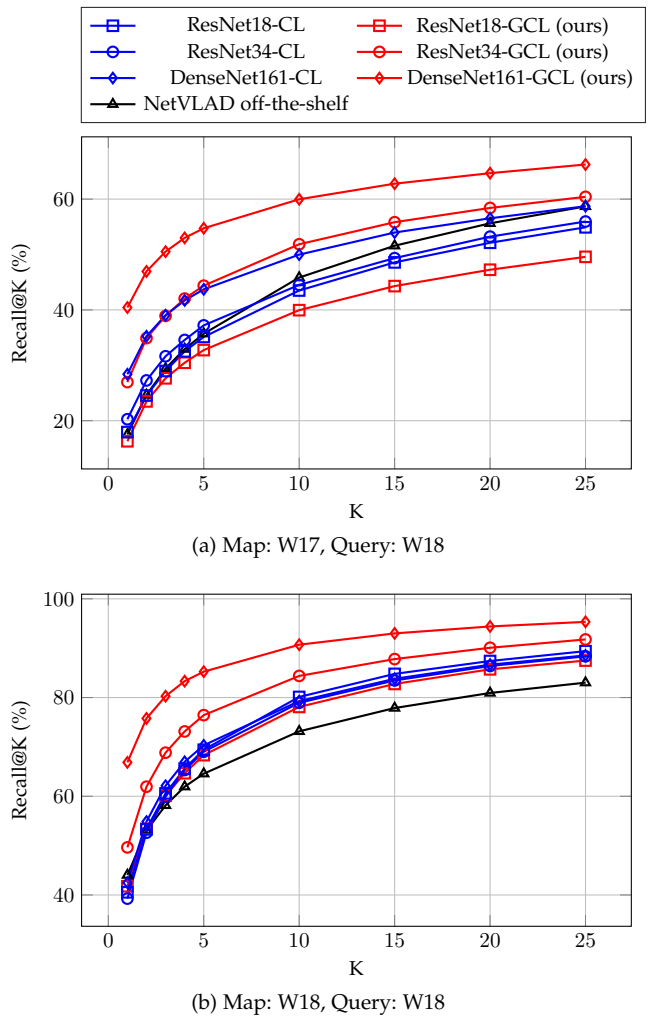

We trained our method on the TB-Places dataset and show the results that we achieved in Fig.~\ref{fig:tb_places_recall}. We considered three different backbones, namely ResNet18, ResNet34, and DenseNet161, which learn representations of size 512, 512 and 2208, respectively. We trained them with the binary Contrastive Loss function and with our Generalized Contrastive Loss function. Furthermore, we compare them with the NetVLAD off-the-shelf model. For these experiments we use only a Global Average Pooling, so this information is omitted from the names of the models.

We report results of two experiments. For the first one, we used the training set (W17) as map, and the W18 set as query. With this experiment we tested the strength of our learned representations against significant variations between the query and the map set. We show the results that we obtained in Fig.~\ref{fig:w17_w18_recall}. For the second experiment, we divided the W18 set into map and query, to test the generalization capabilities when both map and query sets are unknown to the place recognition model (i.e. not seen during training). The results are displayed in Fig.~\ref{fig:w18_map_query_recall}.
For both experiments the models trained with our GCL function and the proposed similarity ground truth consistently achieve better recall than the ones trained with the binary Contrastive Loss, with the exception of the ResNet18 model. In some cases the models trained with the binary ground truth do not outperform the NetVLAD model, even though this one is not trained on the TB-Places dataset.

\subsection{Indoor place recognition}
\label{sec:results7scenes}

\begin{figure*}[ht!]
\centering
\hspace{-5pt}
\subfloat[heads]{
\begin{tikzpicture}[thick, scale=1, every node/.style={scale=.8}]
\begin{axis}
[xlabel=K,
ylabel=Recall@K (\%), ylabel style={at={(axis description cs:-0.08,0.5)}},grid, height=4.75cm, width=.61\columnwidth,xtick distance=5, ytick distance=5]
\addplot[black, thick, mark size=2pt, mark=triangle] table [x=k, y=r, col sep=comma] {plot_data/7Scenes/PR_7scenes_recallatk_50FOV_heads_NetVLAD_recalls.csv};
\addplot[blue, thick, mark size=2pt, mark=o] table [x=k, y=r, col sep=comma] {plot_data/7Scenes/PR_7scenes_recallatk_50FOV_heads_ResNet18_crisp_recalls.csv};
\addplot[red, thick, mark size=2pt, mark=o] table [x=k, y=r, col sep=comma] {plot_data/7Scenes/PR_7scenes_recallatk_50FOV_heads_ResNet18_soft_recalls.csv};

\addplot[blue, thick, mark size=2pt, mark=square] table [x=k, y=r, col sep=comma] {plot_data/7Scenes/PR_7scenes_recallatk_50FOV_heads_ResNet34_crisp_recalls.csv};
\addplot[red, thick, mark size=2pt, mark=square] table [x=k, y=r, col sep=comma] {plot_data/7Scenes/PR_7scenes_recallatk_50FOV_heads_ResNet34_soft_recalls.csv};
\end{axis}
\end{tikzpicture}
    \label{fig:7Scenes_heads_recall}
}
\hspace{-7pt}~\subfloat[stairs]{
\begin{tikzpicture}[thick, scale=1, every node/.style={scale=.8}]
\begin{axis}
[xlabel=K,grid, height=4.75cm, width=.61\columnwidth,xtick distance=5, ytick distance=5]
\addplot[black, thick, mark size=2pt, mark=triangle] table [x=k, y=r, col sep=comma] {plot_data/7Scenes/PR_7scenes_recallatk_50FOV_stairs_NetVLAD_recalls.csv};
\addplot[blue, thick, mark size=2pt, mark=o] table [x=k, y=r, col sep=comma] {plot_data/7Scenes/PR_7scenes_recallatk_50FOV_stairs_ResNet18_crisp_recalls.csv};
\addplot[red, thick, mark size=2pt, mark=o] table [x=k, y=r, col sep=comma] {plot_data/7Scenes/PR_7scenes_recallatk_50FOV_stairs_ResNet18_soft_recalls.csv};
\addplot[blue, thick, mark size=2pt, mark=square] table [x=k, y=r, col sep=comma] {plot_data/7Scenes/PR_7scenes_recallatk_50FOV_stairs_ResNet34_crisp_recalls.csv};
\addplot[red, thick, mark size=2pt, mark=square] table [x=k, y=r, col sep=comma] {plot_data/7Scenes/PR_7scenes_recallatk_50FOV_stairs_ResNet34_soft_recalls.csv};
\end{axis}
\end{tikzpicture}
    \label{fig:7Scenes_stairs_recall}
}
\hspace{-7pt}~\subfloat[pumpkin]{
\begin{tikzpicture}[thick, scale=1, every node/.style={scale=.8}]
\begin{axis}
[xlabel=K,
grid, height=4.75cm, width=.61\columnwidth,xtick distance=5, ytick distance=5]
\addplot[black, thick, mark size=2pt, mark=triangle] table [x=k, y=r, col sep=comma] {plot_data/7Scenes/PR_7scenes_recallatk_50FOV_pumpkin_NetVLAD_recalls.csv};
\addplot[blue, thick, mark size=2pt, mark=o] table [x=k, y=r, col sep=comma] {plot_data/7Scenes/PR_7scenes_recallatk_50FOV_pumpkin_ResNet18_crisp_recalls.csv};
\addplot[red, thick, mark size=2pt, mark=o] table [x=k, y=r, col sep=comma] {plot_data/7Scenes/PR_7scenes_recallatk_50FOV_pumpkin_ResNet18_soft_recalls.csv};

\addplot[blue, thick, mark size=2pt, mark=square] table [x=k, y=r, col sep=comma] {plot_data/7Scenes/PR_7scenes_recallatk_50FOV_pumpkin_ResNet34_crisp_recalls.csv};
\addplot[red, thick, mark size=2pt, mark=square] table [x=k, y=r, col sep=comma] {plot_data/7Scenes/PR_7scenes_recallatk_50FOV_pumpkin_ResNet34_soft_recalls.csv};
\end{axis}
\end{tikzpicture}
    \label{fig:7Scenes_pumpkin_recall}
}
~\subfloat{
\begin{minipage}[c][2cm]{.47\columnwidth}
\vspace{-120pt}
\begin{tikzpicture}[thick, scale=1, every node/.style={scale=.8}] 
    \begin{axis}[%
    hide axis,
    xmin=10,
    xmax=70,
    ymin=0,
    ymax=0.4,
    legend pos =north west,
    legend style={draw=white!20!black,legend cell align=left}, height=4.75cm, width=.61\columnwidth
    ]
    \addlegendimage{black, thick, mark size=2pt, mark=triangle}
    \addlegendentry{NetVLAD off-the-shelf};
    \addlegendimage{blue, thick, mark size=2pt, mark=o}
    \addlegendentry{ResNet18-CL};
    \addlegendimage{red, thick, mark size=2pt, mark=o}
    \addlegendentry{ResNet18-GCL (ours)};
    \addlegendimage{blue, thick, mark size=2pt, mark=square}
    \addlegendentry{ResNet34-CL};
    \addlegendimage{red, thick, mark size=2pt, mark=square}
    \addlegendentry{ResNet34-GCL (ours)};
    \end{axis}
    \end{tikzpicture} 
    \end{minipage}
}
\setcounter{subfigure}{3}

\vspace{-5mm}
\subfloat[fire]{
\begin{tikzpicture}[thick, scale=1, every node/.style={scale=.8}]
\begin{axis}
[xlabel=K,
ylabel=Recall@K (\%), ylabel style={at={(axis description cs:-0.08,0.5)}},grid, height=4.75cm, width=.61\columnwidth,xtick distance=5, ytick distance=5]
\addplot[black, thick, mark size=2pt, mark=triangle] table [x=k, y=r, col sep=comma] {plot_data/7Scenes/PR_7scenes_recallatk_50FOV_fire_NetVLAD_recalls.csv};
\addplot[blue, thick, mark size=2pt, mark=o] table [x=k, y=r, col sep=comma] {plot_data/7Scenes/PR_7scenes_recallatk_50FOV_fire_ResNet18_crisp_recalls.csv};
\addplot[red, thick, mark size=2pt, mark=o] table [x=k, y=r, col sep=comma] {plot_data/7Scenes/PR_7scenes_recallatk_50FOV_fire_ResNet18_soft_recalls.csv};

\addplot[blue, thick, mark size=2pt, mark=square] table [x=k, y=r, col sep=comma] {plot_data/7Scenes/PR_7scenes_recallatk_50FOV_fire_ResNet34_crisp_recalls.csv};
\addplot[red, thick, mark size=2pt, mark=square] table [x=k, y=r, col sep=comma] {plot_data/7Scenes/PR_7scenes_recallatk_50FOV_fire_ResNet34_soft_recalls.csv};
\end{axis}
\end{tikzpicture}
    \label{fig:7Scenes_fire_recall}
}
\hspace{-7pt}~\subfloat[redkitchen]{
\begin{tikzpicture}[thick, scale=1, every node/.style={scale=.8}]
\begin{axis}
[xlabel=K,grid, height=4.75cm, width=.61\columnwidth,xtick distance=5, ytick distance=5]
\addplot[black, thick, mark size=2pt, mark=triangle] table [x=k, y=r, col sep=comma] {plot_data/7Scenes/PR_7scenes_recallatk_50FOV_redkitchen_NetVLAD_recalls.csv};
\addplot[blue, thick, mark size=2pt, mark=o] table [x=k, y=r, col sep=comma] {plot_data/7Scenes/PR_7scenes_recallatk_50FOV_redkitchen_ResNet18_crisp_recalls.csv};
\addplot[red, thick, mark size=2pt, mark=o] table [x=k, y=r, col sep=comma] {plot_data/7Scenes/PR_7scenes_recallatk_50FOV_redkitchen_ResNet18_soft_recalls.csv};

\addplot[blue, thick, mark size=2pt, mark=square] table [x=k, y=r, col sep=comma] {plot_data/7Scenes/PR_7scenes_recallatk_50FOV_redkitchen_ResNet34_crisp_recalls.csv};
\addplot[red, thick, mark size=2pt, mark=square] table [x=k, y=r, col sep=comma] {plot_data/7Scenes/PR_7scenes_recallatk_50FOV_redkitchen_ResNet34_soft_recalls.csv};
\end{axis}
\end{tikzpicture}
    \label{fig:7Scenes_redkitchen_recall}
}
\hspace{-7pt}~\subfloat[chess]{
\begin{tikzpicture}[thick, scale=1, every node/.style={scale=.8}]
\begin{axis}
[xlabel=K,grid, height=4.75cm, width=.61\columnwidth,xtick distance=5, ytick distance=2]
\addplot[black, thick, mark size=2pt, mark=triangle] table [x=k, y=r, col sep=comma] {plot_data/7Scenes/PR_7scenes_recallatk_50FOV_chess_NetVLAD_recalls.csv};
\addplot[blue, thick, mark size=2pt, mark=o] table [x=k, y=r, col sep=comma] {plot_data/7Scenes/PR_7scenes_recallatk_50FOV_chess_ResNet18_crisp_recalls.csv};
\addplot[red, thick, mark size=2pt, mark=o] table [x=k, y=r, col sep=comma] {plot_data/7Scenes/PR_7scenes_recallatk_50FOV_chess_ResNet18_soft_recalls.csv};

\addplot[blue, thick, mark size=2pt, mark=square] table [x=k, y=r, col sep=comma] {plot_data/7Scenes/PR_7scenes_recallatk_50FOV_chess_ResNet34_crisp_recalls.csv};
\addplot[red, thick, mark size=2pt, mark=square] table [x=k, y=r, col sep=comma] {plot_data/7Scenes/PR_7scenes_recallatk_50FOV_chess_ResNet34_soft_recalls.csv};
\end{axis}
\end{tikzpicture}
    \label{fig:7Scenes_chess_recall}
}
\hspace{-7pt}~\subfloat[office]{
\begin{tikzpicture}[thick, scale=1, every node/.style={scale=.8}]
\begin{axis}
[xlabel=K,grid, height=4.75cm, width=.61\columnwidth,xtick distance=5, ytick distance=5]
\addplot[black, thick, mark size=2pt, mark=triangle] table [x=k, y=r, col sep=comma] {plot_data/7Scenes/PR_7scenes_recallatk_50FOV_office_NetVLAD_recalls.csv};
\addplot[blue, thick, mark size=2pt, mark=o] table [x=k, y=r, col sep=comma] {plot_data/7Scenes/PR_7scenes_recallatk_50FOV_office_ResNet18_crisp_recalls.csv};
\addplot[red, thick, mark size=2pt, mark=o] table [x=k, y=r, col sep=comma] {plot_data/7Scenes/PR_7scenes_recallatk_50FOV_office_ResNet18_soft_recalls.csv};

\addplot[blue, thick, mark size=2pt, mark=square] table [x=k, y=r, col sep=comma] {plot_data/7Scenes/PR_7scenes_recallatk_50FOV_office_ResNet34_crisp_recalls.csv};
\addplot[red, thick, mark size=2pt, mark=square] table [x=k, y=r, col sep=comma] {plot_data/7Scenes/PR_7scenes_recallatk_50FOV_office_ResNet34_soft_recalls.csv};
\end{axis}
\end{tikzpicture}
    \label{fig:7Scenes_office_recall}
}
\caption{Recall@K results achieved on the 7 Scenes dataset. The results of the models trained with our Generalized Contrastive Loss are shown in red, while those of the models trained with the binary Contrastive Loss are shown in blue. The Recall@K achieved by NetVLAD off-the-shelf is plotted in black.}
\label{fig:recallat_k_7scenes}
\end{figure*}
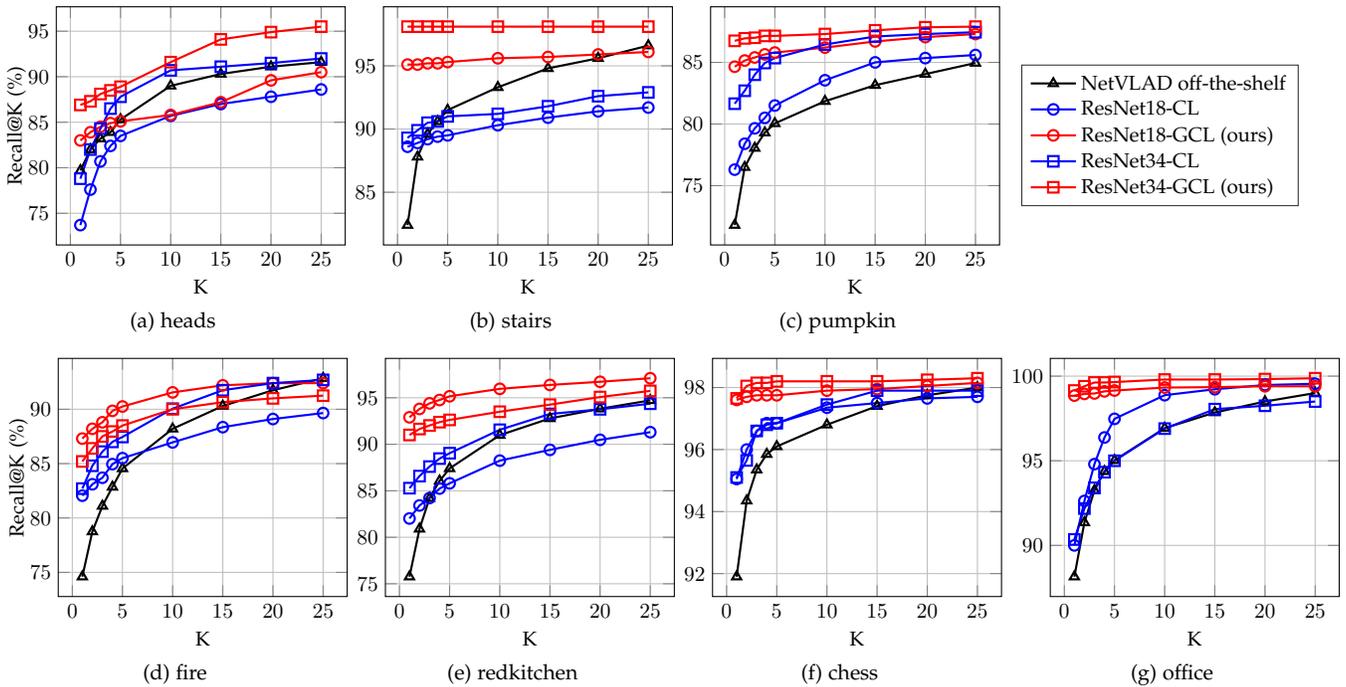

We report the results that we achieved on the seven sets of the 7scenes dataset in Fig.~\ref{fig:recallat_k_7scenes}. We used the ResNet18 and ResNet34 architectures as backbones with a Global Average Pooling layer and we compare them to NetVLAD off-the-shelf.  Since we only use one type of pooling for this experiments this information is omitted from the model names. We achieved generally higher Recall@K results for the models trained using the GCL function, for all scenes. The cases of the \emph{stairs}  (Fig.~\ref{fig:7Scenes_stairs_recall}), \emph{chess}  (Fig.~\ref{fig:7Scenes_chess_recall}) and \emph{office} (Fig.~\ref{fig:7Scenes_office_recall}) scenes are particularly interesting, since with the GCL descriptors we are able to retrieve positive matches for nearly all the query images, with a top-5 recall for \mbox{ResNet34-GCL} of 98.1\%, 98.2\%  and 99.7\%, respectively.

Furthermore, we report the average precision results in Table~\ref{tab:ap-7scenes}. This measure is an approximation of the area under the precision-recall curve and measures the number of pairs that can be correctly discriminated into positive and negative on the whole dataset. The models trained with the GCL function achieved higher AP than their corresponding models trained with the binary CL function. We achieved an average AP equal to 0.89 using the ResNet34-GCL model. \revised{Following the general expectation that improved retrieval results would contribute to a better localization, we performed a small-scale localization experiment to compare the contribution of the GCL and CL training to the localization performance of an InLoc pipeline~\cite{taira2018inloc}. We report the results in the supplementary materials.}



\subsection{Discussion}
\label{sec:discussion}
\begin{table}[t!]
\centering
\caption{Average Precision results obtained by the networks trained with the proposed Generalized Contrastive loss function on the 7Scenes  dataset, compared with those achieved by the same network architectures trained using the binary Contrastive loss function and by the NetVLAD off-the-shelf model. }
\label{tab:ap-7scenes}

\renewcommand{\arraystretch}{1.0}
\begin{tabular}{@{\extracolsep{4pt}}lccccc@{}}
\toprule
\textbf{}   & \textbf{NetVLAD}       & \multicolumn{2}{c}{\textbf{ResNet18}} & \multicolumn{2}{c}{\textbf{ResNet34}} \\ 
\cmidrule{2-2}
\cmidrule{3-4}
\cmidrule{5-6}
\textbf{Scene}       & \textbf{off-the-shelf} & \textbf{CL}       & \textbf{GCL}               & \textbf{CL}       & \textbf{GCL}             \\ \midrule 
Heads      & 0.587        & 0.739   & 0.807           & 0.759   & \textbf{0.853}   \\
Stairs     & 0.533        & 0.855   & 0.883            & 0.884   & \textbf{0.944}   \\
Pumpkin    & 0.491        & 0.768   & 0.849            & 0.782   & \textbf{0.914}   \\
Fire       & 0.539        & 0.786   & \textbf{0.811}   & 0.796   & 0.803            \\
Redkitchen & 0.439        & 0.790   & 0.876            & 0.782   & \textbf{0.902}   \\
Chess      & 0.645        & 0.943   & 0.964 & 0.945   & \textbf{0.974}   \\
Office     & 0.399        & 0.794   & 0.890            & 0.802   & \textbf{0.896}   \\ \midrule
\textit{Mean }      & 0.519        & 0.811   & 0.868            & 0.821   & \textbf{0.898}   \\ \bottomrule
\end{tabular}
\end{table}



State-of-the-art visual place recognition pipelines rely on difficult pair and triplet mining strategies for training~\cite{Arandjelovic2017,msls}. Their goal is to compose batches of training images that substantially contribute to the learning process, evaluating their potential impact on the loss function, at the cost of GPU memory and computing time. As we discussed in Section~\ref{sec:training_pairs}, we opted for a na\"{i}ve mining strategy, based only on the annotated graded similarity of image pairs and not on their embedded representations. 

\begin{figure}[t!]
     \centering
     \subfloat[Composition A]{\includegraphics[width=.5\columnwidth]{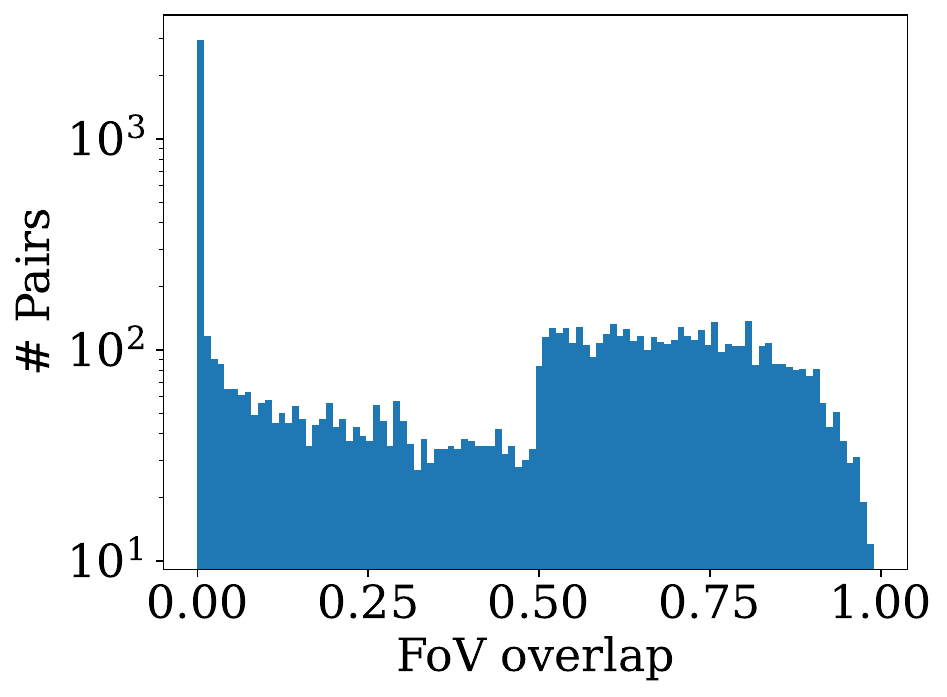}\label{fig:pairs_50_25_25}}
      \subfloat[Composition B]{\includegraphics[width=.5\columnwidth]{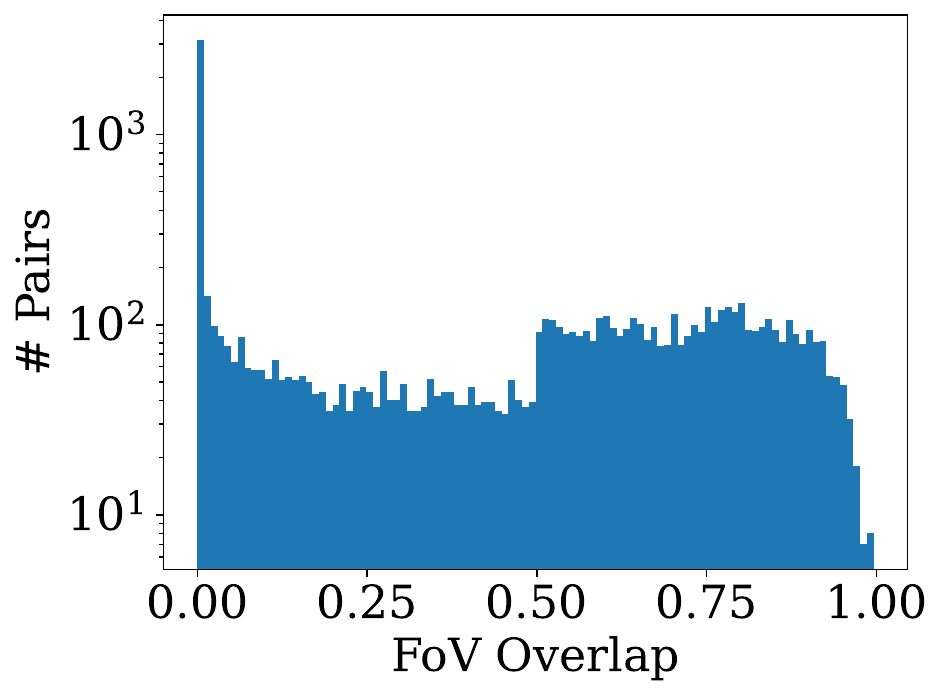}\label{fig:pairs_25_25_25_25}}\\
       \subfloat[Composition C]{\includegraphics[width=.5\columnwidth]{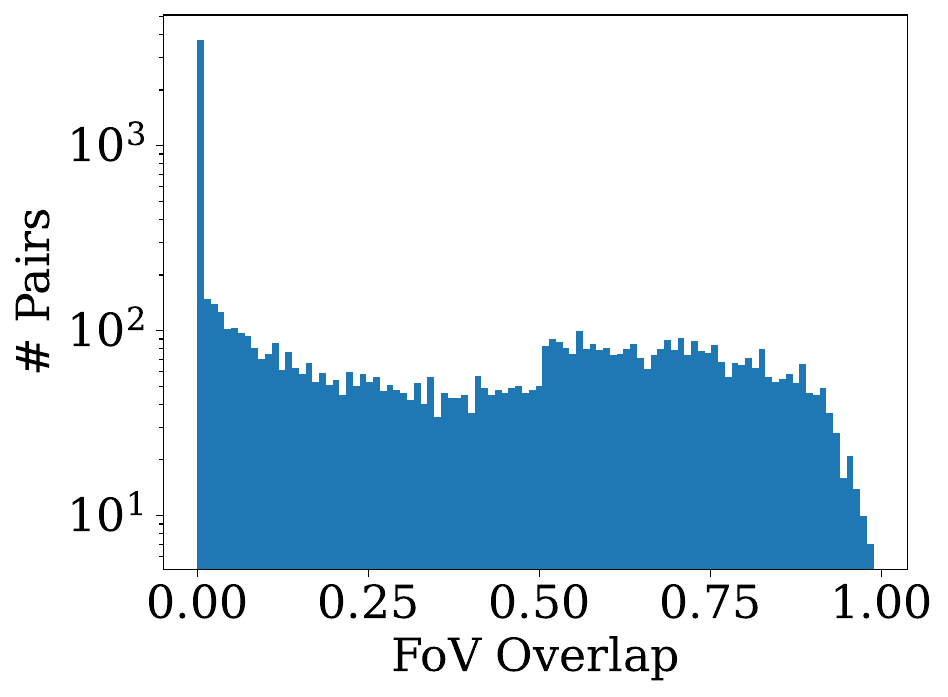}\label{fig:pairs_33_33_33}}
       \subfloat[Composition D]{\includegraphics[width=.5\columnwidth]{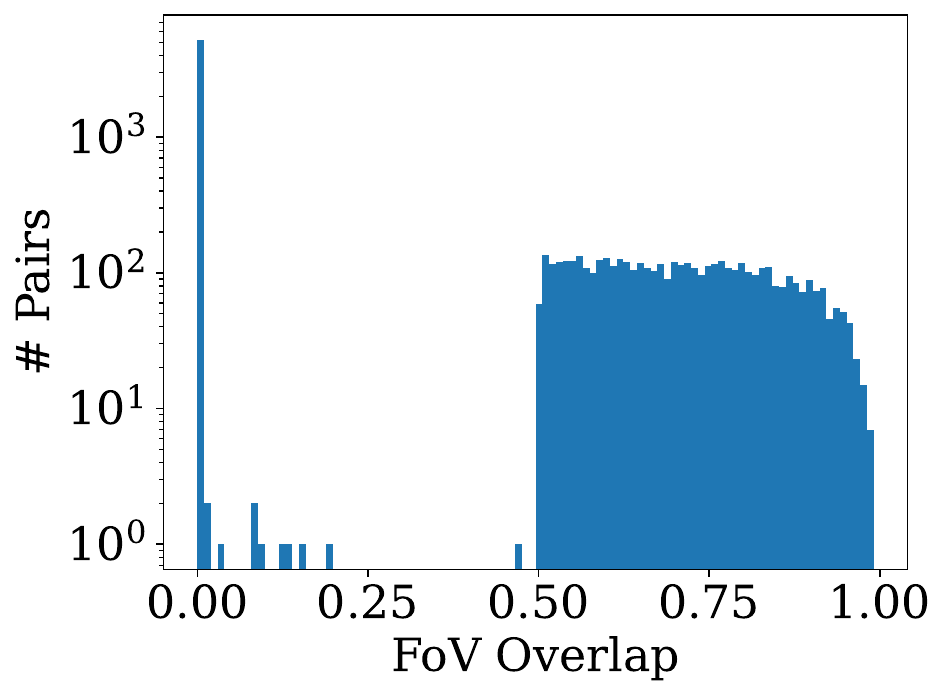}\label{fig:pairs_50_50}}
     \caption{Similarity ground truth distribution for 10000 randomly selected pairs in the MSLS train set when using different batch composition strategies. The vertical axis is in log scale.}
     \label{fig:batch_composition}
\end{figure}

\revised{
\textbf{Pair selection in na\"ive mining.} The composition of the training batches, in terms of distribution of image pairs with different ground truth graded similarity, resulting from the na\"ive mining strategy plays an important role in the effective convergence of the training process. 
In order to test the influence that different pair sampling strategies have on the effectiveness of the training process and accuracy of retrieval results, we carried out several experiments (using the four considered backbones), for which we report results in Table~\ref{tab:batch}. We thus defined four strategies of na\"ive mining for the composition of training batches, which we refer to as strategy A, B, C and D. 
For strategy A, we compose a training batch including 50\% of positive pairs ($\psi \in [0.5, 1]$), 25\% of soft negative pairs ($\psi \in (0, 0.5)$) and 25\% of hard negative pairs ($\psi=0$). For strategy B, we select 25\% hard positive pairs ($\psi \in [0.75, 1]$), 25\% soft positive pairs ($\psi \in [0.5, 0.75)$), 25\% of soft negative pairs ($\psi \in (0, 0.5)$) and 25\% of hard negative pairs ($\psi=0$). For strategy C, we prepare the batches by selecting 33.3\% positive pairs ($\psi \in [0.5, 1]$), 33.3\% of soft negative pairs ($\psi \in (0, 0.5)$) and 33.3\% of hard negative pairs ($\psi=0$). Finally, for strategy D we randomly sample 50\% positive pairs ($\psi \in [0.5, 1]$) and 50\% negative pairs ($\psi \in [0, 0.5)$). 
In Fig.~\ref{fig:batch_composition}, we show histograms of the pair similarity distribution of the composition of training batches resulting from the four considered na\"ive mining strategies.

The results in Table~\ref{tab:batch} clearly indicate that the most important factor to consider when composing the training batches is to select an adequate number of soft negative pairs, i.e. at least 25\% of the pairs included in a batch have similarity $\psi \in (0, 0.5)$, as done for strategy A, B and C. 
For strategy D, we can observe an improvement with respect to the binary similarity case (i.e. with only pairs with similarity $\psi=1$ and $\psi=0$, respectively), due to the use of soft positive samples. However, the use of soft- and hard positive pairs (i.e. pairs with similarity $\psi \in (0.5, 1]$) together with hard negatives only in strategy D does not guarantee optimal performance. Due to the very large amount of hard positive pairs in the dataset, sampling randomly among negatives does ensure the presence of soft negative pairs in the training batches and results in a non-uniform distribution of pair similarity (see Fig.~\ref{fig:batch_composition}d). The generalized contrastive optimization that we proposed, instead, largely benefits from a fairly uniform distribution of soft positive and soft negative pairs in the composition of the training batches.}

\begin{table}[t!]
\centering
\revised{\caption{Recall@5 in the MSLS validation and test sets for GCL models when using different batch composition strategies. Strategy A is 50\% $\psi \in [0.5, 1]$, 25\% $\psi \in (0, 0.5)$ and 25\% $\psi=0$. B is 25\% $\psi \in [0.75, 1]$, 25\% $\psi \in [0.5, 0.75)$, 25\% $\psi \in (0, 0.5)$ and 25\% $\psi=0$. For C, we have 33.3\% $\psi \in [0.5, 1]$, 33.3\% $\psi \in (0, 0.5)$ and 33.3\% $\psi=0$. Finally, for strategy D we have 50\% $\psi \in [0.5, 1]$ and 50\% $\psi \in [0, 0.5)$. }

\label{tab:batch}
\begin{tabular}{@{}llcc@{}}
\toprule
Model & Strategy & Val & Test \\ \midrule
\multirow{4}{*}{VGG-GeM-GCL} & A & 77.8 & \textbf{55.7} \\
 & B  & \textbf{78.8} & 54.9 \\
 & C & 78.2 & 56 \\
 & D & 73.4 & 49.3 \\ \midrule
\multirow{4}{*}{ResNet50-GeM-GCL} & A & \textbf{78.9} & \textbf{59.1} \\
 & B & 76.5 & 54.5 \\
 & C & 75.7 & 52.4 \\
 & D & 75.4 &  54.7\\\midrule
\multirow{4}{*}{ResNet152-GeM-GCL} & A & \textbf{82} & \textbf{62.3} \\
 & B & 80.4 & 59.7 \\
 & C & 78.6 & 58.4 \\
 & D & 78.5 &  64.8\\ \midrule
\multirow{4}{*}{Resnext-GeM-GCL} & A & 86.1 & 70.8 \\
 & B & 86.2 & 70.7 \\
 & C & \textbf{87.3} & \textbf{72.4} \\
 & D & 81.9 &  64.8\\ \bottomrule
\end{tabular}%

}
\end{table}

We have demonstrated that a na\"{i}ve mining strategy can be as effective as or better than a more complex one, provided that the ground truth is informative enough. The results that we achieved are attributable to the use of the proposed GCL function, which allows a graded similarity ground truth to be considered. Furthermore, since our na\"{i}ve strategy is much less resource demanding, it is possible to use it while training big CNN backbones in their entirety, and with batch size up to 512.
\begin{table}[t]
\centering
\caption{Results achieved on the MSLS validation set by our models trained by backpropagating the gradient only partially through the network. The \emph{last trained layer} column indicates the block of the backbone until which the gradient is backpropagated.}
\label{tab:partialtraining}
\begin{tabular}{@{}lcccccc@{}}
\toprule
                            & \multicolumn{3}{c}{\textbf{ResNet50-GeM}}  & \multicolumn{3}{c}{\textbf{ResNet152-GeM}}    \\ 
\textbf{Last trained layer} & \textbf{R@1} & \textbf{R@5} & \textbf{R@10} & \textbf{R@1} & \textbf{R@5} & \textbf{R@10} \\\midrule
Off-the-shelf  & 27.2 &   41.9   &  47.4   & 26.5 & 39.9   &  44.5    \\
Layer4  & 50.1 &   64.6   &     70 & 40.9& 56.4& 63.8    \\
Layer3       & \textbf{66.2}    & \textbf{78.9} & \textbf{81.9}  & \textbf{70.3}  & \textbf{82} & \textbf{84.9} \\ Layer2   & 60.3 &   74.2   &   77.4    & 65.4& 77.4& 80      \\     
Full        & 65      & 75.8 & 81.1 & 66.6 &  79.2  &  81.8     \\
\bottomrule

\end{tabular}

\end{table}

\revised{\textbf{Partial training. }}We studied the effects that training different layers of the \revised{ResNet50 and ResNet152} backbones with the GCL function has on the performance of our models. We report the results that we achieved on the MSLS validation set in Table~\ref{tab:partialtraining}. The models pre-trained on the ImageNet dataset~\cite{deng2009imagenet} achieved reasonable results. However, we obtained the highest results when the networks were further trained on the MSLS training set until Layer 3 (i.e. updating the weights of the two last convolutional blocks). Training more layers leads to overfitting. 

\revised{\textbf{Whitening and dimensionality reduction.} }The learned representations are of a relatively small size (2048 for ResNet50, ResNet152~\revised{and ResNeXt, 512 for VGG16}) compared to NetVLAD (size of 32768). The dimensionality of the latent space representation influences the performance of image retrieval and place recognition systems. Techniques for dimensionality reduction have been explored~\cite{Arandjelovic2017, radenovic2018fine}, usually combined with whitening, which has been demonstrated to have positive effects on image retrieval, especially in the case of CNN-learned representations~\cite{babenko2015aggregating, sharif2014cnn}. It can be learned end-to-end or used as a postprocessing step.

We evaluated the effect of whitening and dimensionality reduction on our learned descriptors by using PCA as postprocessing. We compute the PCA components on the descriptors of the map images and use them to project both map and query image descriptors. We perform experiments with PCA whitening with different number of dimensions, from 32 to 2048. We applied this strategy to the \revised{VGG16-GeM-GCL, }ResNet50-GeM-GCL, ResNet152-GeM-GCL \revised{ and ResNeXt-GeM-GCL} models (trained on MSLS) and \revised{tested} them on the \revised{validation and test sets of MSLS and the test sets of Pittsburgh30k and Tokyo 24/7}. We observe that PCA whitening further improves the performance of the image representations learned by our models and boosts their generalization capabilities. As shown in Fig.~\ref{fig:whitening}, the whitened descriptors with higher dimensionality achieve \revised{better} results on the considered datasets\revised{. It is worth noting that the ResNeXt-GeM-GCL maintains the same performance when reduced down to 256 dimensions, and that VGG16-GeM-GCL reaches peak top-5 recall when reduced to 256 features}. When using the whitening transform combined with PCA dimensionality reduction down to 128 dimensions of the latent space, our models achieved results comparable to those of \revised{models with full-size dimensions (2048 for ResNet and ResNeXt, and 512 for VGG)}, which outperform their non-whithened counterpart on all the considered datasets. We observed up to a 28.2\% of improvement in the case of Tokyo 24/7 and 12.8\% on the MSLS validation set.

\revised{
\textbf{Training time.} The proposed GCL function and the partial similarity labels contribute to training effective models in a much more efficient way than NetVLAD-based ones, as shown in Table~\ref{tab:training-time}. The training time per epoch of our models is less than that of NetVLAD, as we do not perform hard-pair mining. Furthermore, we train our GCL models for only one epoch on the MSLS dataset, while NetVLAD-16 and NetVLAD-64 converged after 22 and 7 epochs, respectively. 
}

%% file: sections/8_conclusions.tex
\label{sec:conclusions}
We presented a novel Generalized Contrastive Loss function to effectively train \revised{two-ways} siamese networks for image retrieval and visual place recognition. For this purpose we defined continuous measures of the degree of image similarity. We \revised{implemented} three techniques for the estimation of the degree of similarity of pairs of images, based on geometric information such as GPS and compass angle, 6DOF camera pose, and scene 3D reconstruction. We deployed them to re-annotate the MSLS, TB-Places and 7Scenes datasets, providing image pairs with graded similarity, rather than binary labels. The re-annotation process is automatic, thus not requiring any human intervention. 

The CNN architectures that we trained consist of a fully convolutional backbone and a global pooling layer (i.e. global average pooling or GeM). The networks that we optimized using the proposed Generalized Contrastive loss function and the graded similarity label sets consistently achieved higher results (top-5 recall up to $18\%$ higher) than their counterparts trained with a binary Contrastive Loss function on the MSLS, TB-Places and 7Scenes datasets. On the MSLS dataset our \revised{ResNeXt-GeM-GCL} network achieved a top-5 recall of \revised{$76.2\%$}, outperforming NetVLAD, \revised{NetVLAD-SARE, AP-GeM and Patch-NetVLAD (top-5 recall of $58.8\%$, $44.3\%$, $44.5\%$, $57.6\%$, respectively)}, and establishing a new state-of-the-art result. Furthermore, as opposed to the case of NetVLAD\revised{-based models}, the training procedure that we deploy does not require a complex pair mining process: we only ensure that each batch contains approximately the same number of positive and negative pairs.  Our model also reaches competitive performance \revised{by generalizing well to} the Pittsburgh\revised{30k, Tokyo 24/7, RobotCar Seasons v2 and Extended CMU Seasons} datasets.

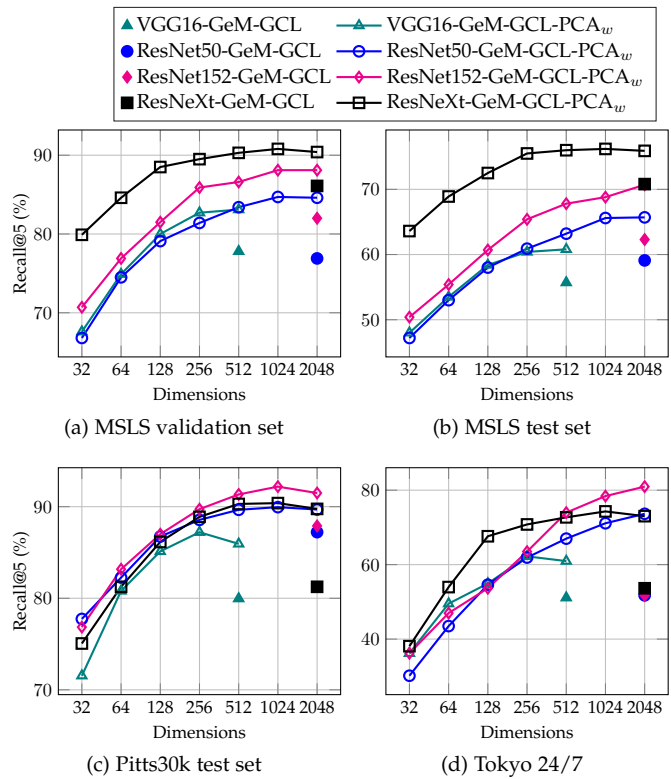
\begin{figure}
    \centering
\subfloat{
\centering
\begin{minipage}[c]{.65\columnwidth}
\begin{tikzpicture}[thick, scale=1, every node/.style={scale=.8}] 
    \begin{axis}[%
    hide axis,
    xmin=20,
    xmax=50,
    ymin=0,
    ymax=0.4,
    legend style={draw=white!15!black,legend cell align=left,legend columns=2}, height=4.75cm, width=.61\columnwidth
    ]
\addlegendimage{teal,thick, mark=triangle*, mark size=2pt, only marks}
\addlegendentry{VGG16-GeM-GCL};
\addlegendimage{teal,thick, mark=triangle, mark size=2pt}
\addlegendentry{VGG16-GeM-GCL-PCA$_w$};

\addlegendimage{blue,thick, mark=*, mark size=2pt, only marks}
\addlegendentry{ResNet50-GeM-GCL};
\addlegendimage{blue,thick, mark=o, mark size=2pt}
\addlegendentry{ResNet50-GeM-GCL-PCA$_w$};
\addlegendimage{magenta,thick, mark=diamond*, mark size=2pt, only marks}
\addlegendentry{ResNet152-GeM-GCL};
\addlegendimage{magenta,thick, mark=diamond, mark size=2pt}
\addlegendentry{ResNet152-GeM-GCL-PCA$_w$};

\addlegendimage{black,thick, mark=square*, mark size=2pt, only marks}
\addlegendentry{ResNeXt-GeM-GCL};
\addlegendimage{black,thick, mark=square, mark size=2pt}
\addlegendentry{ResNeXt-GeM-GCL-PCA$_w$};

    \end{axis}
    \end{tikzpicture} 
    \end{minipage}
}
\vspace{-7pt}
\setcounter{subfigure}{0}

    \subfloat[MSLS validation set]{
\begin{tikzpicture}[thick, scale=1, every node/.style={scale=.7}]
\begin{axis}
[xlabel=Dimensions,
ylabel=Recall@5 (\%), ylabel style={at={(axis description cs:-0.08,0.5)}},grid, xmode=log,log basis x={2}, xtick={32,64,128,256, 512,1024,2048}, xticklabels={32,64,128,256, 512,1024,2048}, width=.6\columnwidth
]
\addplot[teal,thick, mark=triangle*, mark size=2pt] table [x=d, y=r, col sep=comma] {plot_data/whitening/MSLS/MSLS_soft_vgg16_GeM_L2_0-1_2_nowhiten_toplot};
\addplot[teal, mark=triangle, thick, mark size=2pt] table [x=d, y=r, col sep=comma] {plot_data/whitening/MSLS/MSLS_soft_vgg16_GeM_L2_0-1_2_whiten_toplot};
\addplot[blue,thick, mark=*, mark size=2pt] table [x=d, y=r, col sep=comma] {plot_data/whitening/MSLS/MSLS_soft_ResNet50_GeM_L2_0-1_2_nowhiten_toplot};
\addplot[blue, mark=o, thick, mark size=2pt] table [x=d, y=r, col sep=comma] {plot_data/whitening/MSLS/MSLS_soft_ResNet50_GeM_L2_0-1_2_whiten_toplot};
\addplot[magenta,thick, mark=diamond*, mark size=2pt] table [x=d, y=r, col sep=comma] {plot_data/whitening/MSLS/MSLS_soft_ResNet152_GeM_L2_0-1_2_nowhiten_toplot};
\addplot[magenta, mark=diamond, thick, mark size=2pt] table [x=d, y=r, col sep=comma] {plot_data/whitening/MSLS/MSLS_soft_ResNet152_GeM_L2_0-1_2_whiten_toplot};
\addplot[black,thick, mark=square*, mark size=2pt] table [x=d, y=r, col sep=comma] {plot_data/whitening/MSLS/MSLS_soft_resnext_GeM_L2_0-1_2_nowhiten_toplot};
\addplot[black, mark=square, thick, mark size=2pt] table [x=d, y=r, col sep=comma] {plot_data/whitening/MSLS/MSLS_soft_resnext_GeM_L2_0-1_2_whiten_toplot};
\end{axis}
 \label{fig:whitening_msls_val}
\end{tikzpicture}
   
}
    \subfloat[MSLS test set]{
\begin{tikzpicture}[thick, scale=1, every node/.style={scale=.7}]
\begin{axis}
[xlabel=Dimensions, ylabel style={at={(axis description cs:-0.08,0.5)}},grid, xmode=log,log basis x={2}, xtick={32,64,128,256, 512,1024,2048}, xticklabels={32,64,128,256, 512,1024,2048}, width=.6\columnwidth
]
\addplot[teal,thick, mark=triangle*, mark size=2pt] table [x=d, y=r, col sep=comma] {plot_data/whitening/MSLS_test/MSLS_soft_vgg16_GeM_L2_0-1_2_nowhiten_toplot.txt};
\addplot[teal, mark=triangle, thick, mark size=2pt] table [x=d, y=r, col sep=comma] {plot_data/whitening/MSLS_test/MSLS_soft_vgg16_GeM_L2_0-1_2_whiten_toplot.txt};
\addplot[blue,thick, mark=*, mark size=2pt] table [x=d, y=r, col sep=comma] {plot_data/whitening/MSLS_test/MSLS_soft_ResNet50_GeM_L2_0-1_2_nowhiten_toplot.txt};
\addplot[blue, mark=o, thick, mark size=2pt] table [x=d, y=r, col sep=comma] {plot_data/whitening/MSLS_test/MSLS_soft_ResNet50_GeM_L2_0-1_2_whiten_toplot.txt};
\addplot[magenta,thick, mark=diamond*, mark size=2pt] table [x=d, y=r, col sep=comma] {plot_data/whitening/MSLS_test/MSLS_soft_ResNet152_GeM_L2_0-1_2_nowhiten_toplot.txt};
\addplot[magenta, mark=diamond, thick, mark size=2pt] table [x=d, y=r, col sep=comma] {plot_data/whitening/MSLS_test/MSLS_soft_ResNet152_GeM_L2_0-1_2_whiten_toplot.txt};
\addplot[black,thick, mark=square*, mark size=2pt] table [x=d, y=r, col sep=comma] {plot_data/whitening/MSLS_test/MSLS_soft_resnext_GeM_L2_0-1_2_nowhiten_toplot.txt};
\addplot[black, mark=square, thick, mark size=2pt] table [x=d, y=r, col sep=comma] {plot_data/whitening/MSLS_test/MSLS_soft_resnext_GeM_L2_0-1_2_whiten_toplot.txt};
\end{axis}
 \label{fig:whitening_msls_test}
\end{tikzpicture}
   
}
\\
\subfloat[Pitts30k test set]{
\begin{tikzpicture}[thick, scale=1, every node/.style={scale=.7}]
\begin{axis}
[xlabel=Dimensions,
ylabel=Recall@5 (\%), ylabel style={at={(axis description cs:-0.08,0.5)}},grid, xmode=log,log basis x={2}, xtick={32,64,128,256, 512,1024,2048}, xticklabels={32,64,128,256, 512,1024,2048}, width=.6\columnwidth
]
\addplot[teal,thick, mark=triangle*, mark size=2pt] table [x=d, y=r, col sep=comma] {plot_data/whitening/Pitts30k/MSLS_soft_vgg16_GeM_L2_0-1_2_nowhiten_toplot};
\addplot[teal, mark=triangle, thick, mark size=2pt] table [x=d, y=r, col sep=comma] {plot_data/whitening/Pitts30k/MSLS_soft_vgg16_GeM_L2_0-1_2_whiten_toplot};
\addplot[blue,thick, mark=*, mark size=2pt] table [x=d, y=r, col sep=comma] {plot_data/whitening/Pitts30k/MSLS_soft_ResNet50_GeM_L2_0-1_2_nowhiten_toplot};
\addplot[blue, mark=o, thick, mark size=2pt] table [x=d, y=r, col sep=comma] {plot_data/whitening/Pitts30k/MSLS_soft_ResNet50_GeM_L2_0-1_2_whiten_toplot};
\addplot[magenta,thick, mark=diamond*, mark size=2pt] table [x=d, y=r, col sep=comma] {plot_data/whitening/Pitts30k/MSLS_soft_ResNet152_GeM_L2_0-1_2_nowhiten_toplot};
\addplot[magenta, mark=diamond, thick, mark size=2pt] table [x=d, y=r, col sep=comma] {plot_data/whitening/Pitts30k/MSLS_soft_ResNet152_GeM_L2_0-1_2_whiten_toplot};
\addplot[black,thick, mark=square*, mark size=2pt] table [x=d, y=r, col sep=comma] {plot_data/whitening/Pitts30k/MSLS_soft_resnext_GeM_L2_0-1_2_nowhiten_toplot};
\addplot[black, mark=square, thick, mark size=2pt] table [x=d, y=r, col sep=comma] {plot_data/whitening/Pitts30k/MSLS_soft_resnext_GeM_L2_0-1_2_whiten_toplot};
\end{axis}
 \label{fig:whitening_msls_val}
\end{tikzpicture}
}
\subfloat[Tokyo 24/7]{
\begin{tikzpicture}[thick, scale=1, every node/.style={scale=.7}]
\begin{axis}
[xlabel=Dimensions, ylabel style={at={(axis description cs:-0.08,0.5)}},grid, xmode=log,log basis x={2}, xtick={32,64,128,256, 512,1024,2048}, xticklabels={32,64,128,256, 512,1024,2048}, width=.6\columnwidth
]
\addplot[teal,thick, mark=triangle*, mark size=2pt] table [x=d, y=r, col sep=comma] {plot_data/whitening/Tokyo247/MSLS_soft_vgg16_GeM_L2_0-1_2_nowhiten_toplot};
\addplot[teal, mark=triangle, thick, mark size=2pt] table [x=d, y=r, col sep=comma] {plot_data/whitening/Tokyo247/MSLS_soft_vgg16_GeM_L2_0-1_2_whiten_toplot};
\addplot[blue,thick, mark=*, mark size=2pt] table [x=d, y=r, col sep=comma] {plot_data/whitening/Tokyo247/MSLS_soft_ResNet50_GeM_L2_0-1_2_nowhiten_toplot};
\addplot[blue, mark=o, thick, mark size=2pt] table [x=d, y=r, col sep=comma] {plot_data/whitening/Tokyo247/MSLS_soft_ResNet50_GeM_L2_0-1_2_whiten_toplot};
\addplot[magenta,thick, mark=diamond*, mark size=2pt] table [x=d, y=r, col sep=comma] {plot_data/whitening/Tokyo247/MSLS_soft_ResNet152_GeM_L2_0-1_2_nowhiten_toplot};
\addplot[black,thick, mark=square*, mark size=2pt] table [x=d, y=r, col sep=comma] {plot_data/whitening/Tokyo247/MSLS_soft_resnext_GeM_L2_0-1_2_nowhiten_toplot};

\addplot[magenta, mark=diamond, thick, mark size=2pt] table [x=d, y=r, col sep=comma] {plot_data/whitening/Tokyo247/MSLS_soft_ResNet152_GeM_L2_0-1_2_whiten_toplot};

\addplot[black, mark=square, thick, mark size=2pt] table [x=d, y=r, col sep=comma] {plot_data/whitening/Tokyo247/MSLS_soft_resnext_GeM_L2_0-1_2_whiten_toplot};
\end{axis}
\label{fig:whitening_tokyo247}
\end{tikzpicture}

}

    \caption{Results obtained on the MSLS \revised{validation, MSLS test}, Pittsburgh\revised{30k and} Tokyo 24/7 datasets by our models with and without PCA withening. For all the datasets, reducing the dimensionality of the latent space vectors and applying the whitening transform contribute to an increase of the retrieval performance.}
    \label{fig:whitening}
\end{figure}

\begin{table}[t!]
\centering
\revised{\caption{Training time on MSLS using one NVIDIA V100 GPU. In the second column, the number in parenthesis is the amount of time a model took to converge. The total training time we report is until convergence.}
\label{tab:training-time}
\resizebox{\columnwidth}{!}{%
\begin{tabular}{@{}lccc@{}}
\toprule
\textbf{Model} & \textbf{ Epochs} & \textbf{Time per epoch} & \textbf{Total training time} \\ \midrule
NetVLAD-16 & 30 (22) & 24h & 22d \\
NetVLAD-64 & 10 (7) & 36h & 10.5d \\
VGG16-GeM-GCL & 1 & 5h & 5h \\
ResNet50-GeM-GCL & 1 & 6h & 6h \\
ResNet152-GeM-GCL & 1 & 14h & 14h \\
ResNetXt-GeM-GCL & 1 (1/2) & 28h & 14h \\ \bottomrule
\end{tabular}%
}}
\end{table}

%% file: sections/bios.tex
\begin{IEEEbiography}[{\includegraphics[width=1in,height=1.25in,clip,keepaspectratio]{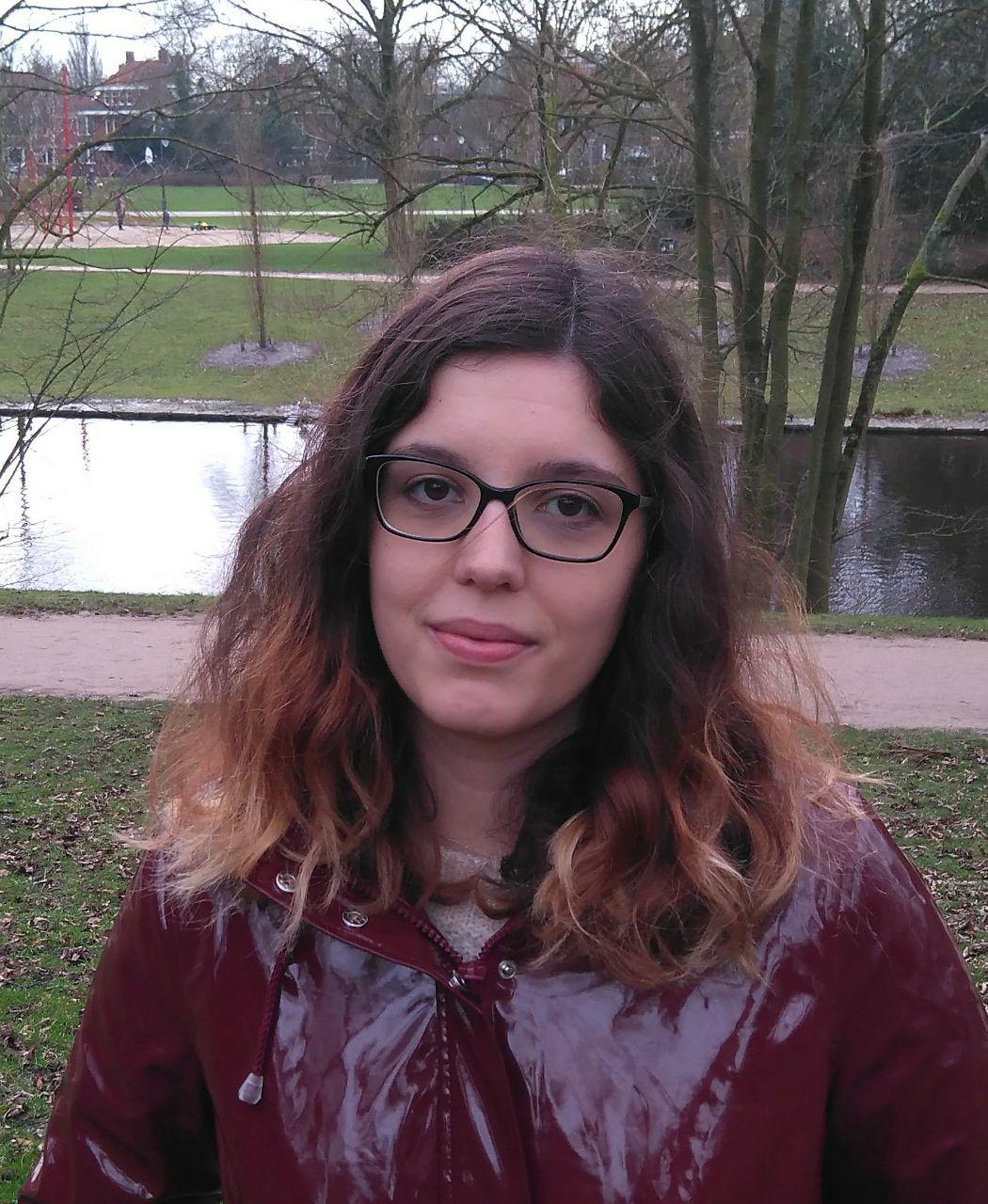}}]{Mar\'ia Leyva-Vallina} received her BSc in Software Engineering from the University of Oviedo in 2015 and her MSc in Artificial Intelligence from the Polytechnic University of Catalonia in 2017. She is currently pursuing her PhD with the Intelligent Systems group in the University of Groningen. Her main research interests are in representation learning for computer vision.\end{IEEEbiography}

\begin{IEEEbiography}[{\includegraphics[width=1in,height=1.25in,clip,keepaspectratio]{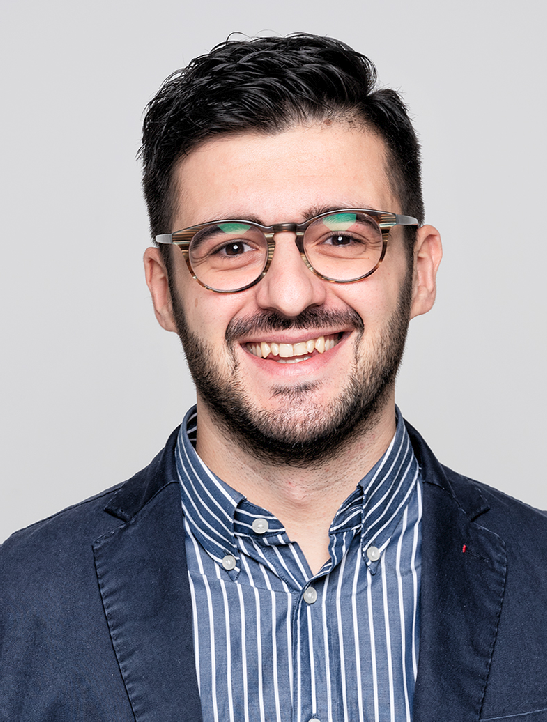}}]{Nicola Strisciuglio}
received the Ph.D. degree (cum laude) in Computer Science from the University of Groningen (Netherlands) and the Ph.D. degree in Information
Engineering from the University of Salerno (Italy).  
He is currently an Assistant Professor at the Faculty of Electrical Engineering, Mathematics and Computer Science, University of Twente (Netherlands). He has been the
General Co-Chair of the 1st, 2nd and 3rd International Conference on Applications of Intelligent Systems (APPIS). His research interests include machine learning, signal processing and computer vision.
\end{IEEEbiography}


\begin{IEEEbiography}[{\includegraphics[width=1in,height=1.25in,clip,keepaspectratio]{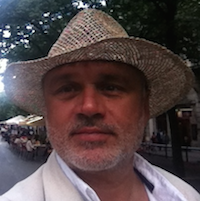}}]{Nicolai Petkov} received the Dr. sc. techn. degree in computer engineering (informationstechnik) from the Dresden University of Technology, Dresden, Germany. Since 1991, he has been a Professor of computer science and the Head of the Intelligent Systems Group, University of Groningen. He has authored two monographs, and has authored or co-authored over 150 scientific papers. He holds four patents. His current research interests include pattern recognition, machine learning, data analytics, and brain-inspired computing, with applications in various areas. He is a member of the editorial boards of several journals.

\end{IEEEbiography}

%% file: sections/appendix_extra_results_msls.tex
\section{Additional results}
In the following, we report extra results and observations on the performance of the proposed methods. In particular, we show how the Generalized Contrastive Loss function contributes to learn representations that better characterize the relevant parts of the input images for a robust computation of their similarity. Furthermore, we presents insights about how the GCL function contributes to a better regularization of the learned latent space.

\revised{We studied the effect of a different pooling layer by training our models on the MSLS dataset using a Global Average Pooling layer and report the results that we obtained.  
Furthermore, we evaluate our MSLS models on the Pittsburgh250k~\cite{Torii-PAMI2015} and TokyoTM~\cite{Torii-PAMI2015} benchmarks, and provide detailed results for the Extended CMU Seasons~\cite{sattler2018benchmarking} and RobotCar Seasons V2 datasets~\cite{sattler2018benchmarking}, divided by environment and weather condition. 
We also provide a comparison with the work of Thoma et al~\cite{thoma2020soft} on the CMU Seasons dataset. 
}

We study the effect that different threshold values for the ground truth similarity associated to the images have on the performance of our models. We consider two ground truth thresholds: one based on the GPS distance in meters between the locations in which images were taken, and another based on the annotated degree of similarity, $\psi$. 
 \revised{Finally, we provide extended results on the 7Scenes dataset, including the precision recall curves and a small test to evaluate our descriptors in a visual localization pipeline.}

\begin{figure}[!b]
    \centering
\subfloat[ResNet50-GeM-CL]{
\label{fig:tsneCL}
\includegraphics[width=.45\columnwidth]{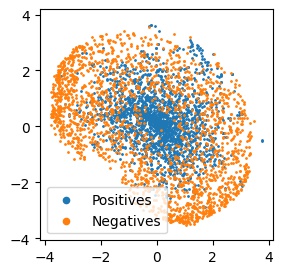}}~\subfloat[ResNet50-GeM-GCL]{
\label{fig:tsneGCL}
\includegraphics[width=.45\columnwidth]{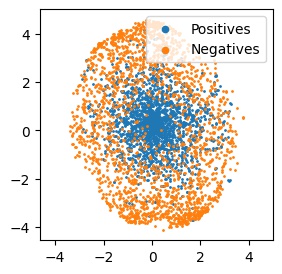}}
    \caption{Visualization of the learned embedded space. We selected 1000 random positive pairs and 1000 random negative pairs from the MSLS Copenhagen set, computed the differences between their representations and projected them into a 2D space using T-SNE.}
    \label{fig:tsne}
\end{figure}

\subsection{Learned latent space }

In Fig.~\ref{fig:tsne}, we show the 2D projection of the vectors representing the difference of the latent space representation of 2000 image pairs (1000 positive and 1000 negative) randomly selected from the Copenhagen set of the MSLS dataset. For each pair, we compute the difference between the map and the query image latent representation. We use this as input to the t-SNE algorithm~\cite{tsne}, which projects the representations onto a 2D space. We visualize the vectors produced by two models with a ResNet50-GeM backbone, one trained using the CL function (Fig.~\ref{fig:tsneCL}) and the other using the GCL (Fig.~\ref{fig:tsneGCL}) function. The effect of the proposed GCL function is evident in the better regularized latent space, where the representation of similar image pairs (blue dots) are more consistently distributed towards the center of the space. The representations learned using the CL function, instead, form a more scattered and noisy distribution.

\begin{figure*}[!t]
    \centering
    {\fontsize{8pt}{11pt}
    \def\svgwidth{\textwidth}
      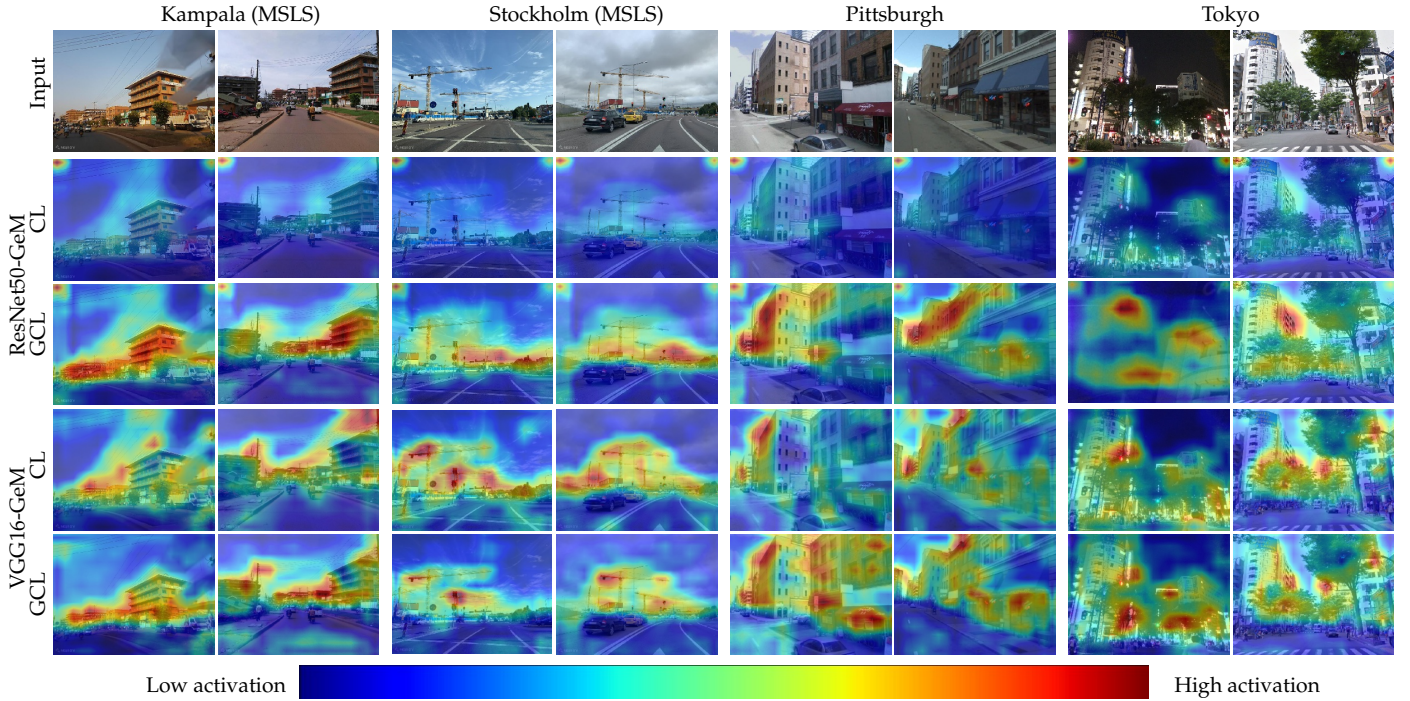}
    \caption{CNN activations for the ResNet50-GeM and the VGG16-GeM models with CL and GCL for several input image pairs. The first two pairs, corresponding to the first four columns, are part of the MSLS test set. The third and fourth belong to the Pittsburgh30k and Tokyo 24/7 test set, respectively. We show the activations for the last layer of the backbone overlapped with the input images.}
    \label{fig:activations}
\end{figure*}

\begin{table*}[t!]
\centering
\caption{Ablation study on the considered datasets: all the models are trained on the MSLS train set and deploy a global average pooling layer. CL stands for Contrastive Loss and while GCL for our Generalized Contrastive Loss. For the cases in which PCA whitening is applied we report the dimensionality that achieves the best results on the MSLS validation set.}
\label{tab:ablation-avg}
\resizebox{\textwidth}{!}{%
\setlength\tabcolsep{1.5pt}
\begin{tabular}
{@{}lcccccccccccccccccccc@{}}
\toprule
\textbf{} & \textbf{} & \textbf{} & \multicolumn{3}{c}{\textbf{MSLS-Val}} & \multicolumn{3}{c}{\textbf{MSLS-Test}} & \multicolumn{3}{c}{\textbf{Pittsburgh30k}} & \multicolumn{3}{c}{\textbf{Tokyo24/7}} & \multicolumn{3}{c}{\textbf{RobotCar Seasons V2}} & \multicolumn{3}{l}{\textbf{Extended CMU Seasons}} \\
\textbf{Method} & \textbf{PCA$_w$} & \textbf{Dim} & \textbf{R@1} & \textbf{R@5} & \textbf{R@10} & \textbf{R@1} & \textbf{R@5} & \textbf{R@10} & \textbf{R@1} & \textbf{R@5} & \textbf{R@10} & \textbf{R@1} & \textbf{R@5} & \textbf{R@10} & \textbf{0.25m/2$\degree$} & \textbf{0.5m/5$\degree$} & \textbf{5.0m/10$\degree$} & \textbf{0.25m/2$\degree$} & \textbf{0.5m/5$\degree$} & \textbf{5.0m/10$\degree$} \\ \midrule
VGG-avg-CL & No & 512 & 28.8 & 47.0 & 53.9 & 16.9 & 30.5 & 36.4 & 28.9 & 53.7 & 64.8 & 10.5 & 24.1 & 35.6 & 1.0 & 6.4 & 34.4 & 0.9 & 3.0 & 22.7 \\
VGG-avg-GCL & No & 512 & 48.8 & 67.8 & 73.1 & 21.5 & 31.4 & 37.9 & 42.1 & 66.9 & 77.3 & 20.0 & 42.5 & 52.1 & 2.3 & 12.5 & 51.7 & 2.3 & 7.2 & 43.3 \\
VGG-avg-CL & Yes & 128 & 35.0 & 53.4 & 60.1 & 35.2 & 47.3 & 54.1 & 47.1 & 69.5 & 78.3 & 16.2 & 28.9 & 40.0 & 2.3 & 10.0 & 42.0 & 2.1 & 6.6 & 34.9 \\
VGG-avg-GCL & Yes & 256 & 54.5 & 72.6 & 78.2 & 32.9 & 49.0 & 56.5 & 56.2 & 76.7 & 83.9 & 28.6 & 45.7 & 54.9 & 3.6 & 15.3 & 57.1 & 3.7 & 11.2 & 52.5 \\ \midrule
ResNet50-avg-CL & No & 2048 & 44.3 & 60.3 & 65.9 & 24.9 & 39.0 & 44.6 & 54.0 & 75.7 & 83.1 & 20.6 & 40.0 & 50.2 & 3.1 & 12.6 & 51.0 & 2.6 & 7.8 & 43.4 \\
ResNet50-avg-GCL & No & 2048 & 59.6 & 72.3 & 76.2 & 35.8 & 52.0 & 59.0 & 62.5 & 82.7 & 88.4 & 24.1 & 44.1 & 54.6 & 2.9 & 12.9 & 56.3 & 3.1 & 9.7 & 55.1 \\
ResNet50-avg-CL & Yes & 2048 & 58.8 & 71.4 & 75.8 & 33.1 & 46.5 & 53.3 & 65.8 & 82.6 & 88.2 & 48.6 & 63.2 & 70.5 & 5.9 & 22.8 & 62.7 & 4.7 & 13.8 & 50.8 \\
ResNet50-avg-GCL & Yes & 1024 & 69.5 & 81.2 & 85.5 & 44.2 & 57.8 & 63.4 & 73.3 & 87.1 & 91.2 & 52.1 & 68.9 & 72.7 & 5.9 & 23.1 & 69.6 & 5.4 & 16.2 & 66.5 \\ \midrule
ResNet152-avg-CL & No & 2048 & 53.1 & 70.1 & 75.4 & 29.7 & 44.2 & 51.3 & 59.7 & 80.3 & 87.0 & 27.0 & 48.6 & 58.4 & 2.5 & 13.1 & 56.6 & 3.0 & 9.2 & 49.9 \\
ResNet152-avg-GCL & No & 2048 & 65.1 & 80.0 & 83.8 & 43.5 & 59.2 & 65.2 & 69.3 & 87.2 & 91.3 & 32.1 & 52.1 & 62.2 & 3.3 & 13.5 & 59.9 & 3.6 & 11.0 & 61.2 \\
ResNet152-avg-CL & Yes & 1024 & 63.0 & 77.7 & 81.5 & 37.7 & 51.6 & 56.9 & 68.8 & 85.9 & 90.4 & 49.8 & 67.3 & 74.3 & 5.4 & 22.2 & 64.8 & 4.8 & 14.3 & 59.9 \\
ResNet152-avg-GCL & Yes & 2048 & 75.8 & 87.4 & 89.7 & 52.7 & 68.1 & 74.2 & \textbf{77.9} & \textbf{90.4} & \textbf{93.5} & \textbf{64.4} & \textbf{77.8} & \textbf{83.2} & \textbf{6.2} & \textbf{23.9} & 70.0 & \textbf{5.7} & \textbf{17.0} & 66.5 \\ \midrule
ResNeXt-avg-CL & No & 2048 & 58.9 & 75.1 & 79.9 & 34.5 & 50.1 & 57.7 & 51.3 & 73.6 & 81.9 & 24.8 & 47.3 & 56.8 & 2.7 & 10.9 & 61.0 & 2.0 & 6.1 & 40.0 \\
ResNeXt-avg-GCL & No & 2048 & 72.2 & 85.1 & 87.3 & 51.5 & 66.9 & 71.7 & 62.9 & 81.0 & 87.1 & 39.4 & 58.1 & 68.9 & 2.7 & 12.4 & 63.1 & 3.3 & 10.2 & 57.7 \\
ResNeXt-avg-CL & Yes & 1024 & 71.6 & 84.7 & 88.0 & 46.5 & 62.9 & 68.9 & 69.2 & 85.3 & 89.6 & 44.8 & 63.5 & 73.6 & 4.8 & 20.4 & \textbf{70.6} & 4.6 & 13.4 & 58.6 \\
ResNeXt-avg-GCL & Yes & 1024 & \textbf{79.3} & \textbf{89.2} & \textbf{90.3} & \textbf{57.8} & \textbf{72.3} & \textbf{77.1} & 74.8 & 88.2 & 91.8 & 53.0 & 76.2 & 80.6 & 2.7 & 10.9 & 61.0 & 5.6 & 16.6 & \textbf{70.7} \\ \bottomrule

\end{tabular}%
}
\end{table*}

\subsection{Network activation}

In Fig.~\ref{fig:activations}, we show the activation maps of the last convolutional layer of our models with a \revised{VGG16-GeM and a} ResNet50-GeM backbone, both trained using the Contrastive Loss function (CL) and the proposed Generalized Contrastive Loss (GCL) function.  We selected two example image pairs from the MSLS test set~\cite{msls} corresponding to the cities of Kampala and Stockholm, one from the Pittsburgh30k test set~\cite{Arandjelovic2017}, and one from the Tokyo 24/7 dataset~\cite{Torii-CVPR2013}. For all cases, we observed that the model trained with the GCL function produces higher activation for the common visual features of the images, and lower for the irrelevant parts (i.e. the road or the sky), in contrast to the model trained with the binary CL, which focuses less in the concerned areas of the pictures. In the example from Stockholm we can observe that our model does not respond to the cars (which vary from picture to picture), while it does respond strongly to the cranes (which are a more permanent feature). The example from Tokyo 24/7 is also particularly interesting: our model trained with the GCL function has high responses on the common parts of the images even under big changes of illumination. \revised{We observed that ResNet architectures tend to produce a peak of activation on the top left corner of the images. This does not seem to occur with the VGG architecture, so our intuition is that this artifact is due to the residuals. }

\revised{\subsection{Results with Global Average Pooling}
In addition to using GeM, we also trained the considered backbones (i.e. VGG16, ResNet50, ResNet152 and ResNeXt) using a Global Average Pooling on the MSLS dataset. We show the results in Table~\ref{tab:ablation-avg}. We observed that our method can reach good results with an even simpler pooling layer as is a Global Average Pooling, although a GeM layer usually leads to better results (see main paper). Our methods reaches good results on the validation and test sets of MSLS and generalizes well to unseen datasets such as Pittsburgh30k~\cite{Arandjelovic2017}, Tokyo24/7~\cite{Arandjelovic2017}, RobotCar Seasons V2~\cite{sattler2018benchmarking} and Extended CMU Seasons~\cite{sattler2018benchmarking}. As we observed also with the GeM models, the Global Average Pooling models can reach better performance when PCA whitening is applied, up to a 72.3\% top-5 recall on the test set of MSLS.
}

\begin{table}[t!]
\centering
\caption{Generalization results of the models trained on the MSLS dataset for the Pittsburgh250k and TokyoTM benchmarks.}
\label{tab:ablation-pitts250k}
\resizebox{\columnwidth}{!}{%
\begin{tabular}{@{}lcccccccc@{}}
\toprule
\textbf{} & \textbf{} & \textbf{} & \multicolumn{3}{c}{\textbf{Pittsburgh250k}} & \multicolumn{3}{c}{\textbf{TokyoTM}} \\ 
\textbf{Method} & \textbf{PCA$_w$} & \textbf{Dim} & \textbf{R@1} & \textbf{R@5} & \textbf{R@10} & \textbf{R@1} & \textbf{R@5} & \textbf{R@10} \\\midrule
 VGG-avg-CL & No & 512 & 20.2 & 38.0 & 47.2 & 38.9 & 58.5 & 67.2 \\
VGG-avg-GCL & No & 512 & 32.1 & 53.4 & 62.7 & 65.6 & 80.8 & 85.2 \\
VGG-avg-CL & Yes & 128 & 39.3 & 58.9 & 67.4 & 66.0 & 80.3 & 85.0 \\
VGG-avg-GCL & Yes & 256 & 51.3 & 71.0 & 78.0 & 78.6 & 87.7 & 90.8 \\
VGG-GeM-CL & No & 512 & 44.5 & 63.1 & 70.1 & 67.7 & 80.8 & 85.0 \\
VGG-GeM-GCL & No & 512 & 53.3 & 72.4 & 79.2 & 75.5 & 85.4 & 88.6 \\
VGG-GeM-CL & Yes & 512 & 65.1 & 81.2 & 85.8 & 83.7 & 91.1 & 93.3 \\
VGG-GeM-GCL & Yes & 512 & 73.4 & 86.4 & 89.9 & 88.2 & 93.2 & 94.7 \\\midrule
ResNet50-avg-CL & No & 2048 & 45.1 & 65.8 & 73.6 & 69.7 & 83.0 & 87.5 \\
ResNet50-avg-GCL & No & 2048 & 56.2 & 77.1 & 83.8 & 72.3 & 84.8 & 88.4 \\
ResNet50-avg-CL & Yes & 2048 & 70.6 & 85.4 & 89.5 & 90.5 & 94.8 & 96.0 \\
ResNet50-avg-GCL & Yes & 1024 & 74.6 & 87.9 & 91.6 & 91.7 & 95.7 & 96.8 \\
ResNet50-GeM-CL & No & 2048 & 54.8 & 74.2 & 80.7 & 73.3 & 85.4 & 89.2 \\
ResNet50-GeM-GCL & No & 2048 & 68.2 & 84.6 & 89.2 & 80.1 & 88.8 & 91.6 \\
ResNet50-GeM-CL & Yes & 1024 & 72.4 & 86.6 & 90.4 & 88.7 & 93.9 & 95.4 \\
ResNet50-GeM-GCL & Yes & 1024 & 80.9 & 91.4 & 94.3 & 92.2 & 95.6 & 96.8 \\\midrule
ResNet152-avg-CL & No & 2048 & 51.3 & 73.0 & 80.1 & 73.5 & 86.4 & 89.9 \\
ResNet152-avg-GCL & No & 2048 & 64.0 & 83.6 & 89.2 & 78.9 & 89.0 & 91.9 \\
ResNet152-avg-CL & Yes & 1024 & 69.6 & 85.9 & 90.6 & 90.7 & 95.1 & 96.3 \\
ResNet152-avg-GCL & Yes & 2048 & 80.9 & 92.2 & 95.3 & \textbf{94.0} & \textbf{96.7} & \textbf{97.3} \\
ResNet152-GeM-CL & No & 2048 & 60.7 & 79.0 & 85.1 & 77.8 & 88.3 & 91.2 \\
ResNet152-GeM-GCL & No & 2048 & 68.0 & 84.9 & 89.8 & 81.5 & 90.3 & 92.8 \\
ResNet152-GeM-CL & Yes & 2048 & 76.2 & 89.9 & 93.4 & 91.1 & 95.3 & 96.4 \\
ResNet152-GeM-GCL & Yes & 2048 & \textbf{83.8} & \textbf{93.7} & \textbf{96.1} & 93.1 & 96.1 & 96.8 \\\midrule
ResNeXt-avg-CL & No & 2048 & 44.2 & 65.8 & 73.9 & 69.0 & 82.3 & 86.3 \\
ResNeXt-avg-GCL & No & 2048 & 57.9 & 76.1 & 82.6 & 77.7 & 86.9 & 89.7 \\
ResNeXt-avg-CL & Yes & 1024 & 69.6 & 85.4 & 89.8 & 88.9 & 94.1 & 95.6 \\
ResNeXt-avg-GCL & Yes & 1024 & 74.7 & 88.0 & 91.7 & 89.6 & 94.6 & 95.9 \\
ResNeXt-GeM-CL & No & 2048 & 50.2 & 70.8 & 78.8 & 66.1 & 78.5 & 82.7 \\
ResNeXt-GeM-GCL & No & 2048 & 58.6 & 76.2 & 81.8 & 78.7 & 87.1 & 90.1 \\
ResNeXt-GeM-CL & Yes & 1024 & 70.3 & 85.2 & 89.6 & 87.2 & 93.1 & 94.5 \\
ResNeXt-GeM-GCL & Yes & 1024 & 78.2 & 90.1 & 93.1 & 91.8 & 95.4 & 96.6 \\ \bottomrule
\end{tabular}%
}
\end{table}

\revised{\subsection{Results on Pittsburgh250k and TokyoTM}
We evaluated the generalization of our models trained on MSLS to the Pittsburgh250k~\cite{Torii-PAMI2015} and TokyoTM~\cite{Arandjelovic2017} datasets. The test set of the former consists of 83k map and 8k query images taken over the span of several years in Pittsburgh, Pennsylvania, USA. The TokyoTM dataset consists of images collected using the Time Machine tool on Google Street View in Tokyo over several years. Its validation set is divided into a map and a query set, with 49k and 7k images.

The results that we obtained are in line with those reported in the main paper. Our models trained with a GCL function consistently generalize better to unseen datasets than their counterpart trained with a binary Contrastive Loss function. Their performance is further boosted if PCA whitening is applied, up to a top-5 recall of 93.7\% on Pittsburgh250k and 96.7\% on TokyoTM. }

\begin{table}[t!]
\centering
\caption{Detailed results on the Extended CMU dataset. The symbol $^\star$ denotes the models for which PCA whitening has been applied. }
\label{tab:cmu_detailed}
\resizebox{\columnwidth}{!}{%
\setlength\tabcolsep{1.5pt}
\begin{tabular}{@{}lcccc@{}}
\toprule
\textbf{} & \textbf{Mean} & \textbf{Urban} & \textbf{Suburban} & \textbf{Park} \\ 
\textbf{Method} & \textbf{\begin{tabular}[c]{@{}c@{}}0.25/0.5/10m \\ 2/5/10$\degree$\end{tabular}} & \textbf{\begin{tabular}[c]{@{}c@{}}0.25/0.5/10m \\ 2/5/10$\degree$\end{tabular}} & \textbf{\begin{tabular}[c]{@{}c@{}}0.25/0.5/10m \\ 2/5/10$\degree$\end{tabular}} & \textbf{\begin{tabular}[c]{@{}c@{}}0.25/0.5/10m \\ 2/5/10$\degree$\end{tabular}} \\\midrule
NetVLAD-64-Pitts30k & 5.3 / 15.9 / 66.3 & 10.7 / 27.8 / 84.1 & 2.9 / 11.2 / 63.5 & 2.4 / 8.9 / 50.1 \\
NetVLAD-64-Pitts30k$^\star$ & 5.8 / 17.6 / 70.3 & 11.6 / 30.3 / 87.5 & 3.3 / 12.5 / 68 & 2.7 / 10.2 / 54.2 \\
NetVLAD-64-MSLS & 1.3 / 4.5 / 31.9 & 2.9 / 8.4 / 49.6 & 0.8 / 3.4 / 29.7 & 0.3 / 1.6 / 15.2 \\
NetVLAD-64-MSLS$^\star$ & 3.9 / 12.1 / 58.4 & 7.6 / 20.3 / 77.2 & 2.6 / 9.7 / 58.9 & 1.6 / 6.3 / 37.0 \\
NetVLAD-16-MSLS & 1.7 / 5.5 / 39.1 & 3.7 / 10.0 / 57.3 & 1.1 / 4.5 / 38.6 & 0.4 / 1.9 / 19.8 \\
NetVLAD-16-MSLS$^\star$ & 4.4 / 13.7 / 61.4 & 8.4 / 22.1 / 76.9 & 3.0 / 11.4 / 63.1 & 1.9 / 7.4 / 41.9 \\
AP-GeM$^\star$ & 4.9 / 14.7 / 65.2 & 9.8 / 25.2 / 82.6 & 2.8 / 10.9 / 65.3 & 2.1 / 8 / 45.9 \\
NetVLAD-SARE$^\star$ & \textbf{6.4} / \textbf{19.4} / \textbf{75.5} & \textbf{12.5} / \textbf{32.4 }/ \textbf{90.7} & \textbf{3.9} / \textbf{14.7 }/ \textbf{76.3} & 3.0 / 11.3 / 57.6 \\\midrule
VGG-avg-CL & 0.9 / 3.0 / 22.7 & 2.0 / 5.8 / 39.4 & 0.6 / 2.3 / 21 & 0.2 / 0.7 / 6.5 \\
VGG-avg-GCL & 2.3 / 7.2 / 43.3 & 4.7 / 13.0 / 64.3 & 1.6 / 5.8 / 43.4 & 0.7 / 2.5 / 19.8 \\
VGG-avg-CL$^\star$ & 2.1 / 6.6 / 34.9 & 4.4 / 12.3 / 54.1 & 1.1 / 4.4 / 30.5 & 0.9 / 3.1 / 19.4 \\
VGG-avg-GCL$^\star$ & 3.7 / 11.2 / 52.5 & 7.2 / 19.1 / 70.8 & 2.1 / 8.2 / 50.4 & 1.8 / 6.2 / 35.1 \\
VGG-GeM-CL & 2.8 / 8.6 / 44.5 & 5.7 / 15.2 / 63.7 & 1.6 / 6.4 / 44.9 & 1.1 / 4.2 / 22.8 \\
VGG-GeM-GCL & 3.6 / 11.2 / 55.8 & 6.8 / 18.5 / 73.6 & 2.5 / 9.2 / 57.3 & 1.5 / 5.6 / 34.2 \\
VGG-GeM-CL$^\star$ & 4.4 / 13.4 / 56.5 & 8.4 / 21.5 / 72.1 & 2.6 / 10 / 55.2 & 2.3 / 8.7 / 40.8 \\
VGG-GeM-GCL$^\star$ & 5.7 / 17.1 / 66.3 & 10.4 / 26.8 / 82.2 & 3.8 / 13.8 / 67.8 & 2.8 / 10.7 / 46.7 \\\midrule
ResNet50-avg-CL & 2.6 / 7.8 / 43.4 & 5.6 / 14.9 / 64.6 & 1.4 / 5.5 / 42.6 & 0.8 / 2.9 / 21.1 \\
ResNet50-avg-GCL & 3.1 / 9.7 / 55.1 & 6 / 16.7 / 73.6 & 2.0 / 7.7 / 56.8 & 1.2 / 4.7 / 32.5 \\
ResNet50-avg-CL$^\star$ & 4.7 / 13.8 / 50.8 & 9.5 / 24.1 / 70.5 & 2.7 / 9.7 / 48.8 & 2.1 / 7.6 / 31.7 \\
ResNet50-avg-GCL$^\star$ & 5.4 / 16.2 / 66.5 & 10.2 / 26.5 / 81.8 & 3.5 / 12.5 / 69.8 & 2.6 / 9.6 / 45.3 \\
ResNet50-GeM-CL & 3.2 / 9.6 / 49.5 & 6.5 / 17.3 / 70.7 & 1.9 / 7.0 / 49.3 & 1.2 / 4.3 / 26.3 \\
ResNet50-GeM-GCL & 3.8 / 11.8 / 61.6 & 7.4 / 19.9 / 79.2 & 2.4 / 9.4 / 64.8 & 1.5 / 6.1 / 38.1 \\
ResNet50-GeM-CL$^\star$ & 4.7 / 13.4 / 51.6 & 9.5 / 23.9 / 73.5 & 2.8 / 9.6 / 49.7 & 1.9 / 6.8 / 30.0 \\
ResNet50-GeM-GCL$^\star$ & 5.4 / 16.5 / 69.9 & 10.1 / 26.3 / 84.5 & 3.5 / 13.4 / 74.1 & 2.6 / 9.8 / 48.2 \\\midrule
ResNet152-avg-CL & 3.0 / 9.2 / 49.9 & 6.2 / 16.4 / 70.4 & 1.9 / 6.9 / 48.7 & 1.0 / 4.1 / 28.9 \\
ResNet152-avg-GCL & 3.6 / 11.0 / 61.2 & 7.0 / 18.9 / 78.8 & 2.3 / 8.4 / 62.6 & 1.4 / 5.5 / 39.9 \\
ResNet152-avg-CL$^\star$ & 4.8 / 14.3 / 59.9 & 9.4 / 24 / 77.3 & 3.0 / 10.9 / 61 & 2 / 8 / 39.3 \\
ResNet152-avg-GCL$^\star$ & 5.7 / 17.0 / 66.5 & 10.8 / 27.3 / 82.0 & 3.7 / 13.5 / 68.9 & 2.6 / 10.1 / 46.3 \\
ResNet152-GeM-CL & 3.2 / 9.7 / 52.2 & 6.7 / 17.7 / 73.9 & 2.1 / 7.3 / 52.8 & 0.9 / 4.1 / 27.6 \\
ResNet152-GeM-GCL & 3.6 / 11.3 / 63.1 & 6.9 / 18.8 / 79.3 & 2.5 / 9.0 / 64.3 & 1.3 / 6.0 / 43.6 \\
ResNet152-GeM-CL$^\star$ & 4.8 / 14.2 / 55.0 & 9.6 / 24.5 / 75.2 & 3.0 / 10.5 / 54.3 & 1.9 / 7.5 / 33.6 \\
ResNet152-GeM-GCL$^\star$ & 5.3 / 16.1 / 66.4 & 9.9 / 25.8 / 81.6 & 3.5 / 12.8 / 69.3 & 2.4 / 9.7 / 46 \\\midrule
ResNext-avg-CL & 2.0 / 6.1 / 40.0 & 4.6 / 12.2 / 63.6 & 1.2 / 4.4 / 38.5 & 0.3 / 1.7 / 16.0 \\
ResNext-avg-GCL & 3.3 / 10.2 / 57.7 & 6.7 / 17.9 / 77.8 & 2.1 / 8.1 / 58.1 & 1.2 / 4.4 / 35.0 \\
ResNext-avg-CL$^\star$ & 4.6 / 13.4 / 58.6 & 9.3 / 23.2 / 78 & 2.8 / 10.1 / 59 & 1.7 / 6.7 / 36.5 \\
ResNext-avg-GCL$^\star$ & 5.6 / 16.6 / 70.7 & 10.4 / 26.6 / 85.1 & 3.7 / 13.4 / 73.2 & 2.6 / 9.8 / 51.6 \\
ResNext-GeM-CL & 2.9 / 9.0 / 52.6 & 5.9 / 15.6 / 70.2 & 1.7 / 6.8 / 53.0 & 1.2 / 4.7 / 32.6 \\
ResNext-GeM-GCL & 3.5 / 10.5 / 58.8 & 6.5 / 17.7 / 77.8 & 2.5 / 8.7 / 59.7 & 1.3 / 5.0 / 36.6 \\
ResNext-GeM-CL$^\star$ & 4.9 / 14.4 / 61.7 & 9.4 / 24 / 80.2 & 3.1 / 11.1 / 63.8 & 2.2 / 8.0 / 38.6 \\
ResNext-GeM-GCL$^\star$ & 6.1 / 18.2 / 74.9 & 11.1 / 28.7 / 87.4 & 4.2 / 14.6 / 77.4 & \textbf{3.1} / \textbf{11.1} / \textbf{57.7 }\\ \bottomrule
\end{tabular}
}
\end{table}
\begin{table*}[]
\centering
\caption{Detailed results on the RobotCar Seasons v2 dataset, divided by wheather and ilumination conditions. The symbol $^\star$ denotes the models for which PCA whitening has been applied.}
\label{tab:robotcar_extended}
\resizebox{\textwidth}{!}{%
\setlength\tabcolsep{1.75pt}
\begin{tabular}{@{}l|c|c|c|c|c|c|c|c|c|c|c|c@{}}
\toprule
\textbf{} & \textbf{night rain} & \textbf{night} & \textbf{night all} & \textbf{overcast winter} & \textbf{sun} & \textbf{rain} & \textbf{snow} & \textbf{dawn} & \textbf{dusk} & \textbf{overcast summer} & \textbf{day all} & \textbf{mean} \\
\textbf{Method} & \textbf{\begin{tabular}[c]{@{}c@{}}0.25/0.5/10m \\ 2/5/10$\degree$\end{tabular}} & \textbf{\begin{tabular}[c]{@{}c@{}}0.25/0.5/10m \\ 2/5/10$\degree$\end{tabular}} & \textbf{\begin{tabular}[c]{@{}c@{}}0.25/0.5/10m \\ 2/5/10$\degree$\end{tabular}} & \textbf{\begin{tabular}[c]{@{}c@{}}0.25/0.5/10m \\ 2/5/10$\degree$\end{tabular}} & \textbf{\begin{tabular}[c]{@{}c@{}}0.25/0.5/10m \\ 2/5/10$\degree$\end{tabular}} & \textbf{\begin{tabular}[c]{@{}c@{}}0.25/0.5/10m \\ 2/5/10$\degree$\end{tabular}} & \textbf{\begin{tabular}[c]{@{}c@{}}0.25/0.5/10m \\ 2/5/10$\degree$\end{tabular}} & \textbf{\begin{tabular}[c]{@{}c@{}}0.25/0.5/10m \\ 2/5/10$\degree$\end{tabular}} & \textbf{\begin{tabular}[c]{@{}c@{}}0.25/0.5/10m \\ 2/5/10$\degree$\end{tabular}} & \textbf{\begin{tabular}[c]{@{}c@{}}0.25/0.5/10m \\ 2/5/10$\degree$\end{tabular}} & \textbf{\begin{tabular}[c]{@{}c@{}}0.25/0.5/10m \\ 2/5/10$\degree$\end{tabular}} & \textbf{\begin{tabular}[c]{@{}c@{}}0.25/0.5/10m \\ 2/5/10$\degree$\end{tabular}} \\ \midrule
NetVLAD-64-Pitts30k & 1.5 / 1.5 / 10.3 & 0.4 / 1.8 / 8.0 & 0.9 / 1.6 / 9.1 & 1.2 / 23.8 / 97.0 & 5.4 / 14.3 / 74.1 & 10.2 / 44.4 / 99.5 & 9.3 / 31.6 / 94.9 & 9.7 / 26.4 / 82.8 & 5.1 / 26.4 / 92.4 & 6.6 / 30.3 / 88.6 & 7.0 / 28.1 / 89.4 & 5.6 / 22 / 71 \\
NetVLAD-64-Pitts30k$^\star$ & 2.0 / 3.9 / 13.3 & 0.0 / 1.3 / 11.5 & 0.9 / 2.6 / 12.4 & 1.2 / 22.6 / \textbf{100.0} & 4.9 / 18.8 / 80.8 & 9.3 / 42.4 / 98.5 & 10.2 / 32.6 / 94.9 & \textbf{11.5 / 29.5 / 85.5} & 5.6 / 27.9 / 92.4 & 7.1 / 30.3 / 90.0 & 7.3 / 29.2 / 91.3 & 5.8 / 23.1 / 73.2 \\
NetVLAD-64-MSLS & 0.5 / 1.0 / 5.9 & 0.0 / 0.4 / 4.4 & 0.2 / 0.7 / 5.1 & 0.0 / 11.6 / 67.7 & 0.9 / 4.5 / 28.6 & 4.4 / 22.0 / 91.2 & 4.7 / 13.0 / 60.5 & 3.1 / 8.4 / 32.6 & 2.5 / 14.2 / 78.7 & 1.4 / 9.5 / 51.7 & 2.5 / 11.7 / 57.5 & 2 / 9.2 / 45.5 \\
NetVLAD-64-MSLS$^\star$ & 0.0 / 1.5 / 11.8 & 0.0 / 1.3 / 12.4 & 0.0 / 1.4 / 12.1 & 0.0 / 15.2 / 87.2 & 3.1 / 12.1 / 66.5 & 8.8 / 35.6 / 99.0 & 7.0 / 28.4 / 90.2 & 9.3 / 21.6 / 77.5 & 4.1 / 25.4 / 98.0 & 5.2 / 21.3 / 77.7 & 5.5 / 22.9 / 84.7 & 4.2 / 18 / 68.1 \\
NetVLAD-16-MSLS & 0.0 / 0.0 / 3.4 & 0.0 / 0.9 / 5.3 & 0.0 / 0.5 / 4.4 & 1.2 / 10.4 / 78.0 & 0.9 / 6.2 / 33.0 & 5.9 / 25.4 / 93.7 & 3.3 / 10.7 / 62.8 & 1.8 / 6.6 / 29.1 & 1.5 / 16.2 / 86.8 & 1.4 / 8.1 / 57.8 & 2.3 / 11.8 / 61.5 & 1.8 / 9.2 / 48.4 \\
NetVLAD-16-MSLS$^\star$ & 0.0 / 0.0 / 1.0 & 0.0 / 0.9 / 7.1 & 0.0 / 0.5 / 4.2 & 1.8 / 19.5 / 90.2 & 4.9 / 11.6 / 67.4 & 10.7 / 34.1 / 96.1 & 6.5 / 24.7 / 89.8 & 7.9 / 24.2 / 74.4 & 2.5 / 23.4 / 93.9 & 7.6 / 24.6 / 77.3 & 6.2 / 23.1 / 83.6 & 4.8 / 17.9 / 65.3 \\
AP-GeM & 0.0 / 2.0 / 14.3 & 0.0 / 0.9 / 10.6 & 0.0 / 1.4 / 12.4 & 1.2 / 26.2 / 89.6 & 4.0 / 17.4 / 62.1 & 13.7 / 41.0 / 98.0 & 7.0 / 26.0 / 87.9 & 8.4 / 24.7 / 71.8 & 5.1 / 25.4 / 94.4 & 5.7 / 23.7 / 75.8 & 6.6 / 26.2 / 82.1 & 5.1 / 20.5 / 66.1 \\
NetVLAD-SARE$^\star$ & 3.4 / 8.9 / 38.9 & 0.9 / 4.4 / 31.4 & 2.1 / 6.5 / 35.0 & 2.4 / \textbf{29.3} / 97.6 & \textbf{8.9 / 22.3 / 92.9} & 13.2 / 43.9 / \textbf{100.0} & \textbf{12.1 / 36.3 / 98.1} & 10.6 / 28.2 / 89.9 & \textbf{5.6 / 30.5 / 95.9} & \textbf{8.5 / 36.5 / 91.9} & \textbf{9.0 / 32.4 / 95.0} &\textbf{ 7.4 / 26.5 / 81.3} \\ \midrule
VGG-avg-CL & 0.0 / 0.0 / 2.5 & 0.0 / 0.0 / 1.3 & 0.0 / 0.0 / 1.9 & 0.0 / 7.9 / 54.9 & 0.0 / 2.2 / 14.3 & 4.9 / 23.9 / 85.4 & 3.3 / 8.8 / 49.8 & 0.9 / 1.3 / 14.5 & 0.0 / 12.2 / 66.5 & 0.0 / 3.3 / 31.8 & 1.3 / 8.3 / 44.0 & 1 / 6.4 / 34.4 \\
VGG-avg-GCL & 0.0 / 0.5 / 3.9 & 0.0 / 0.0 / 1.8 & 0.0 / 0.2 / 2.8 & 0.6 / 15.2 / 82.9 & 2.2 / 6.7 / 44.6 & 7.8 / 31.7 / 95.1 & 4.2 / 19.5 / 73.0 & 0.4 / 5.7 / 33.9 & 2.0 / 21.8 / 86.3 & 3.8 / 13.7 / 57.8 & 3.0 / 16.1 / 66.3 & 2.3 / 12.5 / 51.7 \\
VGG-avg-CL$^\star$ & 0.0 / 0.0 / 0.0 & 0.0 / 0.0 / 0.0 & 0.0 / 0.0 / 0.0 & 0.0 / 11.0 / 61.6 & 0.4 / 4.0 / 20.5 & 5.9 / 26.3 / 85.4 & 5.1 / 13.5 / 64.2 & 3.1 / 8.4 / 40.1 & 4.1 / 17.3 / 76.6 & 1.9 / 11.8 / 40.3 & 3.0 / 13.0 / 54.5 & 2.3 / 10 / 42 \\
VGG-avg-GCL$^\star$ & 0.0 / 0.0 / 1.0 & 0.0 / 0.0 / 0.4 & 0.0 / 0.0 / 0.7 & 0.6 / 17.1 / 79.3 & 3.1 / 11.2 / 49.1 & 9.3 / 34.6 / 97.6 & 6.5 / 17.7 / 80.0 & 3.5 / 16.3 / 56.4 & 5.6 / 25.4 / 90.9 & 3.8 / 18.0 / 69.7 & 4.7 / 19.9 / 73.9 & 3.6 / 15.3 / 57.1 \\
VGG-GeM-CL & 0.0 / 0.0 / 2.0 & 0.0 / 0.4 / 1.8 & 0.0 / 0.2 / 1.9 & 0.0 / 14.6 / 80.5 & 2.2 / 6.7 / 45.5 & 8.3 / 31.2 / 94.1 & 6.0 / 20.0 / 77.7 & 3.5 / 11.5 / 47.1 & 2.0 / 17.8 / 89.8 & 4.7 / 18.0 / 68.2 & 4.0 / 17.0 / 70.8 & 3.1 / 13.2 / 55 \\
VGG-GeM-GCL & 0.0 / 0.5 / 7.4 & 0.0 / 0.0 / 0.9 & 0.0 / 0.2 / 4.0 & 0.6 / 16.5 / 82.3 & 2.2 / 11.6 / 61.2 & 10.2 / 38.0 / 99.0 & 7.9 / 23.3 / 83.7 & 2.6 / 11.5 / 51.1 & 2.0 / 22.3 / 90.9 & 7.1 / 20.9 / 70.6 & 4.8 / 20.4 / 76.2 & 3.7 / 15.8 / 59.7 \\
VGG-GeM-CL$^\star$ & 0.0 / 0.5 / 2.0 & 0.0 / 0.0 / 0.0 & 0.0 / 0.2 / 0.9 & 0.6 / 18.9 / 84.8 & 2.7 / 13.4 / 60.3 & 8.3 / 36.6 / 98.5 & 8.4 / 27.0 / 85.1 & 6.6 / 22.5 / 70.5 & 5.6 / 29.9 / 95.9 & 4.7 / 21.3 / 74.9 & 5.4 / 24.2 / 80.8 & 4.2 / 18.7 / 62.5 \\
VGG-GeM-GCL$^\star$ & 0.5 / 1.0 / 6.4 & 0.0 / 1.3 / 6.6 & 0.2 / 1.2 / 6.5 & 1.2 / 23.2 / 90.9 & 6.7 / 21.4 / 78.1 & 11.2 / 42.4 / \textbf{100.0} & 9.3 / 32.6 / 95.3 & 7.5 / 20.7 / 76.7 & 4.6 / 28.9 / 97.5 & 6.2 / 27.0 / 79.6 & 6.9 / 28.0 / 87.9 & 5.4 / 21.9 / 69.2 \\ \midrule
ResNet50-avg-CL & 0.5 / 1.0 / 9.4 & 0.4 / 2.2 / 11.1 & 0.5 / 1.6 / 10.3 & 0.0 / 13.4 / 71.3 & 1.8 / 6.2 / 30.4 & 8.8 / 32.7 / 97.6 & 5.6 / 18.6 / 64.7 & 4.4 / 15.4 / 53.7 & 3.0 / 14.7 / 79.2 & 2.8 / 10.4 / 51.7 & 3.9 / 15.9 / 63.1 & 3.1 / 12.6 / 51 \\
ResNet50-avg-GCL & 1.0 / 2.0 / 12.8 & 0.4 / 1.3 / 8.4 & 0.7 / 1.6 / 10.5 & 0.6 / 18.3 / 80.5 & 1.3 / 6.7 / 43.8 & 8.3 / 34.1 / 96.1 & 4.7 / 15.3 / 72.6 & 2.6 / 11.0 / 51.1 & 2.0 / 16.2 / 88.8 & 4.3 / 14.2 / 64.0 & 3.5 / 16.3 / 69.9 & 2.9 / 12.9 / 56.3 \\
ResNet50-avg-CL$^\star$ & 0.0 / 0.0 / 3.0 & 0.0 / 0.4 / 1.8 & 0.0 / 0.2 / 2.3 & 2.4 / 26.2 / 89.6 & 3.6 / 14.3 / 53.1 & \textbf{14.6 / 45.4 }/ 98.0 & 8.4 / 30.7 / 84.2 & 10.1 / 31.7 / 75.3 & 5.6 / 24.9 / 93.4 & 7.6 / 33.6 / 76.8 & 7.6 / 29.5 / 80.7 & 5.9 / 22.8 / 62.7 \\
ResNet50-avg-GCL$^\star$ & 0.0 / 1.0 / 6.4 & 0.4 / 0.4 / 10.2 & 0.2 / 0.7 / 8.4 & \textbf{3.0} / 28.0 / 98.2 & 4.0 / 15.6 / 68.8 & 13.2 / 41.5 / 97.6 & 8.4 / 31.2 / 87.9 & 8.4 / 30.4 / 84.1 & 5.6 / 27.9 / 96.4 & 9.5 / 33.6 / 86.3 & 7.6 / 29.7 / 87.8 & 5.9 / 23.1 / 69.6 \\
ResNet50-GeM-CL & 0.5 / 1.5 / 11.8 & 0.4 / 1.3 / 8.0 & 0.5 / 1.4 / 9.8 & 0.0 / 16.5 / 82.9 & 0.9 / 9.8 / 61.2 & 9.8 / 33.7 / 97.1 & 5.6 / 23.7 / 81.4 & 3.1 / 15.0 / 62.6 & 1.5 / 19.8 / 93.4 & 6.2 / 18.5 / 64.9 & 4.0 / 19.5 / 76.9 & 3.2 / 15.4 / 61.5 \\
ResNet50-GeM-GCL & 0.5 / 2.0 / 9.9 & 0.0 / 1.3 / 11.9 & 0.2 / 1.6 / 11.0 & 1.8 / 21.3 / 84.8 & 1.8 / 10.3 / 47.3 & 8.8 / 33.7 / 97.1 & 5.1 / 18.1 / 80.9 & 1.8 / 8.8 / 49.8 & 2.0 / 17.8 / 91.9 & 4.7 / 16.6 / 69.7 & 3.7 / 17.7 / 73.4 & 2.9 / 14 / 58.8 \\
ResNet50-GeM-CL$^\star$ & 0.5 / 0.5 / 3.9 & 0.0 / 0.4 / 2.7 & 0.2 / 0.5 / 3.3 & \textbf{3.0} / 28.0 / 95.1 & 2.2 / 13.8 / 56.7 & 9.8 / 37.6 / 98.5 & 8.4 / 32.6 / 94.0 & 7.0 / 23.8 / 83.7 & 5.6 / 24.9 / 94.4 & 8.1 / 30.8 / 79.6 & 6.4 / 27.2 / 85.3 & 5 / 21.1 / 66.5 \\
ResNet50-GeM-GCL$^\star$ & 0.0 / 0.0 / 8.4 & 0.0 / 0.0 / 9.3 & 0.0 / 0.0 / 8.9 & \textbf{3.0} / 25.6 / 95.7 & 3.1 / 14.7 / 69.2 & 9.3 / 34.6 / 98.0 & 6.5 / 28.4 / 90.7 & 9.3 / 28.6 / 86.3 & 3.6 / 25.9 / 96.4 & 7.1 / 26.1 / 83.9 & 6.1 / 26.2 / 88.1 & 4.7 / 20.2 / 70 \\ \midrule
ResNet152-avg-CL & 0.0 / 0.0 / 8.4 & 0.0 / 0.0 / 8.0 & 0.0 / 0.0 / 8.2 & 1.2 / 17.1 / 86.0 & 1.8 / 11.2 / 51.3 & 6.8 / 34.1 / 92.7 & 4.7 / 15.8 / 74.0 & 2.6 / 11.9 / 53.3 & 3.0 / 17.3 / 82.7 & 2.4 / 13.3 / 64.0 & 3.3 / 17.0 / 71.0 & 2.5 / 13.1 / 56.6 \\
ResNet152-avg-GCL & 0.0 / 1.5 / 11.3 & 0.4 / 0.4 / 10.2 & 0.2 / 0.9 / 10.7 & 1.8 / 20.1 / 90.2 & 3.6 / 8.5 / 52.7 & 9.3 / 29.3 / 95.6 & 4.7 / 19.1 / 84.2 & 2.6 / 12.8 / 47.6 & 3.0 / 16.8 / 88.8 & 4.3 / 16.1 / 70.6 & 4.2 / 17.3 / 74.5 & 3.3 / 13.5 / 59.9 \\
ResNet152-avg-CL$^\star$ & 3.0 / 10.9 / 61.0 & 4.3 / 13.6 / 57.6 & 9.4 / 24.0 / 77.3 & 2.0 / 8.0 / 39.3 & 5.1 / 14.0 / 60.8 & 4.9 / 14.2 / 58.1 & 4.8 / 14.1 / 59.8 & 4.5 / 13.3 / 55.1 & 5.3 / 17.0 / 69.7 & 3.6 / 12.3 / 53.7 & 5.5 / 16.2 / 67.2 & 5.4 / 22.2 / 64.8 \\
ResNet152-avg-GCL$^\star$ & \textbf{3.7 / 13.5 / 68.9} & \textbf{5.3 / 16.6 / 65.0} & \textbf{10.8 / 27.3 / 82.0} & 2.6 / 10.1 / 46.3 & 6.1 / 16.7 / 68.2 & 5.7 / 16.8 / 63.1 & 5.6 / 16.4 / 65.0 & 5.7 / 16.4 / 63.9 & 6.1 / 19.6 / 74.9 & 4.3 / 14.6 / 61.6 & 6.2 / 18.5 / 73.6 & 6.2 / 23.9 / 70 \\
ResNet152-GeM-CL & 2.1 / 7.3 / 52.8 & 3.0 / 9.8 / 53.9 & 6.7 / 17.7 / 73.9 & 0.9 / 4.1 / 27.6 & 3.2 / 8.9 / 52.2 & 3.3 / 9.2 / 47.4 & 3.1 / 8.9 / 47.9 & 3.3 / 9.9 / 53.0 & 3.4 / 11.5 / 61.0 & 1.9 / 7.9 / 45.6 & 3.7 / 11.0 / 56.3 & 3.3 / 15.2 / 64 \\
ResNet152-GeM-GCL & 2.5 / 9.0 / 64.3 & 3.3 / 11.0 / 63.8 & 6.9 / 18.8 / 79.3 & 1.3 / 6.0 / 43.6 & 4.1 / 11.4 / 62.9 & 3.5 / 11.2 / 58.9 & 3.5 / 10.8 / 60.1 & 3.7 / 11.2 / 62.9 & 3.6 / 12.6 / 70.4 & 2.5 / 9.4 / 58.9 & 3.8 / 11.9 / 68.4 & 2.9 / 13.1 / 63.5 \\
ResNet152-GeM-CL$^\star$ & 3.0 / 10.5 / 54.3 & 4.4 / 13.6 / 53.2 & 9.6 / 24.5 / 75.2 & 1.9 / 7.5 / 33.6 & 4.9 / 13.4 / 53.9 & 5.0 / 14.1 / 53.5 & 4.9 / 13.7 / 54.6 & 4.4 / 12.9 / 50.0 & 5.6 / 17.5 / 65.6 & 3.9 / 12.5 / 47.1 & 5.5 / 15.9 / 62.2 & 6.1 / 23.5 / 68.9 \\
ResNet152-GeM-GCL$^\star$ & 3.5 / 12.8 / 69.3 & 4.8 / 16.0 / 65.4 & 9.9 / 25.8 / 81.6 & 2.4 / 9.7 / 46.0 & 5.6 / 15.6 / 67.0 & 5.3 / 15.5 / 62.8 & 5.2 / 15.2 / 64.7 & 5.1 / 15.6 / 63.7 & 5.7 / 19.5 / 75.8 & 4.0 / 14.8 / 62.3 & 6.1 / 17.9 / 74.0 & 6 / 21.6 / 72.5 \\ \midrule
ResNext-avg-CL & 1.0 / 3.4 / 29.1 & 0.0 / 1.3 / 15.5 & 0.5 / 2.3 / 21.9 & 1.2 / 12.8 / 86.0 & 1.3 / 4.9 / 55.8 & 6.8 / 20.5 / 97.6 & 6.5 / 16.7 / 80.0 & 2.2 / 9.7 / 44.5 & 2.5 / 17.8 / 84.8 & 2.8 / 13.3 / 66.8 & 3.4 / 13.5 / 72.6 & 2.7 / 10.9 / 61 \\
ResNext-avg-GCL & 0.0 / 0.5 / 11.8 & 0.0 / 0.9 / 15.9 & 0.0 / 0.7 / 14.0 & 1.2 / 19.5 / 91.5 & 2.7 / 8.9 / 60.3 & 6.8 / 25.4 / 96.6 & 4.2 / 15.3 / 78.1 & 2.6 / 11.9 / 63.4 & 2.5 / 16.2 / 93.9 & 4.3 / 15.6 / 66.8 & 3.5 / 15.9 / 77.7 & 2.7 / 12.4 / 63.1 \\
ResNext-avg-CL$^\star$ & 1.0 / 1.5 / 12.3 & 0.0 / 0.9 / 12.8 & 0.5 / 1.2 / 12.6 & 0.6 / 25.6 / 97.6 & 4.9 / 17.0 / 79.9 & 9.3 / 35.1 / 99.0 & 8.4 / 29.8 / 93.0 & 7.0 / 23.8 / 72.7 & 5.6 / 24.9 / 94.4 & 5.7 / 27.0 / 82.9 & 6.1 / 26.1 / 87.9 & 4.8 / 20.4 / 70.6 \\
ResNext-avg-GCL$^\star$ & 1.0 / 3.4 / 29.1 & 0.0 / 1.3 / 15.5 & 0.5 / 2.3 / 21.9 & 1.2 / 12.8 / 86.0 & 1.3 / 4.9 / 55.8 & 6.8 / 20.5 / 97.6 & 6.5 / 16.7 / 80.0 & 2.2 / 9.7 / 44.5 & 2.5 / 17.8 / 84.8 & 2.8 / 13.3 / 66.8 & 3.4 / 13.5 / 72.6 & 2.7 / 10.9 / 61 \\
ResNext-GeM-CL & 0.0 / 1.5 / 6.9 & 0.0 / 0.4 / 8.0 & 0.0 / 0.9 / 7.5 & 0.0 / 17.1 / 82.3 & 1.3 / 5.4 / 56.7 & 3.9 / 20.5 / 96.6 & 3.7 / 12.6 / 74.9 & 2.2 / 7.0 / 29.5 & 2.0 / 21.3 / 90.4 & 2.8 / 11.4 / 60.7 & 2.4 / 13.2 / 68.9 & 1.9 / 10.4 / 54.8 \\
ResNext-GeM-GCL & 0.0 / 2.5 / 22.2 & 0.4 / 1.3 / 19.9 & 0.2 / 1.9 / 21.0 & 1.2 / 15.2 / 84.8 & 0.9 / 7.6 / 68.8 & 7.3 / 28.3 / 98.5 & 5.1 / 20.0 / 84.2 & 2.2 / 11.0 / 55.5 & 4.1 / 20.8 / 90.9 & 3.3 / 16.1 / 71.6 & 3.5 / 16.8 / 78.4 & 2.7 / 13.4 / 65.2 \\
ResNext-GeM-CL$^\star$ & 0.5 / 2.5 / 10.3 & 0.0 / 0.0 / 11.9 & 0.2 / 1.2 / 11.2 & 1.8 / 20.1 / 92.1 & 4.9 / 15.6 / 79.0 & 7.8 / 31.2 / \textbf{100.0} & 7.0 / 26.0 / 93.0 & 3.5 / 15.0 / 67.0 & 4.6 / 26.9 / 95.9 & 3.8 / 19.4 / 73.0 & 4.9 / 21.9 / 85.1 & 3.8 / 17.2 / 68.2 \\
ResNext-GeM-GCL$^\star$ & 2.5 / 5.4 / 38.4 & 1.8 / 3.5 / 28.3 & 2.1 / 4.4 / 33.1 & 1.2 / 26.8 / 93.9 & 3.6 / 15.2 / 77.2 & 9.8 / 40.0 / 98.5 & 7.9 / 29.3 / 91.2 & 4.8 / 19.8 / 82.4 & 5.1 / 25.9 / 92.9 & 5.7 / 26.1 / 76.8 & 5.5 / 25.9 / 87.1 & 4.7 / 21 / 74.7 \\ \bottomrule
\end{tabular}%

}
\end{table*}

\begin{table*}[t!]
\setlength\tabcolsep{1.5pt}

\centering
\caption{Comparison with Thoma et al.~\cite{thoma2020soft}, that outperformed several triplet-based loss functions on the CMU Seasons dataset.}
\label{tab:comparisonwiththoma}
\begin{tabular*}{\textwidth}{@{}lcccccccccccc@{}}
\toprule
 & \multicolumn{3}{c}{\textbf{All}} & \multicolumn{3}{c}{\textbf{Urban}} & \multicolumn{3}{c}{\textbf{Suburban}} & \multicolumn{3}{c}{\textbf{Park}} \\ 
\textbf{Method} & \textbf{0.25m/2$\degree$} & \textbf{0.5m/5$\degree$} & \textbf{5.0m/10$\degree$} & \textbf{0.25m/2$\degree$} & \textbf{0.5m/5$\degree$} & \textbf{5.0m/10$\degree$} & \textbf{0.25m/2$\degree$} & \textbf{0.5m/5$\degree$} & \textbf{5.0m/10$\degree$} & \textbf{0.25m/2$\degree$} & \textbf{0.5m/5$\degree$} & \textbf{5.0m/10$\degree$} \\\midrule
\textbf{Thoma et al~\cite{thoma2020soft}} & 8.0 & 20.5 & 70.4 & 12.7 & 30.7 & 84.6 & 5.1 & 14.9 & 67.9 & 4.5 & 12.6 & 56.8 \\
\textbf{VGG-GeM-GCL} & 5.9 & 15.4 & 58.9 & 9.6 & 24.1 & 76.4 & 3.1 & 9.8 & 56.3 & 3.3 & 9.0 & 42.0 \\
\textbf{VGG-GeM-GCL+PCA$_w$} & 9.2 & 22.8 & 65.8 & 14.7 & 34.2 & 82.6 & 5.7 & 16.1 & 64.6 & 5.1 & 14.0 & 49.1 \\
\textbf{Resnet50-GeM-GCL} & 6.3 & 16.3 & 62.8 & 10.4 & 25.5 & 81.2 & 3.8 & 11.2 & 63.9 & 3.2 & 9.0 & 43.3 \\
\textbf{Resnet50-GeM-GCL+PCA$_w$} & 9.0 & 22.5 & 69.3 & 14.3 & 33.8 & 85.1 & 6.1 & 17.0 & 71.1 & 4.9 & 13.4 & 52.2 \\
\textbf{Resnet152-GeM-GCL} & 6.2 & 16.2 & 66.2 & 10.1 & 24.8 & 82.0 & 4.1 & 12.2 & 66.4 & 3.1 & 9.2 & 49.9 \\
\textbf{Resnet152-GeM-GCL+PCA$_w$} & 8.8 & 21.6 & 64.0 & 14.1 & 32.4 & 80.0 & 5.9 & 16.8 & 65.2 & 4.7 & 12.7 & 47.0 \\
\textbf{ResNeXt-GeM-GCL} & 5.8 & 15.1 & 63.0 & 9.5 & 24.0 & 81.0 & 3.5 & 10.5 & 62.5 & 2.9 & 7.9 & 44.7 \\
\textbf{ResNeXt-GeM-GCL+PCA$_w$} & \textbf{9.9} & \textbf{24.3} & \textbf{75.5} & \textbf{15.4} & \textbf{36.0} & \textbf{89.6} & \textbf{6.7} & \textbf{18.4} & \textbf{76.8} & \textbf{5.6} & \textbf{14.9} & \textbf{60.3} 
\\ \bottomrule
\end{tabular*}%
\end{table*}

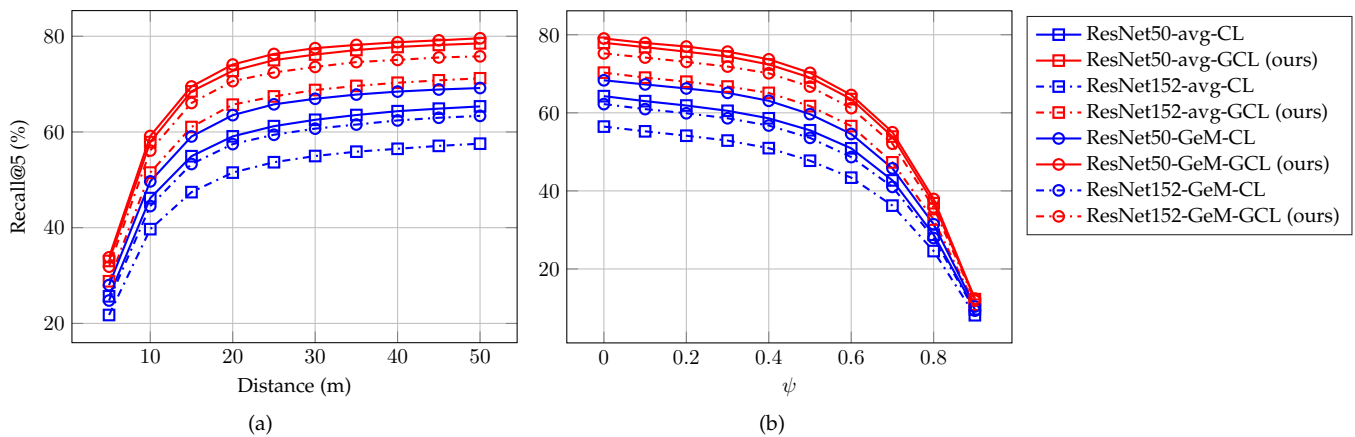
\begin{figure*}[t!]
\centering


\subfloat[]{
\begin{tikzpicture}[thick,scale=1, every node/.style={scale=.8}]
\begin{axis}
[xlabel=Distance (m),
ylabel=Recall@5 (\%), ylabel style={at={(axis description cs:-0.08,0.5)}},
legend pos =south east,grid, height=6cm, width=7.5cm]

\addplot[blue, mark size=2pt, thick , mark=square,mark options=solid, style=dashdotted] table [x=p, y=r, col sep=comma] {plot_data/appendix/MSLS/MSLS_ResNet50_avg_480_CL_recall_vs_dist.txt};

\addplot[red, mark size=2pt, thick , mark=square,mark options=solid, style=dashdotted] table [x=p, y=r, col sep=comma] {plot_data/appendix/MSLS/MSLS_ResNet50_avg_480_GCL_recall_vs_dist.txt};
\addplot[blue, mark size=2pt, thick , mark=o,mark options=solid, style=dashdotted] table [x=p, y=r, col sep=comma] {plot_data/appendix/MSLS/MSLS_ResNet50_GeM_480_CL_recall_vs_dist.txt};
\addplot[red, mark size=2pt, thick , mark=o,mark options=solid, style=dashdotted] table [x=p, y=r, col sep=comma] {plot_data/appendix/MSLS/MSLS_ResNet50_GeM_480_GCL_recall_vs_dist.txt};
\addplot[blue, mark size=2pt, thick , mark=square,mark options=solid, style=solid] table [x=p, y=r, col sep=comma] {plot_data/appendix/MSLS/MSLS_ResNet152_avg_480_CL_recall_vs_dist.txt};
\addplot[red, mark size=2pt, thick , mark=square,mark options=solid, style=solid] table [x=p, y=r, col sep=comma] {plot_data/appendix/MSLS/MSLS_ResNet152_avg_480_GCL_recall_vs_dist.txt};
\addplot[blue, mark size=2pt, thick , mark=o,mark options=solid, style=solid] table [x=p, y=r, col sep=comma] {plot_data/appendix/MSLS/MSLS_ResNet152_GeM_480_CL_recall_vs_dist.txt};
\addplot[red, mark size=2pt, thick , mark=o,mark options=solid, style=solid] table [x=p, y=r, col sep=comma] {plot_data/appendix/MSLS/MSLS_ResNet152_GeM_480_GCL_recall_vs_dist.txt};

\end{axis}

\end{tikzpicture}\label{fig:msls_dist}
}
\subfloat[]{
\begin{tikzpicture}[thick,scale=1, every node/.style={scale=.8}]
\begin{axis}
[xlabel=$\psi$,
legend pos =south east,grid, height=6cm, width=7.5cm]
\addplot[blue, mark size=2pt, thick , mark=square,mark options=solid, style=dashdotted] table [x=p, y=r, col sep=comma] {plot_data/appendix/MSLS/MSLS_ResNet50_avg_480_CL_recall_vs_psi.txt};
\addplot[red, mark size=2pt, thick , mark=square,mark options=solid, style=dashdotted] table [x=p, y=r, col sep=comma] {plot_data/appendix/MSLS/MSLS_ResNet50_avg_480_GCL_recall_vs_psi.txt};
\addplot[blue, mark size=2pt, thick , mark=o,mark options=solid, style=dashdotted] table [x=p, y=r, col sep=comma] {plot_data/appendix/MSLS/MSLS_ResNet50_GeM_480_CL_recall_vs_psi.txt};
\addplot[red, mark size=2pt, thick , mark=o,mark options=solid, style=dashdotted] table [x=p, y=r, col sep=comma] {plot_data/appendix/MSLS/MSLS_ResNet50_GeM_480_GCL_recall_vs_psi.txt};
\addplot[blue, mark size=2pt, thick , mark=square,mark options=solid, style=solid] table [x=p, y=r, col sep=comma] {plot_data/appendix/MSLS/MSLS_ResNet152_avg_480_CL_recall_vs_psi.txt};
\addplot[red, mark size=2pt, thick , mark=square,mark options=solid, style=solid] table [x=p, y=r, col sep=comma] {plot_data/appendix/MSLS/MSLS_ResNet152_avg_480_GCL_recall_vs_psi.txt};
\addplot[blue, mark size=2pt, thick , mark=o,mark options=solid, style=solid] table [x=p, y=r, col sep=comma] {plot_data/appendix/MSLS/MSLS_ResNet152_GeM_480_CL_recall_vs_psi.txt};
\addplot[red, mark size=2pt, thick , mark=o,mark options=solid, style=solid] table [x=p, y=r, col sep=comma] {plot_data/appendix/MSLS/MSLS_ResNet152_GeM_480_GCL_recall_vs_psi.txt};

\end{axis}
\end{tikzpicture}
\label{fig:msls_psi}
}
\subfloat{
\begin{minipage}[t]{.65\columnwidth}
\vspace{-5.15cm}
\begin{tikzpicture}[thick, scale=1, every node/.style={scale=.8}] 
    \begin{axis}[%
    hide axis,
    xmin=20,
    xmax=50,
    ymin=0,
    ymax=0.4,
    height=6cm, width=2cm,
    legend style={draw=white!15!black,legend cell align=left}
    ]
\addlegendimage{blue, mark size=2pt, thick , mark=square, style=solid,mark options=solid}
\addlegendentry{ResNet50-avg-CL};
\addlegendimage{red, mark size=2pt, thick , mark=square, style=solid,mark options=solid}
\addlegendentry{ResNet50-avg-GCL (ours)};
\addlegendimage{blue, mark size=2pt, thick , mark=square, style=dashdotted,mark options=solid}
\addlegendentry{ResNet152-avg-CL};
\addlegendimage{red, mark size=2pt, thick , mark=square, style=dashdotted,mark options=solid}
\addlegendentry{ResNet152-avg-GCL (ours)};
\addlegendimage{blue, mark size=2pt, thick , mark=o, style=solid,mark options=solid}
\addlegendentry{ResNet50-GeM-CL};
\addlegendimage{red, mark size=2pt, thick , mark=o, style=solid,mark options=solid}
\addlegendentry{ResNet50-GeM-GCL (ours)};
\addlegendimage{blue, mark size=2pt, thick , mark=o, style=dashdotted,mark options=solid}
\addlegendentry{ResNet152-GeM-CL};
\addlegendimage{red, mark size=2pt, thick , mark=o, style=dashdotted,mark options=solid}
\addlegendentry{ResNet152-GeM-GCL (ours)};
    \end{axis}
    \end{tikzpicture} 

\end{minipage}
}

   \caption{Comparison of the results achieved by methods trained with GCL and CL for (a) different distance thresholds and (b) different $\psi$ threshold values. GCL-based methods are plotted in red, while CL-based methods are shown in blue.}
       \label{fig:msls_psi_dist}

\end{figure*}

\revised{\subsection{Detailed results on RobotCar Seasons v2 and Extended CMU Seasons}

We provide results divided by type of environment for the Extended CMU Dataset and display them in Table~\ref{tab:cmu_detailed}. We observed that all models (ours and SoTA ones) tend to reach a higher performance on the urban images. This is logical, as they are all trained on datasets with images depicting mainly urban areas. 

We also provide detailed results for the RobotCar Seasons v2 dataset, organized by weather and illumination conditions, in Table~\ref{tab:robotcar_extended}. We observe that all methods obtain a higher successful localization rate on day conditions, but our methods tend to outperform the state-of-the-art on the night-time subsets. The MSLS train dataset includes images taken at night, but those are heavily underrepresented, so it is very interesting that GCL based models can localize images under these conditions.

}

\revised{\subsection{Comparison to Thoma et al~\cite{thoma2020soft}}
In Table~\ref{tab:comparisonwiththoma}, we show the results that we obtained on the first version of the CMU dataset, and we compare our performance to the one achieved by the method presented in~\cite{thoma2020soft}. Our method is trained on the MSLS dataset, while the model of~\cite{thoma2020soft} is a NetVLAD model trained on the Oxford RobotCar dataset~\cite{Maddern2017}. We observe that our method generalizes as well as~\cite{thoma2020soft} to this unseen dataset with the same VGG backbone, and significantly better when using a bigger backbone like ResNeXt. It is worth noting that our method uses a simple pooling method (namely GeM~\cite{radenovic2018fine}), while~\cite{thoma2020soft} deploys a VLAD layer.}

\begin{figure*}[t]
\subfloat[heads]{
\centering
\includegraphics[width=.25\textwidth]{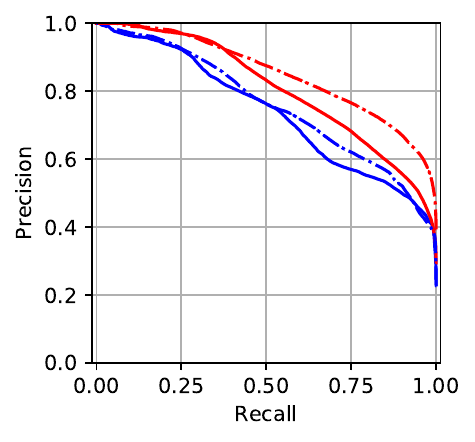}
}\hspace{-0.5cm}
\subfloat[stairs]{
\centering
\includegraphics[width=.25\textwidth]{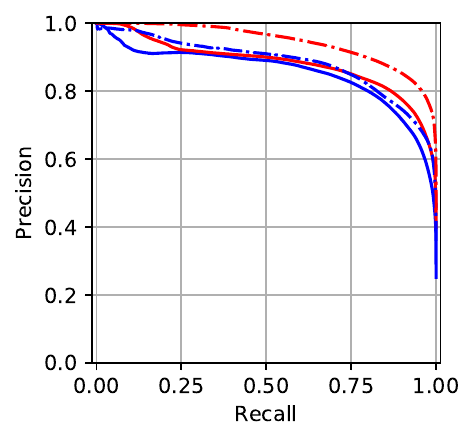}
}\hspace{-0.5cm}
\subfloat[pumpkin]{
\centering
\includegraphics[width=.25\textwidth]{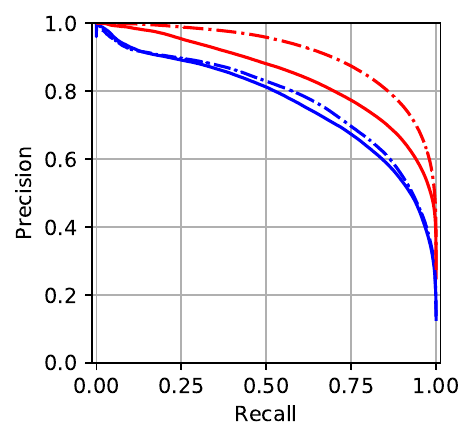}
}
\subfloat{
\centering
\includegraphics[width=.25\textwidth]{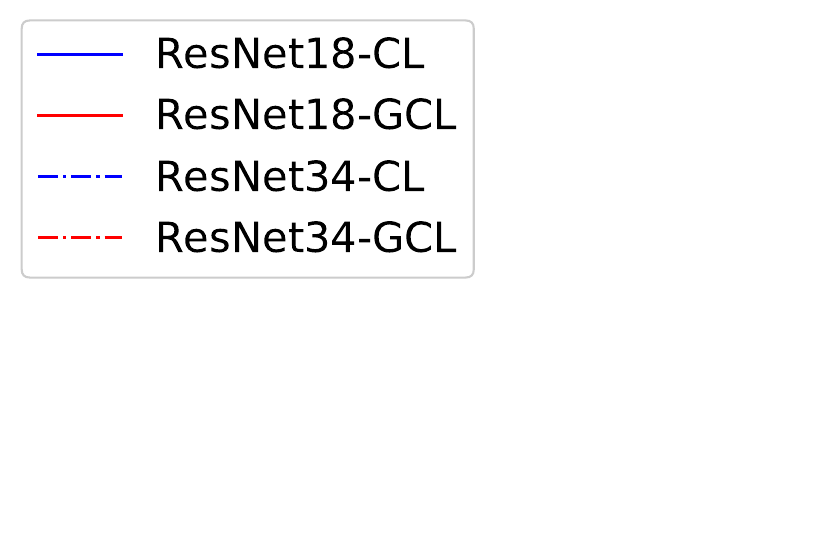}
}
\setcounter{subfigure}{3}
\hspace{-0.5cm}\subfloat[fire]{
\centering
\includegraphics[width=.25\textwidth]{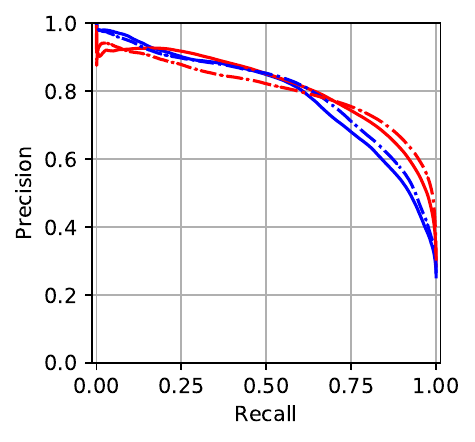}
}\hspace{-0.5cm}
\subfloat[redkitchen]{
\centering
\includegraphics[width=.25\textwidth]{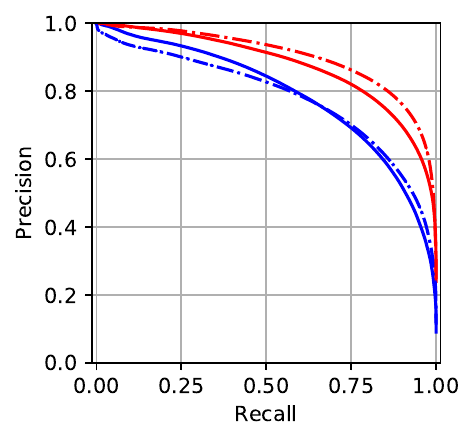}
}\hspace{-0.5cm}
\subfloat[chess]{
\centering
\includegraphics[width=.25\textwidth]{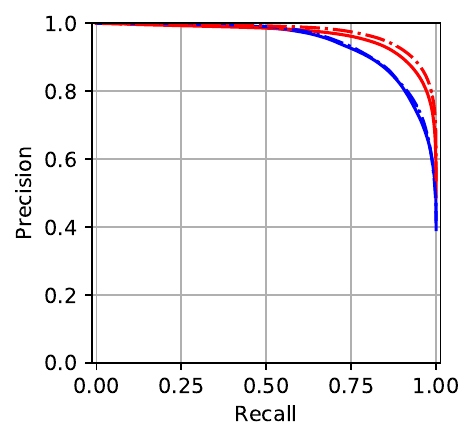}
}\hspace{-0.5cm}
\subfloat[office]{
\centering
\includegraphics[width=.25\textwidth]{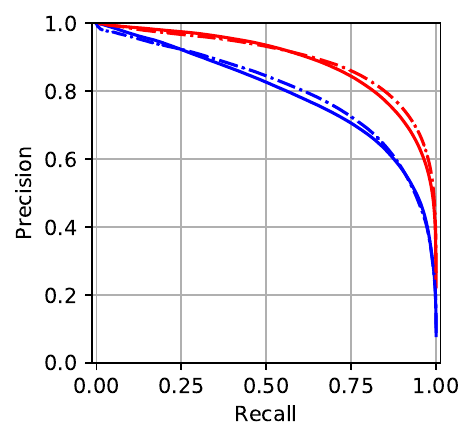}
}
 \caption{Precision-recall curves achieved on the 7Scenes dataset by ResNet18 and ResNet34 models trained using the Contrastive Loss function (CL, blue lines) and the proposed Generalized Contrastive loss (GCL, red lines) function.}
    \label{fig:7scenes_pr}
\end{figure*}


%

\begin{table*}[t!]
\centering
\caption{Median translation and rotation errors (in cm and degrees, respectively) on the 7Scenes dataset using the descriptors computed using the models trained with the binary Contrastive Loss and the Generalized Contrastive Loss for retrieval and InLoc for localization.}
\label{tab:7scenes_loc}
\setlength\tabcolsep{2.5pt}

\begin{tabular*}{\textwidth}{@{}l@{\extracolsep{\fill}}llllllll@{}}
\toprule
\textbf{Model} & \textbf{heads}    & \textbf{stairs} & \textbf{pumpkin} & \textbf{fire} & \textbf{redkitchen} & \textbf{office} & \textbf{chess} & \textbf{Mean}  \\ \midrule
NetVLAD-DensePE & 2cm 1.16$\degree$ & 9cm 2.47$\degree$ & 5cm 1.55$\degree$ & 3cm 1.07$\degree$ &4cm 1.31$\degree$ & 3cm 1.05$\degree$ & 3cm 1.05$\degree$& 4.14cm 1.38$\degree$\\
ResNet34-CL   & 2.98cm 2.44$\degree$      & 36.03cm 4.28$\degree$   & 7.12cm 1.97$\degree$     & 5.87cm 2.39$\degree$  & 6.53cm 2.12$\degree$        & 5.62cm 1.85$\degree$    & 6.01cm 1.99$\degree$   & 10.02cm 2.43$\degree$ \\
ResNet34-GCL  & 2.53cm 2.3$\degree$ & 32.13cm 3.9$\degree$    & 6.4cm 1.95$\degree$      & 5.13cm 2.11$\degree$  & 5.38cm 2.0$\degree$         & 5.16cm 1.7$\degree$     & 5.66cm 1.93$\degree$   & 8.91cm 2.27$\degree$  \\ \bottomrule
\end{tabular*}
\end{table*}

\subsection{Performance for different ground truths}

We performed an evaluation of the performance that the models that we train achieved when varying the ground truth threshold, i.e. when applying different cut values on the criteria that define two images as similar or not. We report results on the MSLS validation dataset.

First, we consider the distance, computed using the GPS coordinates associated to image pairs, between the location from where the images were taken and define a threshold on this value. For instance, we consider as similar all the image pairs that were taken from places with a distance lower than $D$, and vary $D$ between $5$ meters to $50$ meters. We report the result in  Fig.~\ref{fig:msls_dist}. An image is considered as correctly identified if at least one of the top-5 retrieved images is at a distance smaller than the considered threshold.

Furthermore, we consider the defined graded similarity $\psi$ and threshold it with values between $0$ and $0.9$. An image is considered as correctly identified if at least one of the top-5 retrieved images is labeled with a similarity greater than the considered threshold. We report the results in Fig.~\ref{fig:msls_psi_dist}.

In both cases, the models trained using the GCL function are more robust than those trained using the CL function to variations of the criteria and values used to establish the ground truth similarity between image pairs.

\subsection{Extra results on 7Scenes}

\textbf{Precision-recall curve.} We report the precision recall curves that we obtained on the 7Scenes dataset in Fig.~\ref{fig:7scenes_pr}. The models trained with the GCL function (red lines) consistently outperform their counterpart trained with the CL function (blue lines),  when they deploy both the ResNet18 and ResNet34 backbones. The results are consistent for all the sub-sets of the 7Scenes dataset. The case of the \emph{chess} scene is particularly interesting, for which the ResNet34-GCL model achieved near-perfect performance.

\revised{
\textbf{Localization. }
We tested the impact that a model trained with the GCL function and graded similarity has on the performance of a camera localization pipeline. We compared it to the case when the same model trained with the original binary Contrastive Loss function is used for the image retrieval task, and report the localization error that we obtained in Table~\ref{tab:7scenes_loc}. This experiment is meant only to demonstrate that the general expectation of better retrieval results contribute to better localization' is satisfied by the improved retrieval results achieved using the proposed GCL. For this experiment, we employed the InLoc algorithm~\cite{taira2018inloc}, using our learned representations with the ResNet34 backbone for the retrieval step. The images retrieved using the representation learned with GCL consistently led to an improvement of the localization accuracy w.r.t. the case in which the CL function is used. The enhanced performance of the retrieval task has a positive effect on the precision of localization algorithms. We also report the results of the original InLoc pipeline, based on and optimized to work with a NetVLAD backbone. While the GCL function contributes indirectly to better localization w.r.t. its binary CL counterpart, it requires more work to be effectively embedded in a localization pipeline.

}

%% file: figures/appendix/activations.pdf_tex
\begingroup%
  \makeatletter%
  \providecommand\color[2][]{%
    \errmessage{(Inkscape) Color is used for the text in Inkscape, but the package 'color.sty' is not loaded}%
    \renewcommand\color[2][]{}%
  }%
  \providecommand\transparent[1]{%
    \errmessage{(Inkscape) Transparency is used (non-zero) for the text in Inkscape, but the package 'transparent.sty' is not loaded}%
    \renewcommand\transparent[1]{}%
  }%
  \providecommand\rotatebox[2]{#2}%
  \newcommand*\fsize{\dimexpr\f@size pt\relax}%
  \newcommand*\lineheight[1]{\fontsize{\fsize}{#1\fsize}\selectfont}%
  \ifx\svgwidth\undefined%
    \setlength{\unitlength}{2531.79630247bp}%
    \ifx\svgscale\undefined%
      \relax%
    \else%
      \setlength{\unitlength}{\unitlength * \real{\svgscale}}%
    \fi%
  \else%
    \setlength{\unitlength}{\svgwidth}%
  \fi%
  \global\let\svgwidth\undefined%
  \global\let\svgscale\undefined%
  \makeatother%
  \begin{picture}(1,0.4915339)%
    \lineheight{1}%
    \setlength\tabcolsep{0pt}%
    \put(0,0){\includegraphics[width=\unitlength,page=1]{figures/appendix/activations.pdf}}%
    \put(0.10,0.495){\makebox(0,0)[lt]{\lineheight{1.25}\smash{\begin{tabular}[t]{l}Kampala (MSLS)\end{tabular}}}}%
    \put(0.34,0.495){\makebox(0,0)[lt]{\lineheight{1.25}\smash{\begin{tabular}[t]{l}Stockholm (MSLS)\end{tabular}}}}%
    \put(0.6,0.495){\makebox(0,0)[lt]{\lineheight{1.25}\smash{\begin{tabular}[t]{l}Pittsburgh\end{tabular}}}}%
    \put(0.86,0.495){\makebox(0,0)[lt]{\lineheight{1.25}\smash{\begin{tabular}[t]{l}Tokyo\end{tabular}}}}%
    \put(0,0){\includegraphics[width=\unitlength,page=2]{figures/appendix/activations.pdf}}%
    \put(0.14,0.005){\makebox(0,0)[t]{\lineheight{1.25}\smash{\begin{tabular}[t]{c}Low activation\end{tabular}}}}%
    \put(.84,0.005){\makebox(0,0)[lt]{\lineheight{1.25}\smash{\begin{tabular}[t]{l}High activation\end{tabular}}}}%
    \put(0.015,0.45){\rotatebox{90}{\makebox(0,0)[t]{\lineheight{1.25}\smash{\begin{tabular}[t]{c}Input\end{tabular}}}}}%
    \put(0.015,0.34){\rotatebox{90}{\makebox(0,0)[lt]{\lineheight{1.25}\smash{\begin{tabular}[t]{l}CL\end{tabular}}}}}%
    \put(0.015,0.06){\rotatebox{90}{\makebox(0,0)[lt]{\lineheight{1.25}\smash{\begin{tabular}[t]{l}GCL\end{tabular}}}}}%
    \put(0.015,0.16){\rotatebox{90}{\makebox(0,0)[lt]{\lineheight{1.25}\smash{\begin{tabular}[t]{l}CL\end{tabular}}}}}%
    \put(0.015,0.25){\rotatebox{90}{\makebox(0,0)[lt]{\lineheight{1.25}\smash{\begin{tabular}[t]{l}GCL\end{tabular}}}}}%
    \put(0.00,0.08){\rotatebox{90}{\makebox(0,0)[lt]{\lineheight{1.25}\smash{\begin{tabular}[t]{l}VGG16-GeM\end{tabular}}}}}%
    \put(0.00,0.25){\rotatebox{90}{\makebox(0,0)[lt]{\lineheight{1.25}\smash{\begin{tabular}[t]{l}ResNet50-GeM\end{tabular}}}}}%
    \put(0,0){\includegraphics[width=\unitlength,page=3]{figures/appendix/activations.pdf}}%

  \end{picture}%
\endgroup%

%% file: sections/appendix_fov.tex
\begin{figure}[!t]
    \subfloat[]{\centering
    \includegraphics[width=.45\columnwidth]{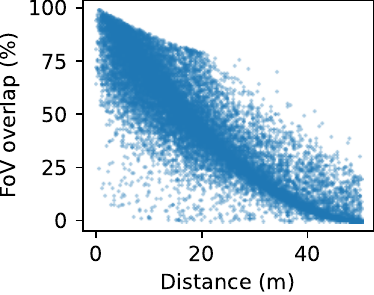}
    \label{fig:supp_fov_tr}
    }~\subfloat[]{\centering
    \includegraphics[width=.45\columnwidth]{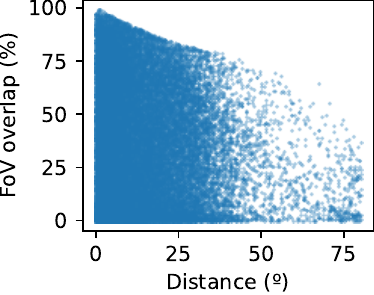}
        \label{fig:supp_fov_rot}
    }\\
     \subfloat[]{\centering
    \includegraphics[width=.45\columnwidth]{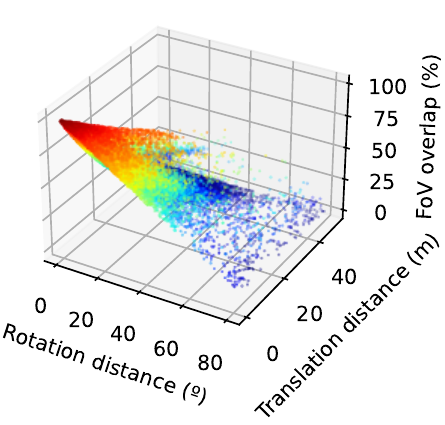}
        \label{fig:supp_fov_3d}
    }~\subfloat[]{\centering
    \includegraphics[width=.45\columnwidth]{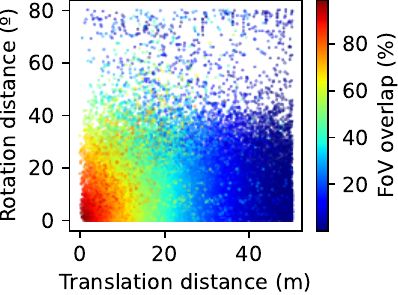}
        \label{fig:supp_fov_colormap}
    }
    \caption{Relation of 2D FoV overlap with translation and orientation distance. This data is computed from a subset of the MSLS training set.}
    \label{fig:supp_fov}
\end{figure}

\revised{\section{Field-of-View overlap}

We studied the relation among translation distance, rotation difference and the resulting FoV overlap measure. 
For that, we selected a subset of pairs of the MSLS training set and measured how the value of the FoV overlap varies with respect to the position and orientation difference of the image pairs. 

In Fig.~\ref{fig:supp_fov_tr}, we plot the relationship between the  translation distance and the value of the FoV overlap. We observed a somewhat linear relationship between increasing translation distance and decreasing FoV overlap. Moreover, many of the image pairs with FoV overlap of approximately 50\% were taken at a distance of about 25m. The variance that we observed in Fig.~\ref{fig:supp_fov_tr} can be attributed to the changes in orientation. Indeed, the relation between FoV overlap and orientation difference is less clear (see Fig.~\ref{fig:supp_fov_rot}). However, in general, the pairs with smaller orientation distance tend to have a higher FoV overlap. 

We plot the relation of the FoV overlap with both translation and orientation distance in a 3-dimensional plane in Fig.~\ref{fig:supp_fov_3d}, and as a heatmap in Fig.~\ref{fig:supp_fov_colormap}. 
Rotation and translation distance jointly influence the FoV overlap: the smaller the translation and orientation distance, the higher the similarity. Furthermore, as one of the distances has a higher value, the computer FoV overlap decreases, resulting in a lower annotated ground truth similarity.
}

%% file: sections/appendix_gradients.tex
\section{Gradient computations}
Let us consider two input images $x_i$ and $x_j$, their latent representations $\hat{f}(x_i)$ and $\hat{f}(x_j)$, and define $ d(x_i, x_j)$ the Euclidean distance between the representations, such as:
\begin{equation}
\nonumber
 d(x_i, x_j) = \left \| \hat{f}(x_i)-\hat{f}(x_j) \right \|_2
\end{equation}
For simplicity of notation, hereinafter we refer to $d(x_i, x_j)$ as $d$.

\subsection{Contrastive loss}

\noindent The Contrastive Loss function is defined as:
\begin{equation}
\nonumber
\mathcal{L}_{CL}=\begin{cases}
\frac{1}{2} d ^2 ,& \text{if } y= 1\\
\frac{1}{2}\max(\tau- d ,0)^2,& \text{if } y=0
\end{cases}
\end{equation}
\noindent where $y$ corresponds to the binary ground truth label and $\tau$ corresponds to the margin.

\noindent In order to compute the gradient for this function, we consider three cases, depending on the ground truth label $y$ and the value of the distance $d$. 

\noindent \textbf{Case 1) $y=1$}

\noindent The loss function becomes:
\begin{equation}
    \nonumber
    \mathcal{L}_{CL}=\frac{1}{2} d ^2
\end{equation}

\noindent and its derivative with respect to $d$ is:
\begin{align}
\nonumber
\begin{split}
    \nabla  \mathcal{L}_{CL} = \frac{\partial}{\partial d} \left( \mathcal{L}_{CL} \right) =\frac{\partial}{\partial d} \left( \frac{1}{2} d^2 \right)=  d
\end{split}
\end{align}

\noindent \textbf{Case 2) $y=0, d < \tau$}
\noindent The loss function becomes:
\begin{equation}
    \nonumber
    \mathcal{L}_{CL}=\frac{1}{2}(\tau- d)^2
\end{equation}

\noindent and its derivative with respect to $d$ is:
\begin{align*}
\nonumber
\begin{split}
\nabla {} \mathcal{L}_{CL}=&
\frac{\partial}{\partial d} \left(  \mathcal{L}_{CL} \right) =\frac{\partial}{\partial d} \left[ \frac{1}{2}(\tau- d)^2 \right]=\\
&=(\tau-d)(-1)= d-\tau
\end{split}
\end{align*}

\noindent  \textbf{Case 3) $y=0, d \geq \tau$}
\begin{equation}
    \nonumber
    \mathcal{L}_{CL} = 0
\end{equation}
and the gradient $\mathcal{L}_{CL} = 0$ as well. Thus, case 2 and case 3, for $y=0$, can be grouped as: 

\begin{equation}
    \nonumber
     \nabla  \mathcal{L}_{CL}=\begin{cases}
     d-\tau ,& \text{if } d < \tau\\
    0 ,& \text{if } d \geq \tau
\end{cases}
\end{equation}

\noindent and simplified as:

\begin{equation}
    \nonumber
     \nabla  \mathcal{L}_{CL}=\min( d-\tau,0)
\end{equation}

\noindent Finally, the \textbf{gradient of the Contrastive Loss function} is:

\begin{equation}
\nonumber
     \nabla  \mathcal{L}_{CL}=\begin{cases}
     d ,& \text{if } y= 1\\
    \min( d-\tau,0) ,& \text{if } y=0
\end{cases}
\end{equation}

\subsection{Generalized Contrastive loss}
We defined the Generalized Contrastive Loss function as:
\begin{equation}
\nonumber
 \mathcal{L}_{GCL}= \psi_{i,j}\cdot \frac{1}{2}d^2 + (1-\psi_{i,j}) \cdot \frac{1}{2}\max(\tau-d,0)^2
\end{equation}
\noindent where $\psi_{i,j}$ is the ground truth degree of similarity between the input images $x_i$ and $x_j$, and its values are in the interval $[0,1]$.
To compute the gradient of the GCL function we consider two cases, namely when 1) the distance $d$ between the representations is lower than the margin $\tau$ and 2) the alternative case when $d$ is larger than $\tau$. 

\noindent \textbf{Case 1) $d < \tau$}
\noindent The Generalized Contrastive loss function becomes:

\begin{equation}
\nonumber
 \mathcal{L}_{GCL}= \psi_{i,j}\cdot \frac{1}{2}d^2 + (1-\psi_{i,j}) \cdot \frac{1}{2}(\tau-d)^2 
\end{equation}

\noindent and its derivative with respect to $d$ is:
\begin{align*} 
\nonumber
\nabla  \mathcal{L}_{GCL}= \\
& = \frac{\partial}{\partial d} \mathcal{L}_{GCL}=\\
& = \frac{\partial}{\partial d} \left[ \psi_{i,j} \cdot \frac{1}{2} d^2  + (1 - \psi_{i,j}) \cdot \frac{1}{2}(\tau - d)^2 \right]=
\\&=\psi_{i,j} \cdot d + (1-\psi_{i,j}) (\tau-d)(-1)=
\\&=\psi_{i,j} \cdot d + d - \tau- \psi_{i,j} \cdot d + \psi_{i,j} \cdot \tau =
\\&=d+\tau(\psi_{i,j} -1)
\end{align*}

\noindent \textbf{Case 2) $d \geq \tau$}

\noindent The Generalized Contrastive loss function becomes:
\begin{equation}
\nonumber
 \mathcal{L}_{GCL}= \psi_{i,j}\cdot \frac{1}{2}d^2 + (1-\psi_{i,j}) \cdot \frac{1}{2}(0)^2 = \psi_{i,j}\cdot \frac{1}{2}d^2
\end{equation}

\noindent and its derivative with respect to $d$ is:
\begin{align*} 
\nonumber
\nabla \mathcal{L}_{GCL}=
\\&=\frac{\partial}{\partial d}\mathcal{L}_{GCL}=
\\&=\frac{\partial}{\partial d} \left[ \psi_{i,j}\cdot \frac{1}{2} d^2  \right]
\\&=d \cdot \psi_{i,j}
\end{align*}

\noindent Finally, the \textbf{gradient of the Generalized Contrastive Loss function} is:
\begin{equation}
     \nabla  \mathcal{L}_{GCL}=\begin{cases}d+\tau(\psi_{i,j}-1),& \text{if } d<\tau\\
     d \cdot \psi_{i,j},& \text{if } d \geq \tau\\
     \end{cases}
\end{equation}